\documentclass[10pt,twocolumn,letterpaper]{article}

\usepackage{iccv}
\usepackage{times}
\usepackage{epsfig}
\usepackage{graphicx}
\usepackage{amsmath}
\usepackage{amssymb}

\usepackage{booktabs}
\usepackage{algorithm}
\usepackage{algorithmic}
\usepackage{float}
\usepackage{color}
\usepackage{colortbl}
\usepackage{verbatim}
\usepackage{tabularx}
\usepackage{multirow}
\usepackage{setspace}

\usepackage{color}
\usepackage{colortbl}
\usepackage{comment}
\usepackage{subfigure}
\usepackage{listings}

\makeatletter
\def\hlinewd#1{%
\noalign{\ifnum0=`}\fi\hrule \@height #1 \futurelet
\reserved@a\@xhline}

\usepackage[breaklinks=true,bookmarks=false]{hyperref}

\iccvfinalcopy % *** Uncomment this line for the final submission

\def\iccvPaperID{****} % *** Enter the ICCV Paper ID here

\ificcvfinal\fi

\begin{document}

%%%%%%%%% TITLE
\title{MaskingDepth: Masked Consistency Regularization for Semi-supervised Monocular Depth Estimation}

\author{
% Authors
    Jongbeom Baek \textsuperscript{\rm 1}\thanks{Equal contribution} \quad 
    Gyeongnyeon Kim \textsuperscript{\rm 1}\samethanks \quad 
    Seonghoon Park \textsuperscript{\rm 1}\samethanks \quad 
    Honggyu An \textsuperscript{\rm 1} \\
    Matteo Poggi \textsuperscript{\rm 2} \quad 
    Seungryong Kim \textsuperscript{\rm 1} 
    \thanks{Corresponding author}\\
\textsuperscript{\rm 1}Korea University, Seoul, Korea \quad \textsuperscript{\rm 2}University of Bologna, Bologna, Italy\\
{\tt\small \{baem0911,kkn9975,seong0905,hg010303,seungryong\_kim\}@korea.ac.kr} \quad 
{\tt\small m.poggi@unibo.it}
}

\newcommand*\samethanks[1][\value{footnote}]{\footnotemark[#1]}
% \author{Jongbeom Baek, Gyongnyeon Kim, Seonghoon Park, Honggyu An\\
% Korea University, Seoul, Korea  University\\
% {\tt\small {baem0911,kkn9975,seong0905,hg010303,seungryong kim}@korea.ac.kr} 
% % {\tt\small m.poggi@unibo.it}
% \and
% % Second Author\\

\maketitle

%%%%%%%%% ABSTRACT

%%%%%%%%% ABSTRACT
\begin{abstract}
We propose MaskingDepth, a novel semi-supervised learning framework for monocular depth estimation to mitigate the reliance on large ground-truth depth quantities. MaskingDepth is designed to enforce consistency between the strongly-augmented unlabeled data and the pseudo-labels derived from weakly-augmented unlabeled data, which enables learning depth without supervision. In this framework, a novel data augmentation is proposed to take the advantage of a na\"ive masking strategy as an augmentation, while avoiding its scale ambiguity problem between depths from weakly- and strongly-augmented branches and risk of missing small-scale instances. To only retain high-confident depth predictions from the weakly-augmented branch as pseudo-labels, we also present an uncertainty estimation technique, which is used to define robust consistency regularization. Experiments on KITTI and NYU-Depth-v2 datasets demonstrate the effectiveness of each component, its robustness to the use of fewer depth-annotated images, and superior performance compared to other state-of-the-art semi-supervised methods for monocular depth estimation. Furthermore, we show our method can be easily extended to domain adaptation task. Our code is available at \url{https://github.com/KU-CVLAB/MaskingDepth}.
\end{abstract}

%%%%%%%%% BODY TEXT
\section{Introduction}
\label{sec:intro}
Monocular depth estimation, aiming to predict a dense depth map from a single image, has been one of the most essential tasks in numerous applications, such as augmented reality (AR)~\cite{lee2011depth}, virtual reality (VR)~\cite{huang20176}, and autonomous driving~\cite{yogamani2019woodscape, kumar2020fisheyedistancenet}. 

As a pioneering work, Eigen et al.~\cite{eigen2014depth} first introduced a deep learning-based approach for this task, and several works~\cite{li2015depth,kim2016,ummenhofer2017demon, liu2015learning, eigen2015predicting,fu2018deep,lee2019big,ranftl2020towards,ranftl2021vision} have achieved higher accuracy throughout the years. These methods are mostly formulated in a supervised learning regime, which requires a large number of images and corresponding ground-truth depths. These are notoriously challenging to obtain~\cite{geiger2012we,silberman2012indoor} compared to other types of annotation, such as image class labels~\cite{sohn2020fixmatch} and segmentation labels~\cite{zou2020pseudoseg}. To overcome this problem, self-supervised learning techniques~\cite{zhou2017unsupervised,godard2017unsupervised,godard2019digging} have emerged, which formulate monocular depth estimation as an image reconstruction problem. Although this strategy seems to be an attractive solution, these methods often require extra data, such as stereo pairs or video sequences which are not always available~\cite{zhou2017unsupervised, godard2017unsupervised, godard2019digging}. In addition, they are known to often generate blurred depth at object boundaries~\cite{choi2021adaptive, baek2022semi}.

\begin{figure}[t]
% \captionsetup[subfigure]{labelformat=empty}
\begin{center}
\renewcommand{\thesubfigure}{}
\subfigure[]
{\includegraphics[width=0.325\linewidth]{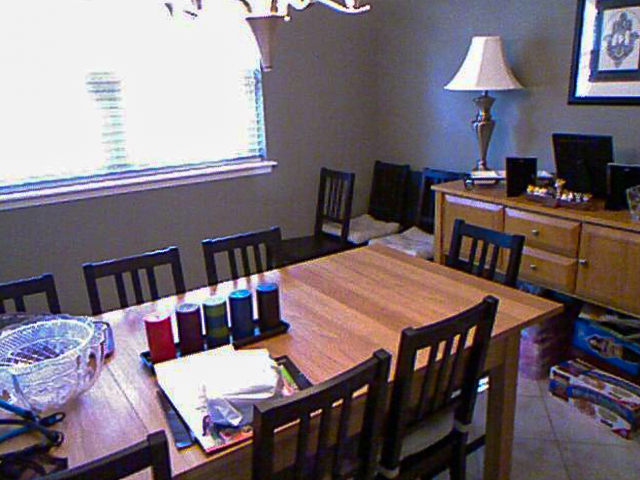}} 
\subfigure[]
{\includegraphics[width=0.325\linewidth]{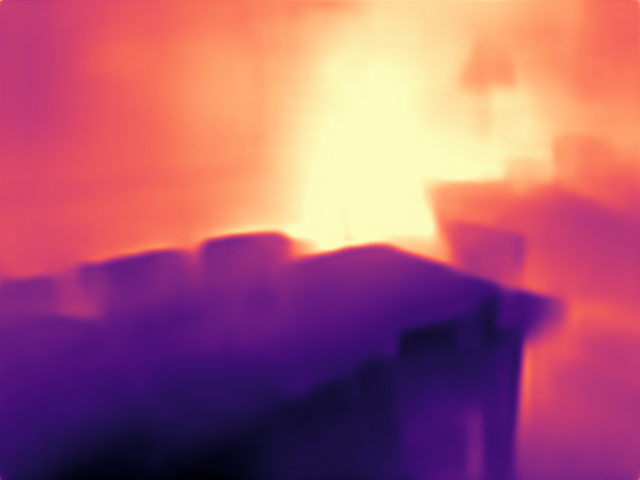}} 
\subfigure[]
{\includegraphics[width=0.325\linewidth]{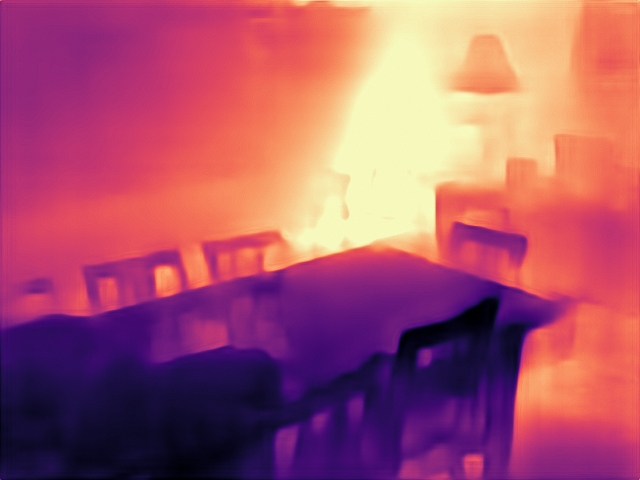}} \\ 
\vspace{-21pt}
\subfigure[(a) RGB images]
{\includegraphics[width=0.325\linewidth]{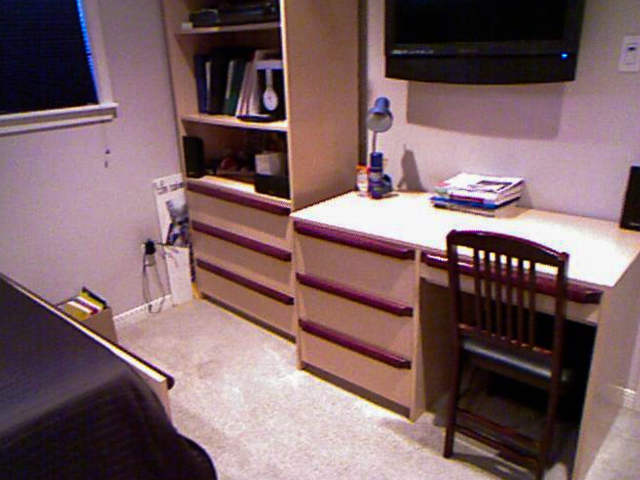}}
\subfigure[(b) Supervised]
{\includegraphics[width=0.325\linewidth]{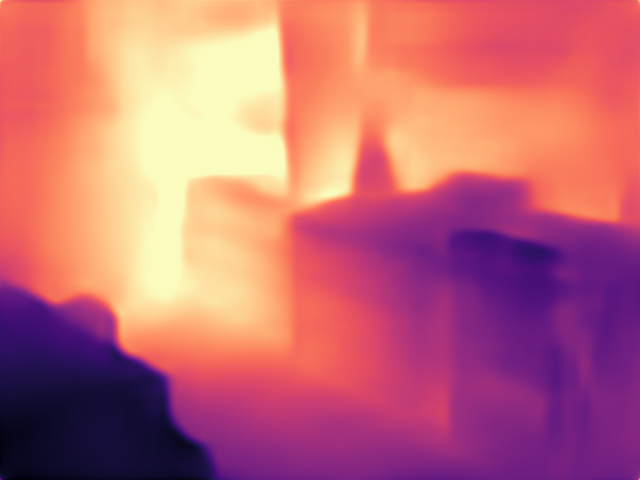}}
\subfigure[(c) \textbf{MaskingDepth}]
{\includegraphics[width=0.325\linewidth]{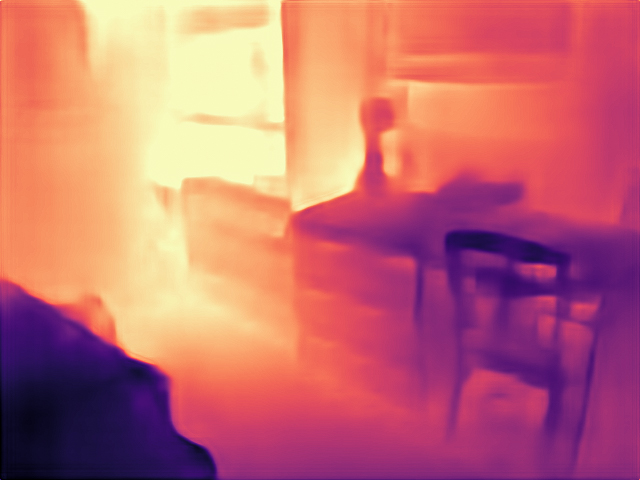}}\\
\end{center}
% \vspace{-20pt}
\vspace{-10pt}
\caption{\textbf{Qualitative results of MaskingDepth:} Our proposed MaskingDepth produces high quality depth maps by leveraging a large amount of unlabeled data. 
}%
\label{fig_main}\vspace{-10pt}
\end{figure}

\begin{figure*}[t]
\begin{center}
\renewcommand{\thesubfigure}{}
\subfigure[(a)]
{\includegraphics[width=0.43\linewidth]{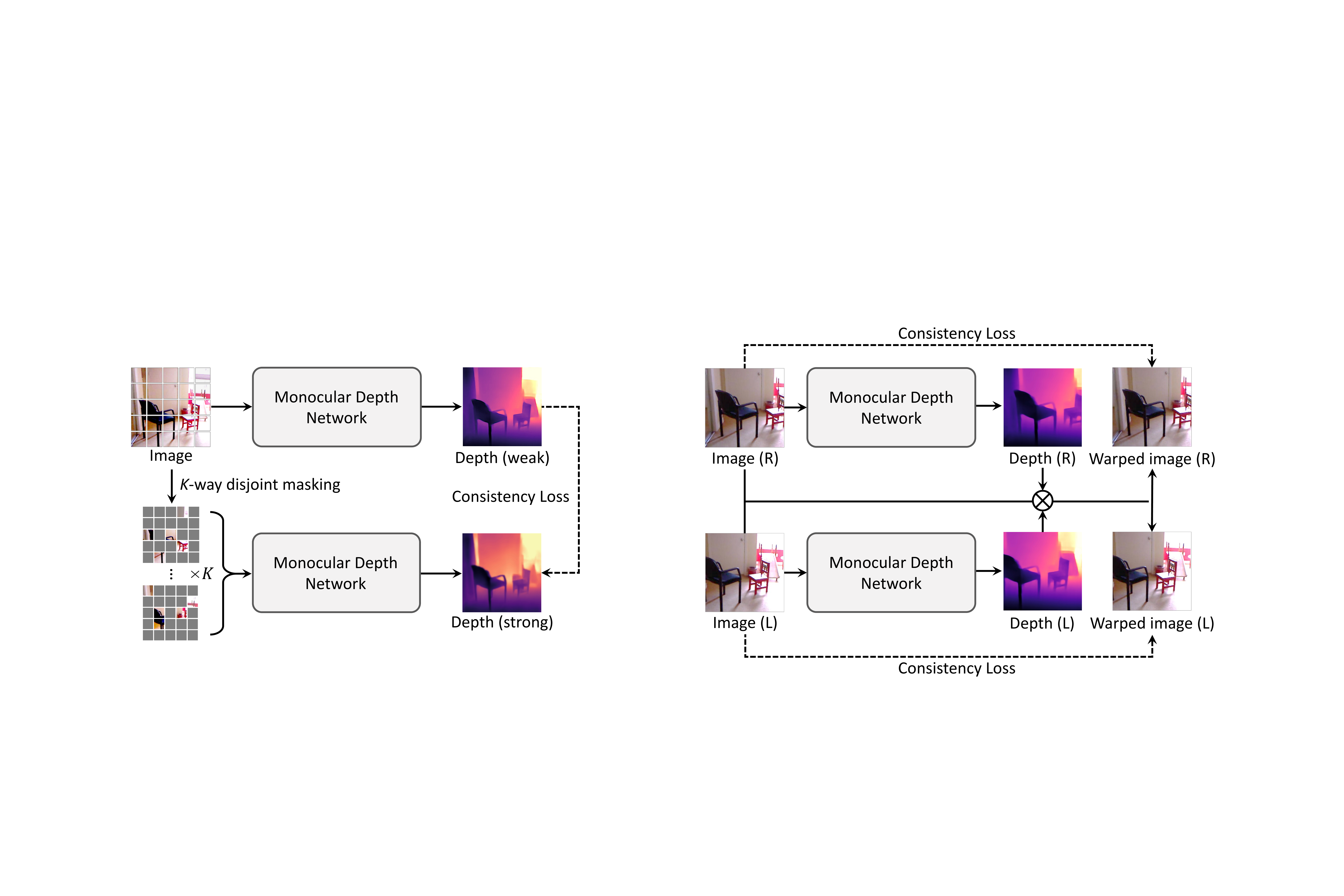}} 
\hspace{20pt}
\subfigure[(b)]
{\includegraphics[width=0.425\linewidth]{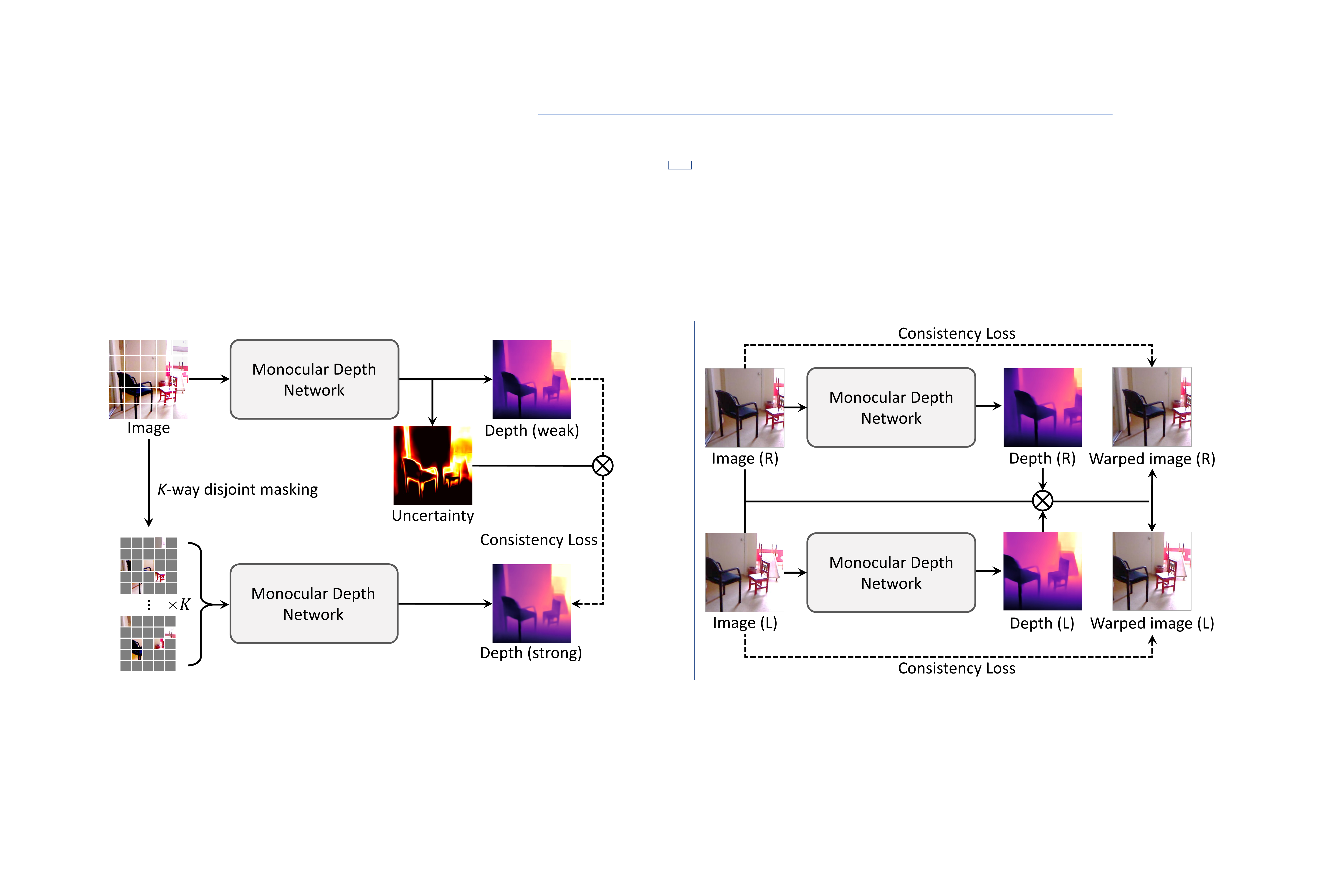}} 
\end{center}
% \vspace{-21pt}
\vspace{-10pt}
\caption{\textbf{Motivation:} (a) existing self-supervised methods~\cite{godard2019digging, garg2016unsupervised, godard2017unsupervised} that enforce consistency between one image and another warped image from stereo or sequential frames, which often results in blurred artifacts at object boundaries, and (b) our framework that uses uncertainty-aware consistency regularization by adopting $K$-way disjoint masking for augmentation, which shows highly improved performance with sparse depth data in a semi-supervised learning setting.}%
\label{fig1}\vspace{-10pt}
\end{figure*}

On the other hand, semi-supervised learning approaches using stereo-based knowledge distillation~\cite{tonioni2019unsupervised, cho2019large,choi2021adaptive} have been proposed. However, they are constrained by the need for stereo image pairs and additional computation costs for training a stereo model. Moreover, some works~\cite{kuznietsov2017semi, amiri2019semi} have attempted to simply combine supervised learning and reconstruction-based self-supervised learning~\cite{zhou2017unsupervised, godard2017unsupervised, godard2019digging}, but they directly inherit of limitations of existing self-supervised learning methods.

In this paper, for the first time, we present a novel semi-supervised learning framework, called MaskingDepth, for monocular depth estimation based on an uncertainty-aware consistency regularization. Our framework enforces consistency between strongly-augmented unlabeled data and pseudo-labels obtained from weakly-augmented unlabeled data. To apply enough perturbations to an input image, we propose a new data augmentation, called $K$-way disjoint masking, inspired by recent masked image modeling strategies for vision Transformers~\cite{bao2021beit, xie2021simmim, he2021masked}. The na\"ive masking technique~\cite{bao2021beit, xie2021simmim, he2021masked} yielded superior performance on classification tasks such as image classification and semantic segmentation. However, adopting this to monocular depth estimation, which heavily relies on context information to estimate the depth, may cause scale ambiguity and omit the context of small objects~\cite{he2021masked}. To overcome this, the $K$-way disjoint masking jointly decodes scattered tokens that are encoded from a $K$-disjoint set of tokens independently, and thus mitigates the scale ambiguity and restores the full context from the image while retaining the benefits of masking-based data augmentation. In our framework, we encourage depth and feature consistencies~\cite{bao2021beit} across two branches from two augmented views by $K$-way disjoint masking. In addition, we propose an uncertainty estimation technique~\cite{kendall2017uncertainties,poggi2020uncertainty} to aid depth consistency and facilitate convergence by filtering out the noise of pseudo labels.

In the experiments, we evaluate MaskingDepth on standard benchmarks, including KITTI~\cite{geiger2012we} and NYU-Depth-v2~\cite{silberman2012indoor}, showing outstanding performance compared to previous methods. 
We validate each component through an extensive ablation study and demonstrate that our approach can be easily used for domain adaptation. 

\section{Related Work}
\label{sec:rel}
\paragraph{Monocular depth estimation.}
Monocular depth estimation aims at estimating a depth from a single image. \cite{eigen2014depth} pioneering work tackled this task with deep neural networks. Since then, several approaches~\cite{li2015depth, ummenhofer2017demon} based on supervised learning, i.e., utilizing ground-truth labels, have been presented to improve performance. Although these methods have achieved remarkable accuracy over traditional, handcrafted approaches~\cite{saxena20083,ladicky2014pulling}, their success depends on massive amounts of ground-truth depth maps that require a labor-intensive process for collection and cleaning~\cite{geiger2012we,silberman2012indoor}.

\begin{figure*}[t]
\centering
\includegraphics[width=1.0\linewidth]{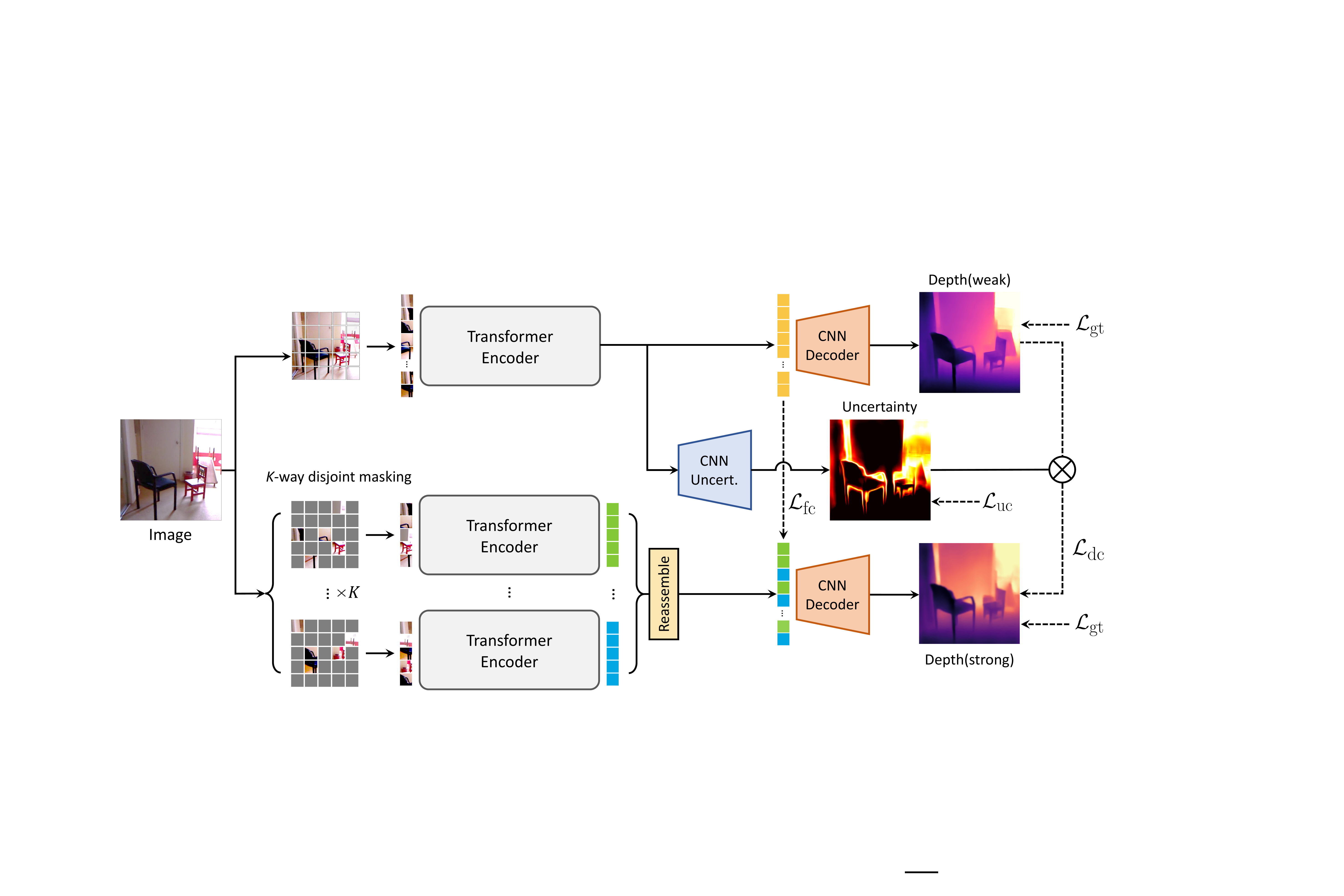}\hfill\\
\vspace{5pt}
\caption{\textbf{Overview of MaskingDepth architecture.} It consists of two main components; a branch using full tokens (top), and a branch using $K$-way disjoint masked tokens (bottom), where $K$-number of subsets are encoded independently and concatenated before decoding. We use consistency loss $\mathcal{L}_\mathrm{dc}$ to make predictions between original and augmented images consistent, aided by an uncertainty measure. Feature consistency loss $\mathcal{L}_\mathrm{fc}$ is also applied to facilitate the convergence.}
\vspace{-10pt}
\label{fig:network}
\end{figure*}

To address this limitation, self-supervised learning methods~\cite{xie2016deep3d,garg2016unsupervised,godard2019digging} formulate the task as an image reconstruction, leveraging geometric information over stereo pairs or a sequence of frames. These approaches have emphasized the importance to mitigate the dependency on annotations. However, they often produces indistinct depth result near object boundaries and disregards occluded pixels~\cite{choi2021adaptive}. Unlike both of the aforementioned approaches, there has not been much work on semi-supervised depth estimation. \cite{kuznietsov2017semi} simply combined supervised and self-supervised loss functions and \cite{amiri2019semi} utilized a left-right consistency to improve performance. Recently, several works~\cite{guo2018learning,cho2019large,tosi2019learning,watson2019self, choi2021adaptive} have attempted to use stereo knowledge to distill labels for monocular depth estimation. However, they are still constrained by the need for specific data (stereo pairs) and additional computation cost (training a stereo module). Our proposed framework alleviates the reliance on labeled data by leveraging consistency regularization which allows us to use unlabeled data.

\vspace{-10pt}
\paragraph{Masked image modeling.}
Masked image modeling is the process of learning representations by reconstructing images that are corrupted by masking~\cite{xie2021simmim,he2021masked,bao2021beit}.
After BERT~\cite{devlin2018bert} proposed the masked language modeling task, one of the most successful methods for pre-training in NLP, related works have explored a variety of masked image prediction strategies suited for Transformers~\cite{bao2021beit, he2021masked}. ViT~\cite{dosovitskiy2020image} studied a masked patch prediction to facilitate representation learning, and BEiT~\cite{bao2021beit} extended upon this by predicting discrete tokens. Recent literature~\cite{xie2021simmim, he2021masked} introduce an extremely simple yet effective approach. Masked autoencoder (MAE)~\cite{he2021masked} utilizes only unmasked tokens to encode meaningful representations. In addition, MRA~\cite{xu2022masked} leverages this strategy to generate augmented images. In this paper, we propose an effective data augmentation strategy that takes advantage of masked image modeling.

\section{Methodology}
\label{sec:method}
\subsection{Problem Formulation}
Let us denote a color image and its corresponding depth map as $I$ and $D$, respectively. The objective of monocular depth estimation is to learn a mapping function $f$ from the image $I$ to its corresponding depth $D$ such that $D=f(I)$. Recent learning-based methods formulate the mapping function with convolutions~\cite{eigen2014depth, wang2015designing, kim2016} or Transformers~\cite{ranftl2021vision, yang2021transformer, li2022depthformer} as a neural network $f_\theta$ with parameters $\theta$. To train the monocular depth estimation networks $f_\theta$ in a supervised manner, the ground-truth $D_\mathrm{gt}$ is required, but building large-scale dense depth data is notoriously challenging~\cite{garg2016unsupervised,xie2016deep3d}. In addition, to alleviate depth capture errors, post-processing~\cite{uhrig2017sparsity,geiger2012we,silberman2012indoor} is essential, which introduces further burden.

\begin{figure*}[t]
\begin{center}
\renewcommand{\thesubfigure}{}
\subfigure[]
{\includegraphics[width=0.245\linewidth]{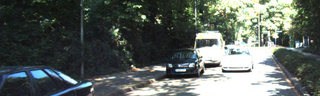}} 
\subfigure[]
{\includegraphics[width=0.245\linewidth]{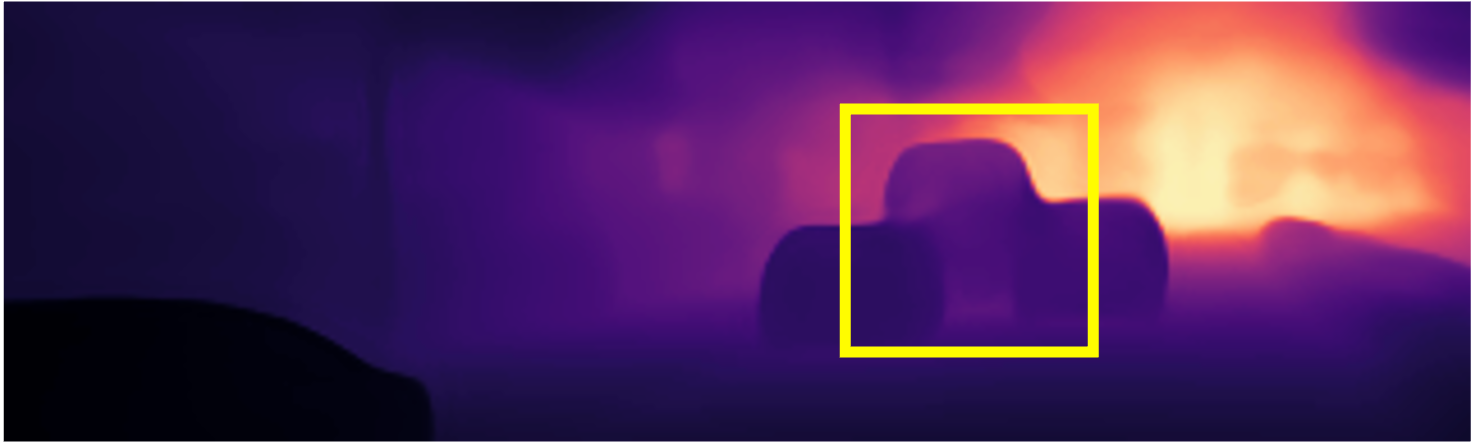}} 
\subfigure[]
{\includegraphics[width=0.245\linewidth]{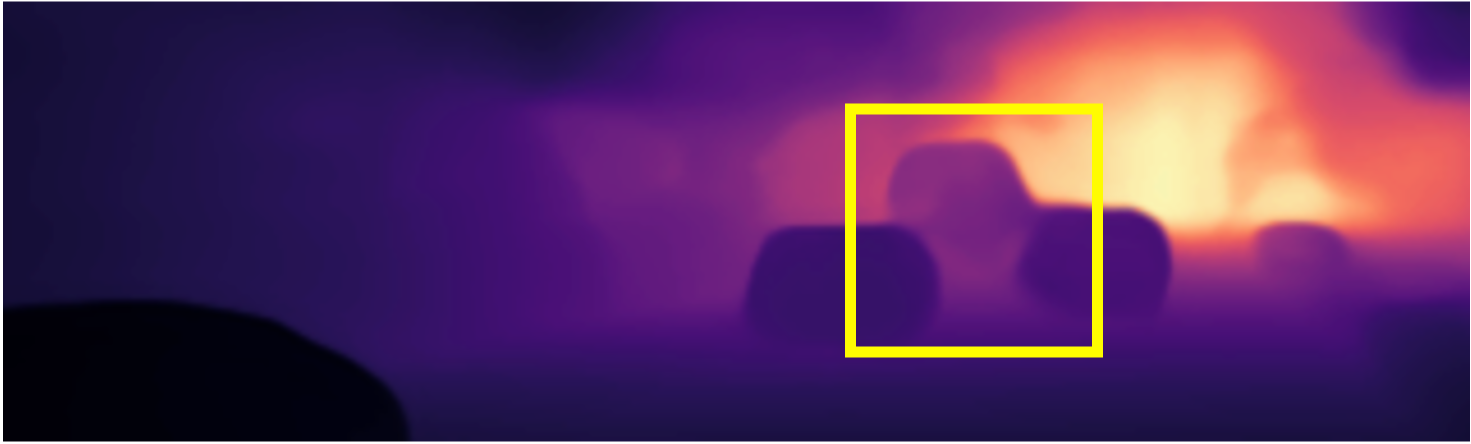}}
\subfigure[]
{\includegraphics[width=0.245\linewidth]{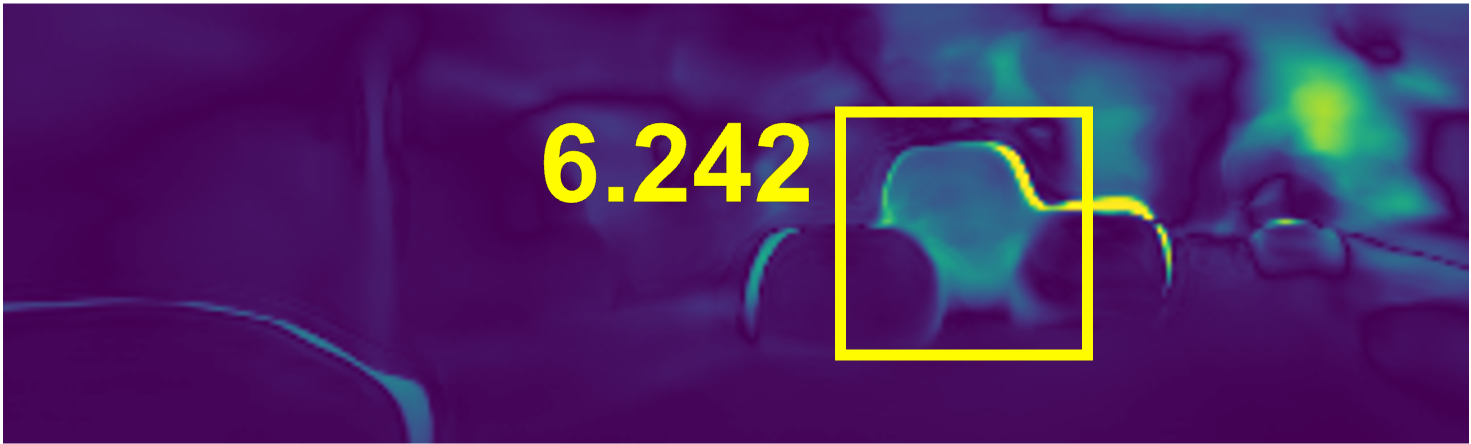}}\\

\vspace{-22pt}
\subfigure[(a) RGB images]
{\includegraphics[width=0.245\linewidth]{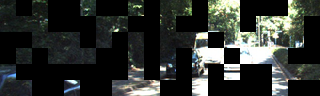}} 
\subfigure[(b) Depths (weak)]
{\includegraphics[width=0.245\linewidth]{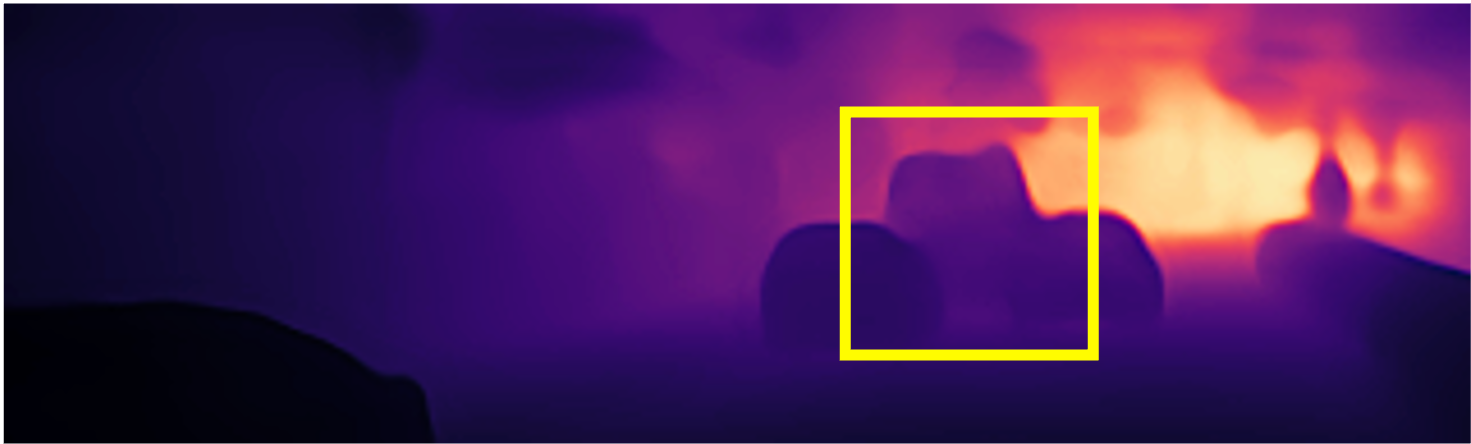}} 
\subfigure[(c) Depths (strong)]
{\includegraphics[width=0.245\linewidth]{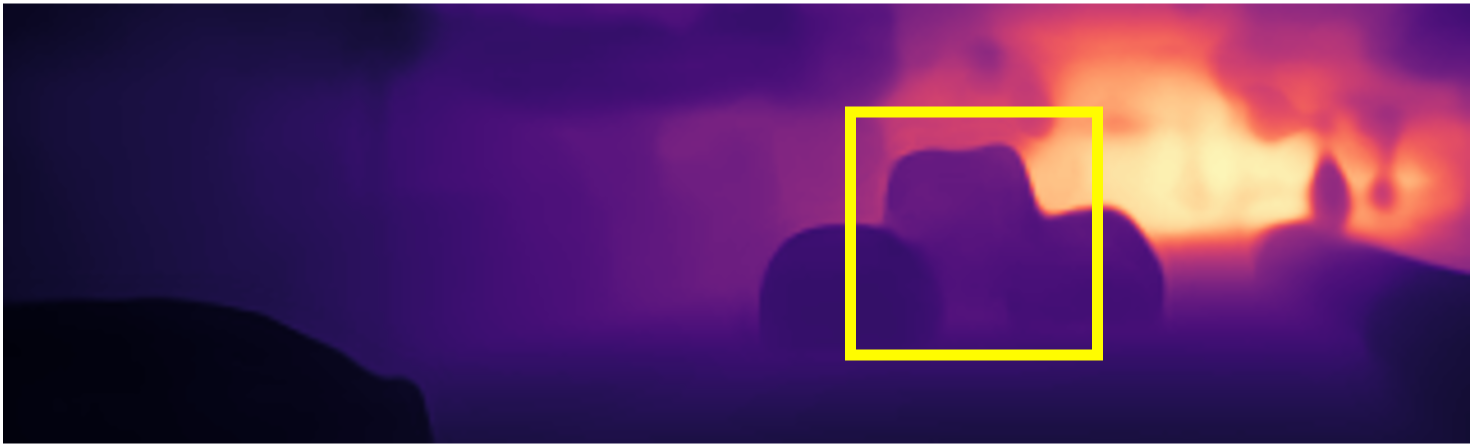}}
\subfigure[(d) Difference maps]
{\includegraphics[width=0.245\linewidth]{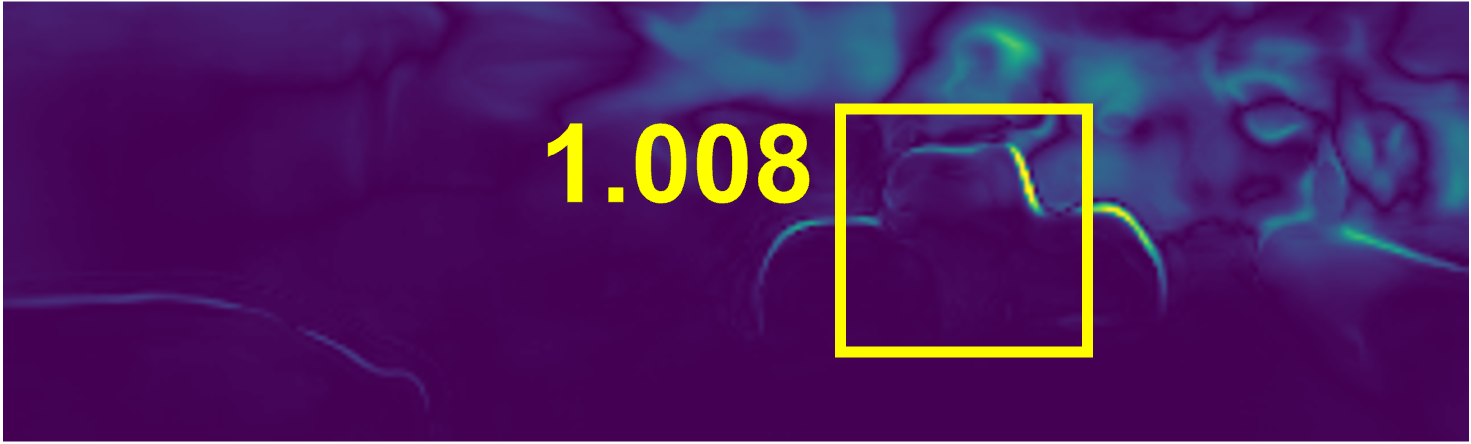}}\\

\end{center}
% \vspace{-17.5pt}
% \vspace{-20pt}
\vspace{-10pt}
\caption{\textbf{Effectiveness of our masking strategy to handle scale ambiguity:} (a) RGB image and its masked one. The first row of (b) and (c) are na\"ive masking results, and the second row of (b) and (c) are our masking results. In (d), we visualize the difference map between (b) and (c). We denote the mean scale difference in the boxed area. Our method better generates scale-consistent results, which in turns helps to better learn the monocular depth estimation networks.}%
\label{fig_sa}\vspace{-10pt}
\end{figure*}

\begin{figure}[t]
\begin{center}
\subfigure[Na\"ive masking]
{\includegraphics[width=0.335\linewidth]{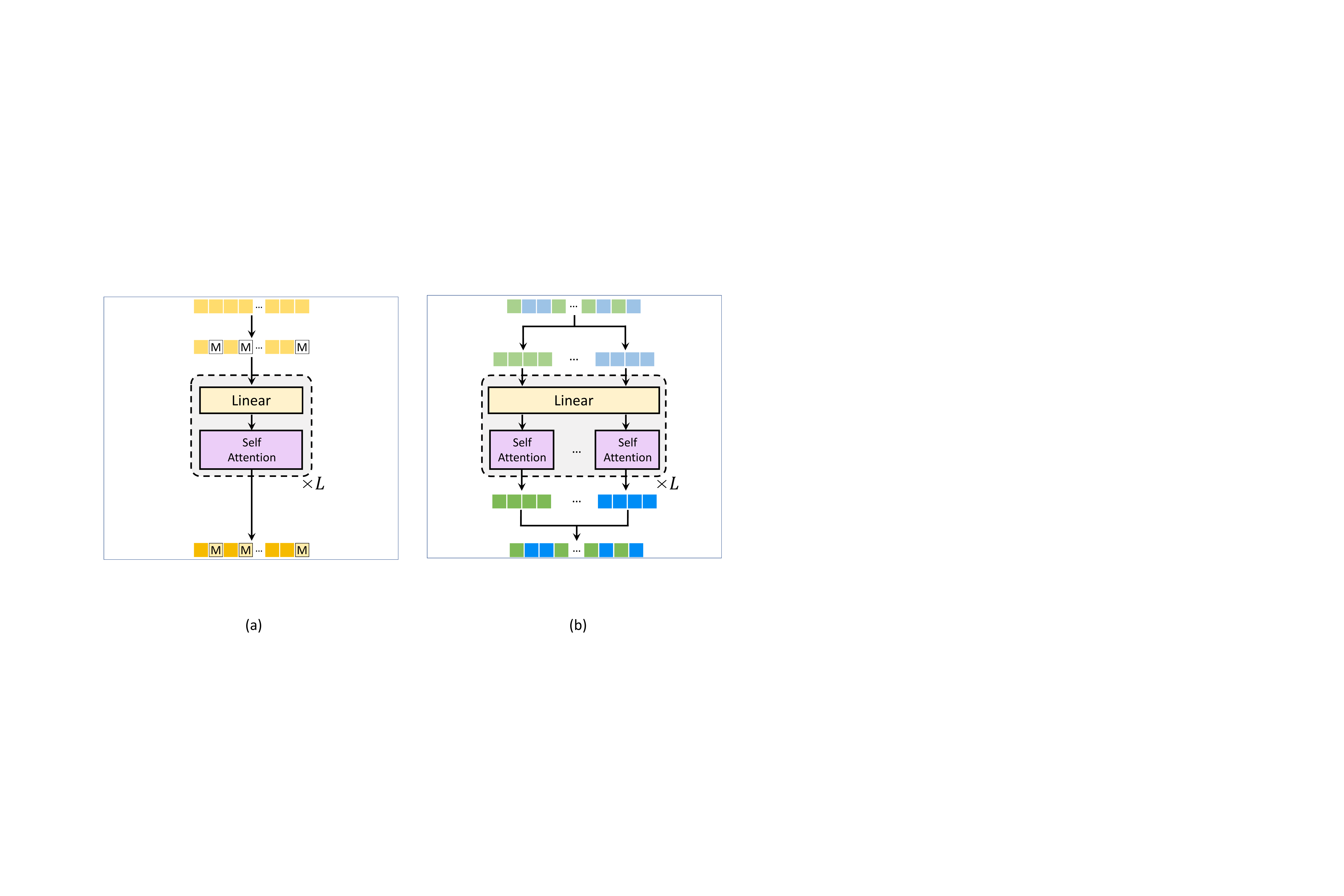}}
\hspace{15pt}
\subfigure[K-way disjoint masking]
{\includegraphics[width=0.491\linewidth]{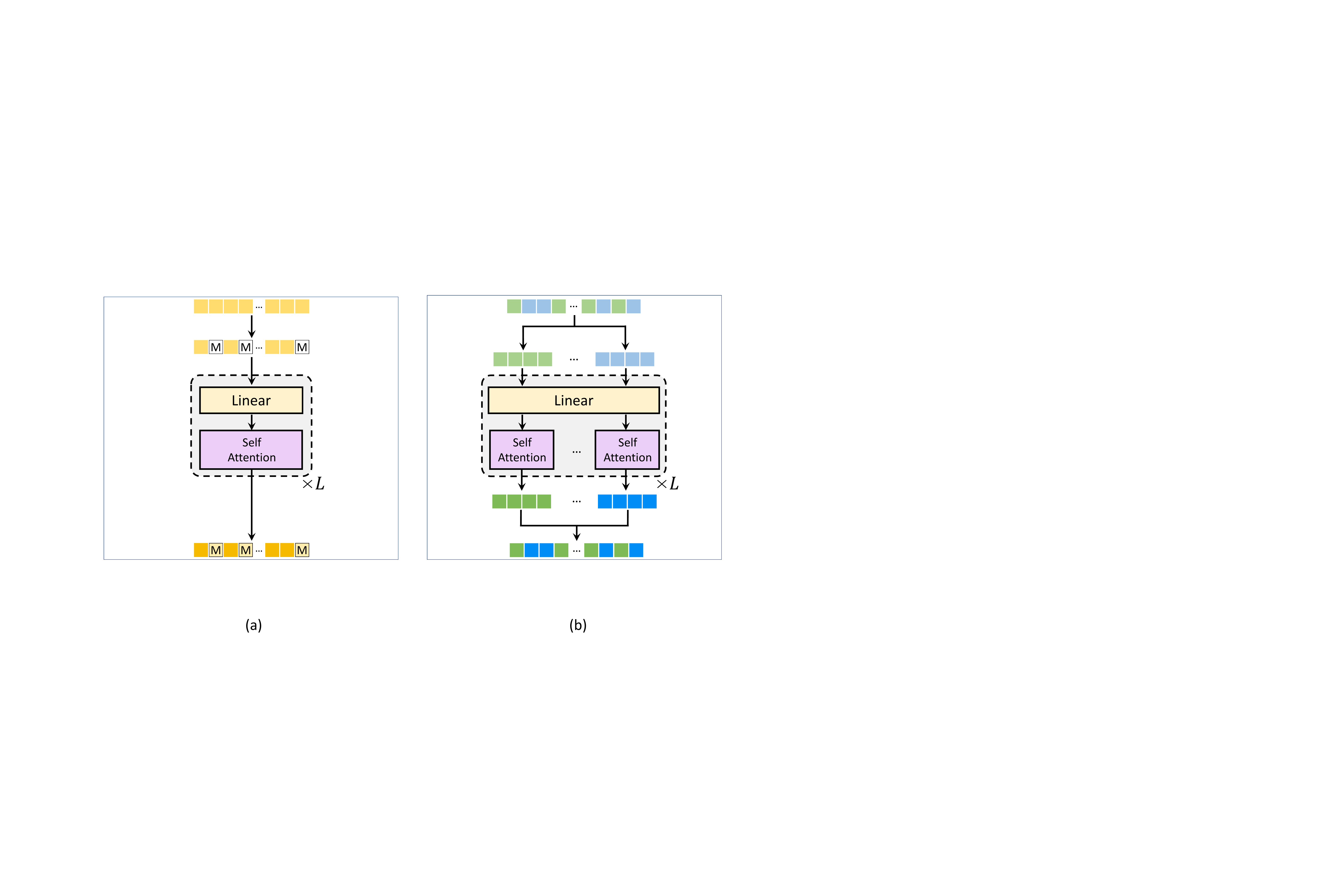}}
\end{center}
% \vspace{-20pt}
\vspace{-10pt}
\caption{\textbf{Illustration of masking:} (a) na\"ive token masking~\cite{xie2021simmim,bao2021beit} and (b) our K-way disjoint masking. Compared to conventional masking~\cite{xie2021simmim,bao2021beit}, our masking strategy consistently well produces dense depth maps because it considers the entire set of tokens during decoding.}
\label{fig:detail}
\vspace{-10pt}
\end{figure}

In this paper, we present a novel semi-supervised learning framework that facilitates the model to learn monocular depth estimation on a large number of unlabeled data with sparse depth annotations. As shown in Fig.~\ref{fig1}, for the first time we introduce consistency regularization between two differently augmented views from the same image to train the monocular depth estimation network.  

\subsection{Overview}
Given an image $I$, we build two different branches, one for processing an image passed through weak augmentation (called weak branch) and the other for a strongly augmented image (called strong branch), where the consistency between the two images through the networks $f_\theta$ is encouraged. In particular, a weakly-augmented image $I_\mathrm{weak}$ and a strongly-augmented image $I_\mathrm{strong}$ are fed to the network $f_\theta$, and then consistency is defined as follows: 
\begin{equation}
    \mathcal{L} =  \mathcal{D}(\mathrm{sg}(f_{\theta}(I_\mathrm{weak})), f_{\theta}(I_\mathrm{strong})),
\end{equation}
where $\mathrm{sg(\cdot)}$ is a stop-gradient operation and $\mathcal{D}(\cdot,\cdot)$ is a distance function, e.g., mean squared error (MSE)~\cite{wang2009mean} or KL-divergence~\cite{kullback1997information}. In this setting, we can interpret $f_{\theta}(I_\mathrm{weak})$ as a pseudo ground-truth or pseudo-label. To effectively implement this strategy, appropriate data augmentation techniques are important.

However, it is challenging to make difference between two branches in monocular depth estimation since depth-specific data augmentation techniques have been rarely studied~\cite{ishii2021cutdepth, kim2022global}. Furthermore, conventional data augmentation techniques such as crop~\cite{devries2017improved} and rotation~\cite{gidaris2018unsupervised}, which have been successfully used in classification tasks, are no longer effective for monocular depth estimation as they can lead to geometric inconsistency~\cite{dijk2019neural}.

To address this issue, we present a novel data augmentation technique, inspired by recent masked image modeling techniques~\cite{xie2021simmim, he2021masked,bachmann2022multimae}, which allows for generating geometrically consistent pseudo depth maps while applying sufficient perturbations to the inputs.
As illustrated in Fig.~\ref{fig:network}, our proposed framework follows the backbone model $f_\theta$~\cite{ranftl2021vision} by consisting of a Transformer-based encoder$f^\mathrm{enc}_\theta$, which takes a tokenized image and outputs encoded features~\cite{dosovitskiy2020image}, and a CNN-based decoder$f^\mathrm{dec}_\theta$. In addition, we encourage not only feature similarities~\cite{bao2021beit} but also depth similarities between the two branches processing the two augmented views. To aid the latter, we present uncertainty estimation~\cite{kendall2017uncertainties,poggi2020uncertainty} that helps the convergence of training by filtering out the noise of pseudo labels.

\begin{figure}[t]
% \captionsetup[subfigure]{labelformat=empty}
\begin{center}
\renewcommand{\thesubfigure}{}
\subfigure[(a)]
{\includegraphics[width=0.325\linewidth]{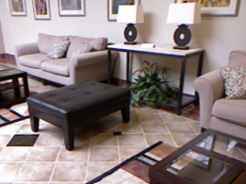}}
\subfigure[(c)]
{\includegraphics[width=0.325\linewidth]{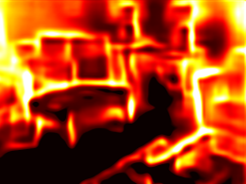}}
\subfigure[(e)]
{\includegraphics[width=0.325\linewidth]{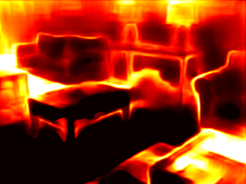}} \hfill \\
\vspace{-7pt}
\subfigure[(b)]
{\includegraphics[width=0.325\linewidth]{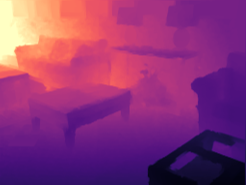}}
\subfigure[(d)]
{\includegraphics[width=0.325\linewidth]{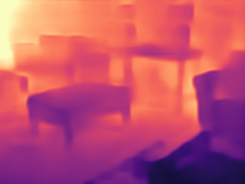}}
\subfigure[(f)]
{\includegraphics[width=0.325\linewidth]{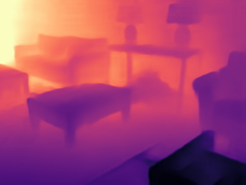}} \hfill

\end{center}
% \vspace{-20pt}
\vspace{-15pt}
\caption{\textbf{Examples of estimated depths and uncertainty maps:} (a) RGB,  (b) ground-truth depth map, predicted depth maps and their corresponding uncertainty maps by the networks trained with (c), (d) 100 and (e), (f) 10,000 labeled frames and $40K$ unlabeled frames.}%
\label{fig4}\vspace{-10pt}
\end{figure}

\begin{figure*}[t]
\begin{center}
\renewcommand{\thesubfigure}{}
\subfigure[]
{\includegraphics[width=0.33\linewidth]{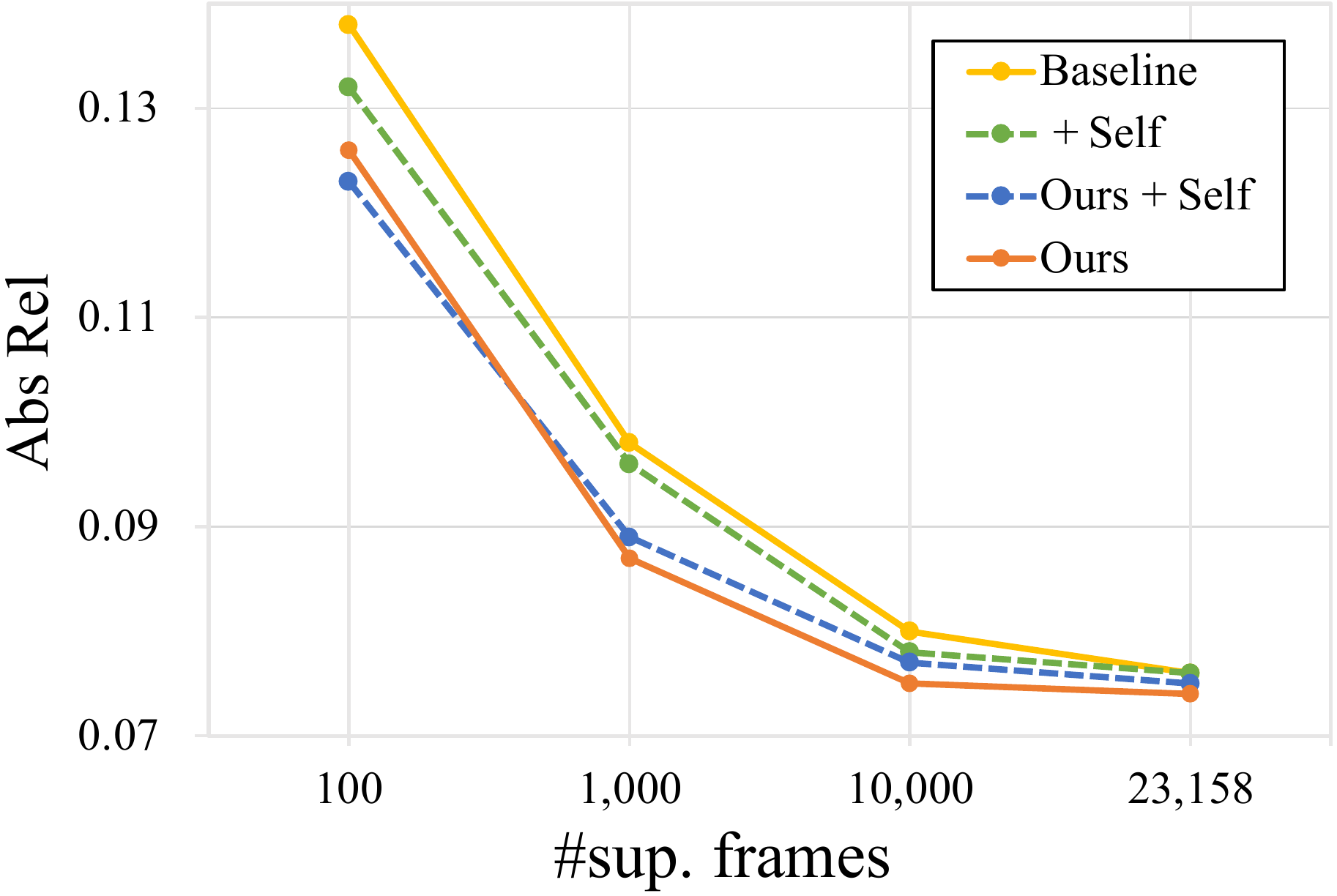}}
\subfigure[]
{\includegraphics[width=0.33\linewidth]{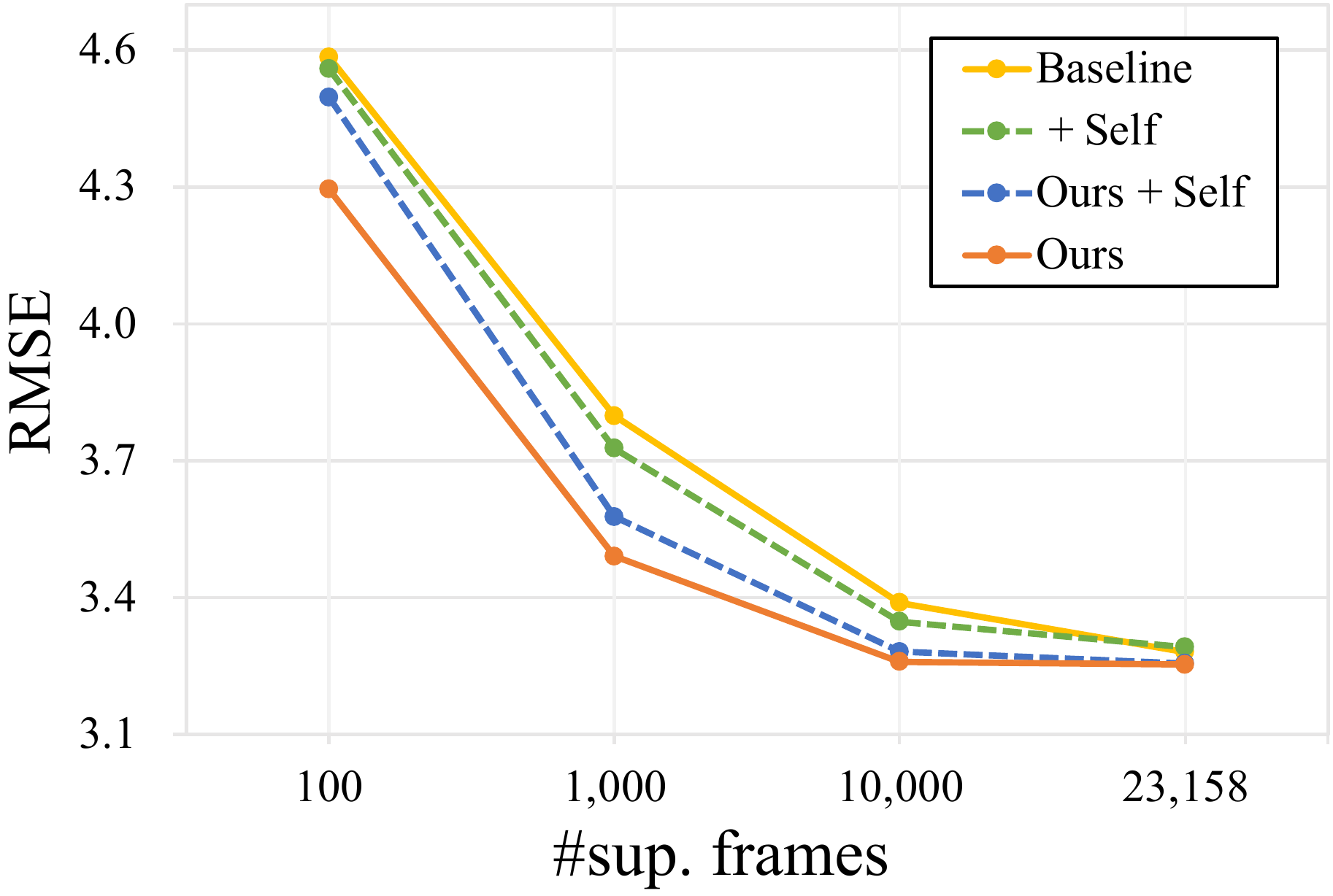}}
\subfigure[]
{\includegraphics[width=0.33\linewidth]{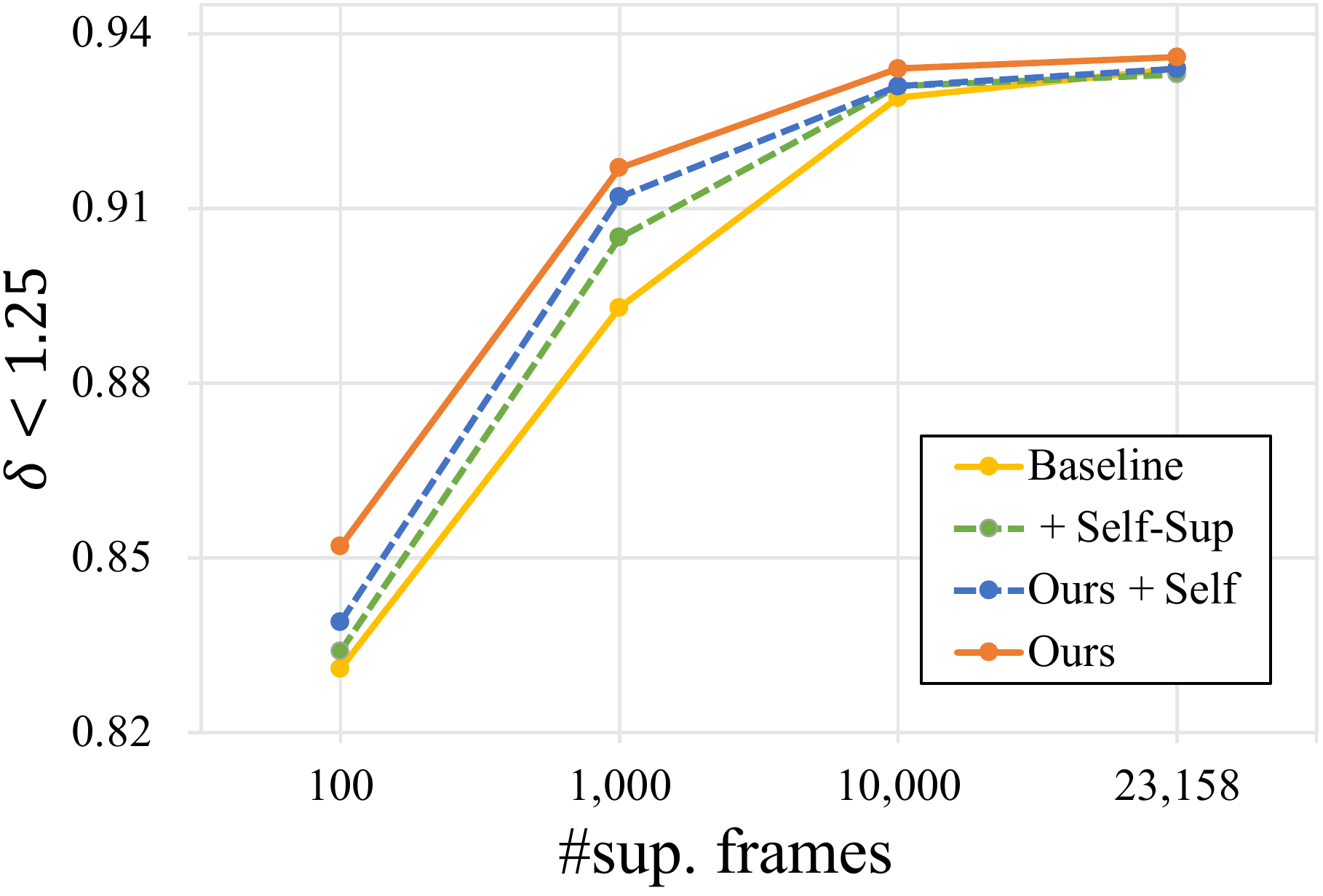}}
\end{center}
\vspace{-20pt}
% \vspace{-10pt}
\caption{\textbf{Quantitative results on the KITTI dataset in a sparsely-supervised setting.} ‘Baseline' only uses a sparse depth, and ‘Self’ indicates existing self-supervised strategies~\cite{garg2016unsupervised, godard2019digging}. ‘Ours’ indicates the proposed semi-supervised framework.}
\label{fig:graph}\vspace{-10pt}
\end{figure*}

\subsection{Na\"ive  Masking and Its Limitations}
Recent masked image modeling techniques for vision Transformers~\cite{bao2021beit,xie2021simmim, he2021masked} are effective as data augmentation, and MRA~\cite{xu2022masked} show promising potential.
The most na\"ive way to formulate masked image modeling as augmentation is to simply mask out the tokens. Specifically, given an image $I\in\mathbb{R}^{h\times w\times 3}$, we reshape it into a sequence of flattened non-overlapped 2D patches $X \in \mathbb{R}^{N\times P}$, where $h\times w$ is the resolution of the original image, $P = p\times p \times 3$ and $p\times p$ is the resolution of each image patch, and $N=hw/p^2$. 
These flattened 2D patches $X$ are embedded by a trainable linear projection~\cite{dosovitskiy2020image} operator, which proceeds to be fed into the Transformer encoder $f^\mathrm{enc}_\theta$. By applying the randomly sampled mask, the sequence of flattened 2D patches $X$ can be transformed into $X'$. Note that similar techniques were also used to increase the robustness of Transformers for image-level or pixel-level classification such as segmentation~\cite{xie2021simmim, he2021masked}.  

Meanwhile, monocular depth estimation heavily depends on contextual information when constructing depth maps. However, na\"ive  masking does not capture the overall context, 
impeding the ability to generate reliable depth predictions. As shown in Fig.~\ref{fig_sa}, applying a na\"ive masking strategy to monocular depth estimation causes scale ambiguity issue between results from weakly- and strongly-augmented branches during filling the missing regions, and the risk of missing out small-scale objects.

\subsection{$K$-way Disjoint Masking}
To address this, we present a $K$-way disjoint masking technique, where a $K$-disjoint set of tokens are encoded independently, then concatenated and decoded simultaneously, as illustrated in Fig.~\ref{fig:detail}. By capturing the entire scene from the partially divided inputs, our method can reduce inherent ambiguity~\cite{pathak2016context} and lead to coherent results, as shown in Fig.~\ref{fig_sa}. Moreover, since it enforces scale consistency by keeping the image size and orientation unaltered, the $K$-disjoint masking strategy can also act as data augmentation. 
Specifically, we divide the sequence of flattened 2D patches $X \in \mathbb{R}^{N\times P}$ into $K$ non-overlapping subsets $X_k$ for $k\in\{1,...,K\}$, with $X_k \in \mathbb{R}^{M\times P}$, where $M$ is set to be a random value smaller than $N$ to avoid learning with a fixed size of the tokens set. In other words, the concatenation of all the $X_k$ tokens should reconstruct the original token representation $X$, while maintaining the proper position ordering such that
\begin{equation}
    X = [X_1,X_2,...,X_K],
\end{equation}
where $[\cdot]$ denotes a concatenation operator. 
By independently encoding $X_k$ to the latent vector $\mathbf{z}_k$ such that $\mathbf{z}_k = f^\mathrm{enc}_\theta(X_k)$, unlike original Transformer-based encoding, i.e., $\mathbf{z} = f^\mathrm{enc}_\theta(X)$, the limited candidates are considered when running self-attention computation, which in turn implements an augmentation over tokens. 

Then, to decode all the $\mathbf{z}_k$, we reassemble them as $\bar{\mathbf{z}} = [\mathbf{z}_1,\mathbf{z}_2,...,\mathbf{z}_K]$ and obtain the final depth $D = f^\mathrm{dec}_\theta(\bar{\mathbf{z}})$. 

In our framework, we control the intensity of augmentations to the networks by adjusting $K$. In our experiments, we empirically set $K=1$ for the weak branch, and $K=64$ for the strong branch, which generate decoded depth maps $D_\mathrm{weak}$ and $D_\mathrm{strong}$, respectively.

\subsection{Loss Functions}
To train the networks we adopt a sparse supervised loss and the proposed unsupervised loss. In addition, we adopt a loss for modeling the uncertainty of the pseudo ground-truth produced by the weak branch and a feature consistency loss. $\|\cdot \|_{1}$ and $\|\cdot \|_{2}$ are $\ell_1$ and $\ell_2$ loss functions, respectively.

\begin{figure*}[t]
\begin{center}
\includegraphics[width=1.0\linewidth]{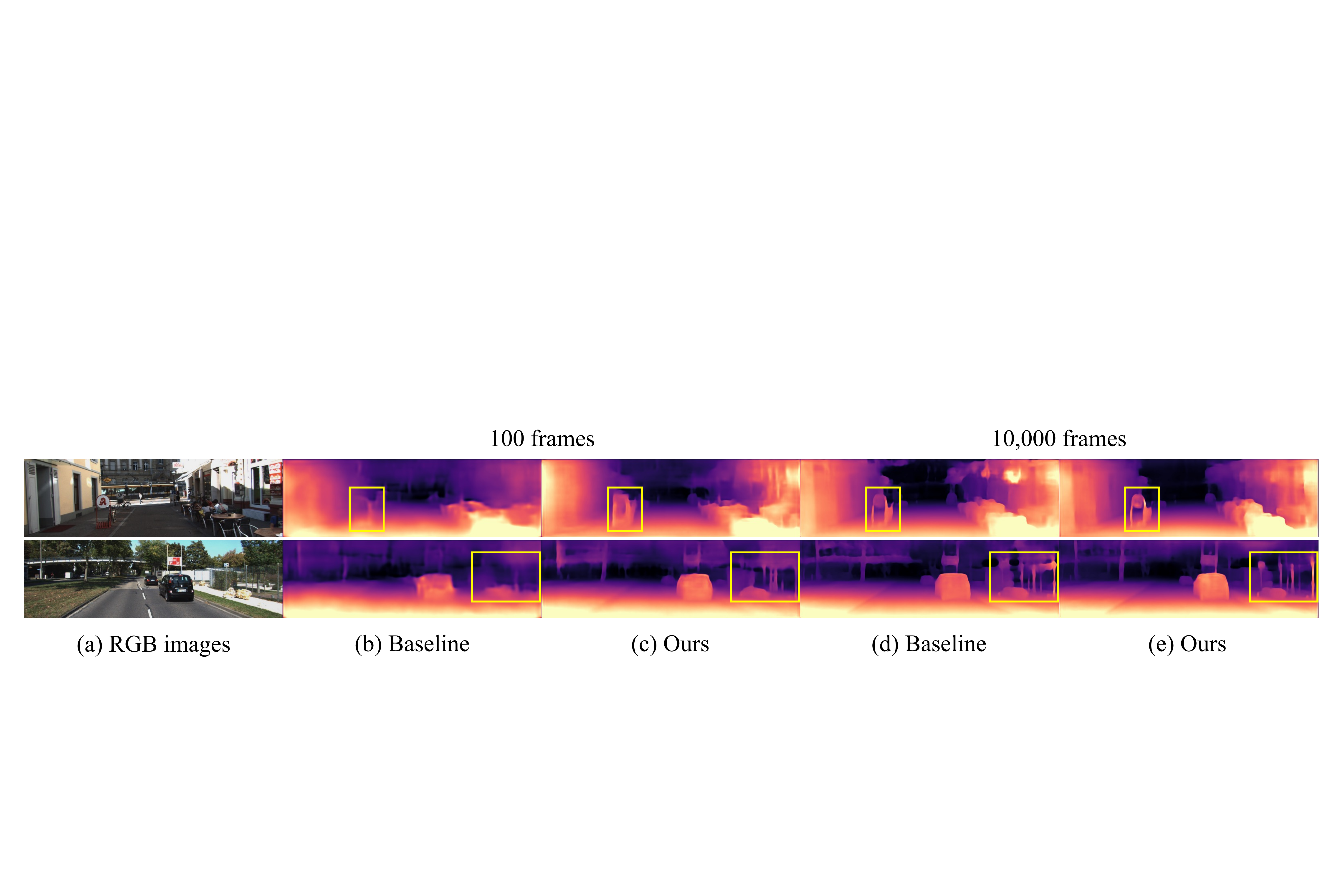}\hfill\\
\end{center}
% \vspace{-15pt}
\vspace{-10pt}
\caption{\textbf{Qualitative results on the KITTI dataset~\cite{geiger2012we}:} (a) RGB images, predicted depth maps by (b), (d) baseline, and (c), (e) ours using 100 and 10,000 labeled frames, respectively. }%
\label{fig6}
\vspace{-10pt}
\end{figure*}

\begin{figure*}[t]
\begin{center}
\includegraphics[width=1.0\linewidth]{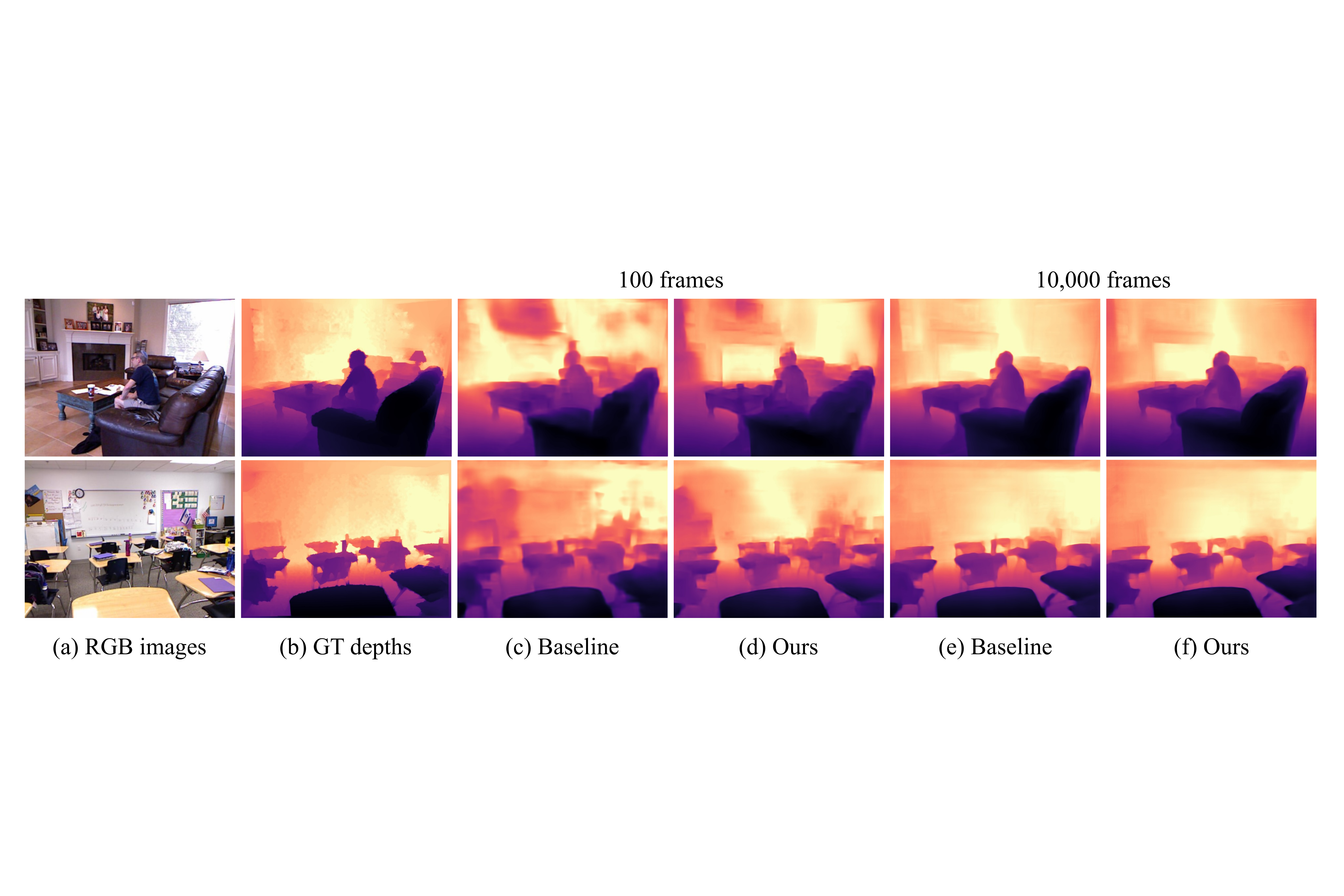}\hfill\\
\end{center}
% \vspace{-15pt}
\vspace{-10pt}
\caption{\textbf{Qualitative results on the NYU-Depth-v2 dataset~\cite{silberman2012indoor}:} (a) RGB images, (b) ground-truth depth maps, and predicted depth maps by (c), (e) baseline, and (d), (f) ours using 100 and 10,000 labeled frames, respectively.}%
\label{fig7}\vspace{-10pt}
\end{figure*}

\vspace{-10pt}
\paragraph{Sparse supervised loss.}
When sparsely depth-labeled data is available to train the network, we can minimize the supervised loss function $\mathcal{L}_{\mathrm{gt}}$ between the predicted $D$ and sparse ground-truth $D_{\mathrm{gt}}$ such that 
\begin{equation}
	\mathcal{L}_\mathrm{gt} = \|D- D_\mathrm{gt}\|_{1}.
	\label{eq:gtloss}
\end{equation}
A small number of fully annotated data used in a supervised manner across both branches allows the network model to ignite the learning process, which is then carried out mainly on unlabeled data through consistency regularization. 

 % both branches 
\vspace{-10pt}
\paragraph{Depth consistency loss.}
Our loss function encourages depth prediction of the strongly-augmented image to be close to the prediction of the weakly-augmented image, enabling pixel-level learning without the need for annotated ground-truths, thus it serving as an effective solution to data hunger caused by sparse annotations. The depth consistency loss $\mathcal{L}_\mathrm{dc}$ assisted by the uncertainty map $U(D_\mathrm{weak})$ can be written as:

\begin{equation}
     \mathcal{L}_\mathrm{dc} = \mathrm{sg}(U(D_\mathrm{weak})) \odot\|\mathrm{sg}(D_\mathrm{weak})- D_\mathrm{strong}\|_{1}.
\end{equation}
where $\odot$ denotes hadamard product. 

\vspace{-10pt}
\paragraph{Uncertainty loss.} 
To filter out the noise on pseudo labels, we train the uncertainty module. This module, which is a key ingredient in our framework, allows for transferring only reliable depth knowledge from weak branch to strongly augmented branch. To model such uncertainty, we leverage a negative log-likelihood minimization~\cite{kendall2017uncertainties} as:
\begin{equation}
    \mathcal{L}_\mathrm{uc} = \frac{\|D- D_\mathrm{gt}\|_{1}}{U(D)} + \mathrm{log}(U(D)),%\|D- D_{gt}\|/\sigma + \mathrm{log}(\sigma),
\end{equation}
where $U(D)$ denotes the uncertainty map related to the predicted depth map $D$. By training the network to model its uncertainty, predictions on unlabeled data will be trusted if highly confident, as shown in Fig.~\ref{fig4}.

\vspace{-10pt}
\paragraph{Feature consistency loss.}
Within our framework, geometric distortions are not applied to the two branches, thus the encoded feature consistency can also be encouraged. The feature consistency loss is then defined as
\begin{equation}
    \mathcal{L}_\mathrm{fc} = 
    \|\mathbf{z}_\mathrm{weak}- h(\mathbf{z}_\mathrm{strong})\|_{2},
\end{equation}
where $h(\cdot)$ is the additional MLP predictor head, which provides better results as shown in the literature~\cite{chen2021exploring,grill2020bootstrap} and prevents collapse~\cite{zhang2022does}.

\vspace{-10pt}
\paragraph{Total loss.}
By considering all the loss terms, the final loss is defined as $\mathcal{L} = \mathcal{L}_\mathrm{gt} + 
    \lambda_\mathrm{dc} \mathcal{L}_\mathrm{dc} + \lambda_\mathrm{uc} \mathcal{L}_\mathrm{uc} + \lambda_\mathrm{fc} \mathcal{L}_\mathrm{fc}$, where $\lambda_\mathrm{uc}$, $\lambda_\mathrm{dc}$, and $\lambda_\mathrm{fc}$ represent hyper-parameters.

% 4_Experiments.tex
\section{Experiments}
\label{sec:exp}
\subsection{Implementation Details}
We implemented our MaskingDepth with the Pytorch library~\cite{paszke2017automatic}. We conduct all our experiments on 24GB RTX-3090 GPUs, using DPT-Base~\cite{ranftl2021vision} as a backbone model. 
We set the learning rate to ${10^{-5}}$ for the encoder and ${10^{-4}}$ for the decoder. The encoder is initialized with ImageNet-pretrained~\cite{deng2009imagenet} weights, whereas the decoder is initialized randomly. We train the entire model with batch size 8 and use Adam optimizer with $\beta_{1} = 0.9$ and $\beta_{2} = 0.999$. 

\begin{table*}[t!]
	\centering
	\scalebox{1}{
		\small
		\setlength{\tabcolsep}{0.5em}
		\begin{tabular}{l|cccc|c|ccccc}
			\toprule
			
			Methods &
			Sup. &
			Self.(M) &
			Self.(S) &
			Self. &
			DB. &
			AbsRel$\downarrow$ &
			SqRel$\downarrow$ &
			RMSE$\downarrow$ &
			RMSElog$\downarrow$ &
			\textbf{$\delta$}$\uparrow$ \\
			\midrule

DORN~\cite{fu2018deep} & $\checkmark$ & - & - & - & K & 0.072 & 0.307 & 2.727 & 0.120 & 0.932 \\

Yin et al.~\cite{yin2019enforcing}
&  $\checkmark$ & - & - & - & K & 0.072 & - & 3.258 & 0.117 & 0.938 \\

DPT-Hybrid~\cite{ranftl2021vision}
&  $\checkmark$ & - & - & - & K+Mix & \textbf{0.062} & \textbf{0.222} & \textbf{2.575} & \textbf{0.092} & \textbf{0.959} \\

DPT-Base~\cite{ranftl2021vision}$^\ast$
&  $\checkmark$ & - & - & - & K & 0.074 & 0.361 & 3.275 & 0.117 & 0.935 \\
\midrule

Monodepth2~\cite{godard2019digging}
&  -  &  $\checkmark$ &  $\checkmark$ & - & K & 0.080 & 0.466 & 3.681 & 0.127 & 0.926 \\

PackNet-SfM~\cite{guizilini20203d}
&  -  &  $\checkmark$ & -  & - & K & 0.078 & 0.420 & 3.485 & 0.121 & 0.931 \\

ManyDepth~\cite{watson2021temporal}
&  -  &  $\checkmark$ & - & - & K & \textbf{0.070} & \textbf{0.399} & \textbf{3.455} & \textbf{0.113} & \textbf{0.941} \\

\midrule

 SVSM FT~\cite{mayer2016large}
&  $\checkmark$ & - &  $\checkmark$ & - & K+F & 0.077 & 0.392 & 3.569 & 0.127 & 0.919 \\

 Kuznietsov et al.~\cite{kuznietsov2017semi}
&    $\checkmark$  & - &  $\checkmark$ & - & K & 0.089 & 0.478 & 3.610 & 0.138 & 0.906 \\

 Baek et al.~\cite{baek2022semi} 
&  $\checkmark$   &  $\checkmark$ &  $\checkmark$ & - & K & 0.078 & 0.381 & 3.404 & 0.121 & 0.930 \\

 Guizilini et al.~\cite{guizilini2020robust}
&  $\checkmark$    &  $\checkmark$ & - & - & K & 0.072 & 0.340 & 3.265 & 0.116 & 0.934 \\

 SemiDepth~\cite{amiri2019semi}
&  $\checkmark$ & - &  $\checkmark$ & - & K+C & 0.078 & 0.417 & 3.464 & 0.126 & 0.923  \\

 \midrule
 \textbf{MaskingDepth (DPT-Base)}
&  $\checkmark$ & - & - &  $\checkmark$ & K+C & \textbf{0.071} & \textbf{0.316} & \textbf{3.049} & \textbf{0.111} & \textbf{0.941}  \\

\bottomrule
		\end{tabular}
	}\vspace{5pt}%\vspace{-5pt}
    \caption{\textbf{Quantitative results on the Eigen split of the KITTI dataset~
    \cite{geiger2012we}.} 
%comparing our method against the existing approaches on the Eigen split of the KITTI dataset~
%\cite{geiger2012we}, using improved ground truth. %Best results are in \textbf{bold}.
    `Sup.', `Self.(M)', and `Self.(S)' indicate supervised, existing self-supervised strategies on video and stereo pairs, respectively. `Self.' denotes our proposed consistency regularization, which needs\textbf{ no stereo images or video sequences}. `K', `C', and `F' indicate KITTI~\cite{geiger2012we}, Cityscapes~\cite{cordts2016cityscapes}, and FlyingThings3D~\cite{mayer2016large}, respectively. 'Mix' indicates the dataset proposed from~\cite{ranftl2021vision}, containing 1.4$M$ images, which is approximately 60 times larger than KITTI. `$\ast$' denotes our re-implementation of the model trained on the full dataset.}
    \label{tab:kitti_sparse}\vspace{-10pt}
\end{table*}

To avoid collapsing, we balance the ratio of labeled and unlabeled data in one batch to 1:7 similarly to~\cite{sohn2020fixmatch}. Besides our new data augmentation approach, we adopt flipping and jittering, widely used in the literature~\cite{dwibedi2017cut,ghiasi2021simple}. For confidence estimation, we train the network to predict the log variance because it is more numerically stable~\cite{kendall2017uncertainties} than regressing the variance itself, as the loss avoids any division by zero. We use an identical hyperparameter set (i.e., $\lambda_\mathrm{uc}=1$, $\lambda_\mathrm{dc}=1$, $\lambda_\mathrm{fc}=1$, $K=64$ for strong augmentation) for all experiments unless otherwise specified.

\subsection{Experimental Settings}
\paragraph{Dataset.}
We first evaluate the performance of MaskingDepth and others on the KITTI dataset~\cite{geiger2012we} and NYU-Depth-v2~\cite{silberman2012indoor}. The KITTI dataset~\cite{geiger2012we} provides outdoor scenes captured by 3D laser data. The RGB images are resized to $640\times192$ resolutions for training. We follow the standard Eigen training/testing split~\cite{eigen2014depth}. We use randomly sampled 10,000, 1,000, and 100 images from $24K$ (i.e., left frames in the Eigen training split) for labeled images during training. We evaluate our trained model on 652 annotated test images for single-view depth estimation, using the improved ground truth by~\cite{uhrig2017sparsity}. The NYU-Depth-v2 dataset~\cite{silberman2012indoor} is composed of various indoor scenes and corresponding depth maps at a resolution of $640\times480$. We train our network on the same number of labeled images as KITTI, randomly sampled from the original 40$K$ total. The remaining images in the training set are used as unlabeled images. We test our trained model on 654 test images.

\begin{table}[t!]
  	\centering
	{
		\small
        \resizebox{1\linewidth}{!}{
		\begin{tabular}{l|c|c|ccc}
			\toprule
			Methods &
			DB. &
			AbsRel$\downarrow$ &
			RMSE$\downarrow$ &
			\textbf{$\delta$}$\uparrow$ \\
			\midrule

DORN~\cite{fu2018deep}
&  N  & 0.115 & 0.509 & 0.828 \\

BTS~\cite{lee2019big}
&  N  & 0.110 & 0.392 & 0.885 \\

DPT-Hybrid~\cite{ranftl2021vision}
& N+Mix & 0.110 & \textbf{0.357} & \textbf{0.904} \\

DPT-Base$^\ast$~\cite{ranftl2021vision}
& N  & 0.106 & 0.380 & 0.899 \\

\midrule
\textbf{MaskingDepth (DPT-Base)}
& N+S  & \textbf{0.104} & 0.372 & \textbf{0.904} \\

\bottomrule
		\end{tabular}
	}
    }\vspace{2pt}%\vspace{-5pt}
    \caption{\textbf{Quantitative results on the NYU-Depth-v2~\cite{silberman2012indoor}. } `N' and `S' indicate NYU-Depth-v2 and SUN RGB-D~\cite{song2015sun} datasets, respectively.}
      \label{tab:nyu_qualitative}\vspace{-10pt}
\end{table}

\vspace{-10pt}
\paragraph{Evaluation metrics.}
In our experiments, we follow the standard evaluation protocol of the prior work~\cite{eigen2014depth} to evaluate the effectiveness of MaskingDepth. The error metrics are defined as Absolute Relative error (AbsRel), Squared Relative error (SqRel), Root Mean Squared Error (RMSE), Root Mean Squared log Error (RMSElog), and accuracy under the threshold ($<1.25$) ($\delta$). 

\subsection{Depth Estimation Results}
In this section, we investigate the effects of sparse labels on supervised depth training, and how MaskingDepth is able to mitigate the degradation of depth maps when the number of labels is significantly limited. Especially, our method is useful for improving performance by utilizing vast unlabeled data in place of expensive ground-truth depth annotations.

\vspace{-10pt}
\paragraph{Robustness of MaskingDepth.} As a baseline, we compare DPT-Base~\cite{ranftl2021vision} trained on supervised and conventional semi-supervised manners obtained with self-supervised losses~\cite{garg2016unsupervised, godard2019digging}. Results for the KITTI dataset~\cite{geiger2012we} using different numbers of supervised frames are shown in Fig.~\ref{fig:graph}. MaskingDepth outperforms the baseline with any arbitrary number of data frames. As the amount is further decreased, the performance of all approaches, except for ours, shows a significant and rapid decline. This is mostly due to the model's inability to learn the appropriate scale and structure of a scene with such sparse information. However, as the labels become sparser, the performance degradation of our proposed method progresses more slowly compared to the baseline or naive semi-supervised approach using self-supervised losses, and the performance gap gets larger. Note that when MaskingDepth is incorporated with the existing self-supervised approach, the performance gain was marginal because the existing solution cannot avoid increasing inherent scale ambiguity from the self-supervised loss function. We also provide a qualitative comparison of the baseline and our method on the KITTI dataset in Fig.~\ref{fig6} and the NYU-Depth-v2 dataset in Fig.~\ref{fig7}. 

\begin{table}[t]
\centering
\resizebox{1.0\linewidth}{!}{%
\begin{tabular}{l|ccc|ccc}
        \toprule
        \multirow{2}{*}{Method}  & \multicolumn{3}{c}{Sup} & \multicolumn{3}{c}{\textbf{Semi-Sup}} \\
        & AbsRel $\downarrow$ & RMSE $\downarrow$ &\textbf{$\delta$}$\uparrow$ & AbsRel $\downarrow$ & RMSE $\downarrow$ &\textbf{$\delta$}$\uparrow$  \\
        \midrule
        Baseline &0.078 &3.370 &0.930 & - & - & -\\
        \midrule
        CutOut~\cite{devries2017improved} &0.076 &3.302 & 0.932 &0.075 & 3.351 &0.931\\
        SimMiM~\cite{xie2021simmim} &0.077 & 3.338 &0.931 &0.076 & 3.363 &0.931\\
        MAE~\cite{he2021masked} &\textbf{0.075} & 3.291 &0.933 &\textbf{0.074} & 3.280  &0.934\\
        \midrule
        \textbf{MaskingDepth} &0.076 & \textbf{3.289} &\textbf{0.934} &0.075 & \textbf{3.239} &\textbf{0.937}\\
        \bottomrule
\end{tabular}%
}\vspace{5pt}%\vspace{-5pt}
\caption{\textbf{Impact of data augmentation.}} 
\label{tab:augmentation}\vspace{-10pt}
\end{table}

\vspace{-10pt}
\paragraph{Comparison to other methods.}
 As our method does not rely on stereo or video sequence frames, it is agnostic to the configuration of the unlabeled training set. 
Table~\ref{tab:kitti_sparse} compares our semi-supervised method that uses additional data against existing approaches. We trained our model on the KITTI dataset~\cite{geiger2012we} as labeled data and the additional Cityscape dataset~\cite{cordts2016cityscapes} as additional  unlabeled data. In the results, our method achieves significant improvement in comparison to the baseline by utilizing unlabeled data and surpasses the state-of-the-art in semi-supervised depth estimation methods. 
Moreover, despite using a smaller model capacity and a fewer annotated data, MaskingDepth shows competitive performance against DPT-Hybrid. A similar trend can be seen in Table~\ref{tab:nyu_qualitative}, where our method utilizes the SUN RGB-D~\cite{song2015sun} dataset as additional unlabeled data and NYU-Depth-v2 as labeled data.
\begin{table}[t]
\centering
\resizebox{1.0\linewidth}{!}{
\begin{tabular}{l|ccc|ccc} 
\toprule
Methods & D & U & F & AbsRel $\downarrow$ & RMSE $\downarrow$ & \textbf{$\delta$}$\uparrow$   \\ 
\midrule
Baseline    & -             & -             & -             & 0.136 & 4.585 & 0.833 \\ 
\midrule
\multirow{4}{*}{MaskingDepth}   & $\checkmark$  & -             & -             & 0.132 & 4.355  &0.848   \\
                                &$\checkmark$  & $\checkmark$  &  -            &  0.126 & 4.296 &0.851\\
                                &-              & -& $\checkmark$  & 0.131 &  4.422  & 0.849 \\ 
                                &$\checkmark$   & $\checkmark$  & $\checkmark$  & \textbf{0.124} & \textbf{4.263} & \textbf{0.855}  \\ 
\bottomrule
\end{tabular}
}
% \vspace{-5pt}
\vspace{2pt}
\caption{\textbf{Ablation study on main components:} Depth consistency (D), Uncertainty (U), and Feature consistency (F).}\label{tab:ablation}\vspace{-5pt}
\end{table}

\begin{table}[t]
    \centering
    \resizebox{1.0
    \linewidth}{!}{
	{
		\setlength{\tabcolsep}{0.3em}
		\begin{tabular}{l|c|ccccc}
			\toprule
			Methods &
                % Dataset &
                Cap &
			AbsRel$\downarrow$ &
			SqRel$\downarrow$ &
			RMSE$\downarrow$ &
			RMSElog$\downarrow$ &
			\textbf{$\delta$}$\uparrow$ \\
			\midrule

 Kundu \textit{et al.}~\cite{kundu2018adadepth} 
& 50m & 0.203 & 1.734 & 6.251 & 0.284 & 0.687 \\

 T2Net~\cite{zheng2018t2net}
& 50m & 0.168 & 1.199 & 4.674 & 0.243 & 0.772 \\

 GASDA~\cite{zhao2019geometry}
& 50m & 0.143 & 0.756 & 3.846 & 0.217 & 0.836 \\

 SharinGAN~\cite{pnvr2020sharingan}
& 50m & \textbf{0.109} & \textbf{0.673} & 3.770 & 0.190 & \textbf{0.864} \\

 \textbf{MaskingDepth}
& 50m & 0.129 & 0.719 & \textbf{3.475} & \textbf{0.180} & 0.845\\

\midrule
 GASDA~\cite{zhao2019geometry} 
& 80m & 0.149 & 1.003 & 4.995 & 0.227 & 0.824 \\

 SharinGAN~\cite{pnvr2020sharingan} 
& 80m & \textbf{0.116} & \textbf{0.939} & 5.068 & 0.203 & \textbf{0.850} \\

 \textbf{MaskingDepth}
& 80m & 0.136 & 1.020 & \textbf{4.726} & \textbf{0.190} & 0.833 \\

\bottomrule
		\end{tabular}
	}
 }
 \vspace{2pt}
% \vspace{-5pt}
  \caption{\textbf{Domain adaptation on the KITTI dataset~\cite{geiger2012we}}. For the training data, KITTI dataset~\cite{geiger2012we} and vKITTI dataset~\cite{gaidon2016virtual} were used. }
	\label{tab:domain}
	\vspace{-5pt}
\end{table}

\subsection{Ablation Study}
We analyze the effectiveness of different design choices in our framework on the KITTI dataset~\cite{geiger2012we}. We exploit 10,000 and 100 randomly sampled images, respectively for data augmentation and other ablation studies.
\vspace{-10pt}

\paragraph{Data augmentation.}
We evaluate our proposed $K$-way disjoint masking as augmentation and compare it to different MIM methods, including SimMIM~\cite{xie2021simmim} and MAE~\cite{he2021masked}, and another data augmentation method, CutOut~\cite{devries2017improved}. We evaluate these methods in both fully supervised and semi-supervised settings using a baseline model, which is trained in a supervised manner and without any augmentation.
The results are summarized in Table~\ref{tab:augmentation}. Although other augmentation methods improved the model performance, they also increased the inherent ambiguity. Our augmentation not only reduces the inherent ambiguity but also achieves the best performance by capturing the entire scene.
\vspace{-10pt}

\paragraph{Loss functions.} We examine each component of the loss function in our method. It consists of four components: the sparse supervised term, the depth consistency term, the feature consistency term, and the uncertainty term. To evaluate the impact of each component, we start by using only the sparse supervised loss as a baseline and then study the effect of adding each of the other three components. As shown in Table~\ref{tab:ablation}, the performance of our network improves as each component of the loss function is added. We can see that using all components together leads to a significant improvement.
\vspace{-10pt}

\paragraph{The number of ${K}$.}
To study the influence of the masking ratio, we train the network by adopting different values of $K$ for the strong branch, respectively $K=4, 16, 64$, and $128$. The quantitative evaluation results for this study is shown in Supplementary materials. Starting from $K=4$, the error decreases with the increase of $K$, until degrading for $K=128$. We set $K=64$ as the default since it yields the best results.

\begin{figure}[t]
% \captionsetup[subfigure]{labelformat=empty}
\begin{center}
\renewcommand{\thesubfigure}{}
\subfigure[]
{\includegraphics[width=0.49\linewidth]{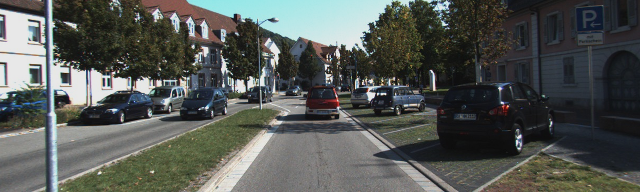}}
\hspace{-2pt}
\subfigure[]
{\includegraphics[width=0.49\linewidth]{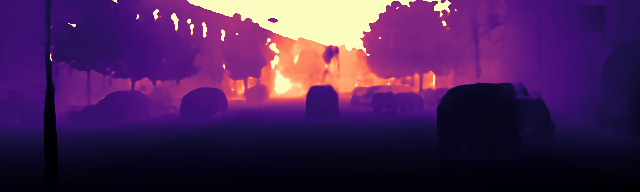}}
\hfill\\
\vspace{-22pt}
\subfigure[]
{\includegraphics[width=0.49\linewidth]{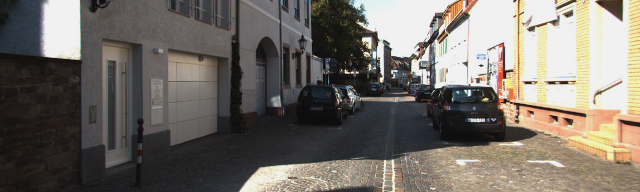}}
\hspace{-2pt}
\subfigure[]
{\includegraphics[width=0.49\linewidth]{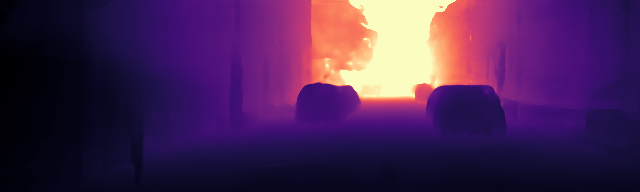}}\hfill\\
\end{center}
\vspace{-20pt}
% \vspace{-10pt}
\caption{\textbf{Qualitative results of KITTI~\cite{geiger2012we} data samples for models trained with vKITTI datasets~\cite{gaidon2016virtual}:} (a) RGB,  (b) predicted depth maps, MaskingDepth proves to work well in domain adaptation task on real-world images.}%
\label{fig8}\vspace{-10pt}
\end{figure}

\subsection{Domain Adaptation}
Inspired by~\cite{melas2021pixmatch}, we simply apply MaskingDepth to the task of unsupervised domain adaptation using virtual KITTI (vKITTI)~\cite{gaidon2016virtual} and KITTI~\cite{geiger2012we} as synthetic and real datasets, respectively.
As demonstrated in Table~\ref{tab:domain}, MaskingDepth achieves reasonable performance compared to others. While other methods rely on generative models which are large and hard to train, our method is lightweight and easily adaptable without an external network. 
Fig.~\ref{fig8} shows some well-adapted results in the real-world domain.

\section{Conclusion}
\label{Conclusion}
In this paper, we presented MaskingDepth, a novel semi-supervised framework for monocular depth estimation using consistency regularization. MaskingDepth allows leveraging unlabeled data without requiring either stereo or sequential frames. To this end, we also proposed a data augmentation method, called $K$-way disjoint masking. We have shown that $K$-way disjoint masking makes consistency regularization framework more stable by utilizing an uncertainty network. Our proposed method showed significant improvement on extremely sparse labeled data and superior results compared to other semi-supervised approaches. For future work, we plan to extend our method to other dense prediction tasks.

\paragraph{Acknowledgements.}
\label{Acknowledgements}
This research was supported in part by Autonomous Driving Center, R\&D Division, Hyundai Motor Company.

\appendix
\onecolumn
\null
   \begin{center}
   
    {\LARGE \bf Appendix\par}
    \vspace*{48pt}

   \end{center}

% \maketitle
In this appendix, we provide more results on experiments and more implementation details of our framework.
    
\vspace{12pt}

\section*{Appendix A. Quantitative Details}
More detailed quantitative results of our proposed method and conventional self-supervised method on the KITTI dataset~\cite{geiger2012we} are shown in Table~\ref{tab:s2}. Moreover, to verify the generalization ability of our framework, we evaluated our framework on the NYU-Depth-v2 dataset~\cite{silberman2012indoor}. Results are shown in Table~\ref{tab:NYU_sparse}.
 
\begin{table*}[h]
	\centering
	\resizebox{0.95\linewidth}{!}{
	{
		\small
		\setlength{\tabcolsep}{0.2em}
		\begin{tabular}{c|c|ccccc}
		    \toprule
			Methods &
			\# sup. frames & 
            AbsRel $\downarrow$ &
			SqRel $\downarrow$&
			RMSE $\downarrow$&
			RMSElog $\downarrow$&
			\textbf{$\delta$}$\uparrow$
			\vspace{0.5mm}\\
			\toprule
			
		\multirow{5}{*}{Baseline}
		& 23,158
		& 0.076 $\pm$ 0.003
        & 0.365 $\pm$ 0.004
        & 3.290 $\pm$ 0.015
        & 0.118 $\pm$ 0.001
        & 0.934 $\pm$ 0.001
        \\
        & 10,000  
        & 0.079 $\pm$ 0.001
        & 0.379 $\pm$ 0.007
        & 3.388 $\pm$ 0.019
        & 0.121 $\pm$ 0.009
        & 0.929 $\pm$ 0.001
        \\
        & 1,000  
        & 0.098 $\pm$ 0.004
        & 0.515 $\pm$ 0.030
        & 3.785 $\pm$ 0.013
        & 0.142 $\pm$ 0.005
        & 0.899 $\pm$ 0.005 
        \\
        & 100  
        & 0.135 $\pm$ 0.005
        & 0.728 $\pm$ 0.019
        & 4.585 $\pm$ 0.048
        & 0.186 $\pm$ 0.011
        & 0.831 $\pm$ 0.005
        \\
        & 10  
        & 0.201 $\pm$ 0.023
        & 1.508 $\pm$ 0.045
        & 6.163 $\pm$ 0.082
        & 0.268 $\pm$ 0.029
        & 0.701 $\pm$ 0.021
        \\
        \cmidrule{1-7} 
        \multirow{5}{*}{Baseline + Self. (M)}
        & 23,158
		& 0.076 $\pm$ 0.002
        & 0.367 $\pm$ 0.007
        & 3.291 $\pm$ 0.020
        & 0.117 $\pm$ 0.001
        & 0.933 $\pm$ 0.002
        \\
        & 10,000  
        & 0.078 $\pm$ 0.001
        & 0.376 $\pm$ 0.006
        & 3.347 $\pm$ 0.043
        & 0.119 $\pm$ 0.002
        & 0.931 $\pm$ 0.001
        \\
        & 1,000  
        & 0.096 $\pm$ 0.002
        & 0.523 $\pm$ 0.024
        & 3.750 $\pm$ 0.033
        & 0.140 $\pm$ 0.002
        & 0.900 $\pm$ 0.004
        \\
        & 100  
        & 0.132 $\pm$ 0.004
        & 0.759 $\pm$ 0.014
        & 4.559 $\pm$ 0.044
        & 0.184 $\pm$ 0.003
        & 0.834 $\pm$ 0.004
        \\
        & 10  
        & 0.210 $\pm$ 0.020
        & 1.322 $\pm$ 0.042
        & \textbf{5.627 $\pm$ 0.080}
        & 0.265 $\pm$ 0.027
        & 0.711 $\pm$ 0.016
        \\
        
        \cmidrule{1-7} 
        \multirow{5}{*}{\textbf{MaskingDepth + Self. (M)}}
        & 23,158
		& 0.079 $\pm$ 0.001
        & 0.379 $\pm$ 0.007
        & 3.388 $\pm$ 0.019
        & 0.121 $\pm$ 0.009
        & 0.929 $\pm$ 0.001
        \\
        & 10,000  
        & 0.076 $\pm$ 0.017
        & 0.369 $\pm$ 0.004
        & 3.311 $\pm$ 0.011
        & 0.117 $\pm$ 0.001
        & \textbf{0.935 $\pm$ 0.002}
        \\
        & 1000  
        & \textbf{0.085 $\pm$ 0.017}
        & 0.430 $\pm$ 0.011
        & 3.521 $\pm$ 0.012
        & 0.129 $\pm$ 0.012
        & \textbf{0.918 $\pm$ 0.010}
        \\
        & 100  
        & \textbf{0.123 $\pm$ 0.003}
        & 0.747 $\pm$ 0.018
        & 4.497 $\pm$ 0.042
        & 0.181 $\pm$ 0.005
        & 0.839 $\pm$ 0.005
        \\
        & 10  
        & \textbf{0.184 $\pm$ 0.011}
        & \textbf{1.265 $\pm$ 0.064}
        & 5.747 $\pm$ 0.080
        & \textbf{0.243 $\pm$ 0.007}
        & 0.727 $\pm$ 0.018
        \\
        
        \cmidrule{1-7} 
        \multirow{5}{*}{\textbf{MaskingDepth}}
        & 23,158
		& \textbf{0.074 $\pm$ 0.001}
        & \textbf{0.362 $\pm$ 0.001}
        & \textbf{3.253 $\pm$ 0.012}
        & \textbf{0.116 $\pm$ 0.001}
        & \textbf{0.935 $\pm$ 0.001}
        \\
        & 10,000  
        & \textbf{0.075 $\pm$ 0.002}
        & \textbf{0.362 $\pm$ 0.006}
        & \textbf{3.259 $\pm$ 0.020}
        & \textbf{0.116 $\pm$ 0.001}
        & 0.934 $\pm$ 0.003
        \\
        & 1,000  
        & 0.088 $\pm$ 0.003
        & \textbf{0.419 $\pm$ 0.007}
        & \textbf{3.490 $\pm$ 0.020}
        & \textbf{0.129 $\pm$ 0.003}
        & 0.917 $\pm$ 0.002
        \\
        & 100  
        & 0.128 $\pm$ 0.004
        & \textbf{0.707 $\pm$ 0.013}
        & \textbf{4.295 $\pm$ 0.037}
        & \textbf{0.173 $\pm$ 0.006}
        & \textbf{0.849 $\pm$ 0.006}
        \\
        & 10  
        & 0.197 $\pm$ 0.019
        & 1.378 $\pm$ 0.032
        & 5.650 $\pm$ 0.091
        & 0.261 $\pm$ 0.030
        & \textbf{0.723 $\pm$ 0.017}
        \\

    \bottomrule
    \end{tabular}
	}}\vspace{8pt}
	\caption{\textbf{Quantitative results} on the KITTI dataset~\cite{geiger2012we} in a sparsely-supervised setting using sparse labels from~\cite{uhrig2017sparsity}. For each row, we trained 5 independent models with randomly selected labels from entire dataset to calculate the mean and variance. `Self. (M)' and 'MaskingDepth' indicate monocular self-supervised learning and proposed consistency regularization, respectively. The best results are in \textbf{bold}.
    }
	\label{tab:s2}
\end{table*}

\clearpage

\begin{table*}[h]
	\centering
	\resizebox{0.75\linewidth}{!}{
	{
		\small
		\setlength{\tabcolsep}{0.3em}
		\begin{tabular}{c|c|cccc}
		    \toprule
			Methods &
			\# sup. frames & 
			AbsRel $\downarrow$&
			RMSE  $\downarrow$&
			log$_{10}$ $\downarrow$&
            % $\delta < 1.25$
            \textbf{$\delta$}$\uparrow$
			\vspace{0.5mm}\\
			\toprule
			
		\multirow{5}{*}{Baseline}
		& 42,602  
        & 0.106 $\pm$ 0.002
        & 0.380 $\pm$ 0.004
        & \textbf{0.053 $\pm$ 0.001}
        & 0.897 $\pm$ 0.001
        \\
        & 10,000  
        & 0.112 $\pm$ 0.004
        & 0.389 $\pm$ 0.006
        & 0.057 $\pm$ 0.003
        & 0.893 $\pm$ 0.003
        \\
        & 1,000  
        & 0.141 $\pm$ 0.008
        & 0.447 $\pm$ 0.009
        & 0.066 $\pm$ 0.004
        & 0.843 $\pm$ 0.006
        \\
        & 100  
        & 0.199 $\pm$ 0.011
        & 0.604 $\pm$ 0.014
        & 0.086 $\pm$ 0.005
        & 0.694 $\pm$ 0.011
        \\
        & 10  
        & 0.321 $\pm$ 0.040
        & 0.872 $\pm$ 0.042
        & 0.124 $\pm$ 0.008
        & 0.523 $\pm$ 0.027
        \\
        \cmidrule{1-6} 
        \multirow{5}{*}{\textbf{MaskingDepth}}
        & 42,602  
        & \textbf{0.105 $\pm$ 0.002}
        & \textbf{0.379 $\pm$ 0.003}
        & \textbf{0.053 $\pm$ 0.001}
        & \textbf{0.899 $\pm$ 0.001}
        \\
        & 10,000  
        & \textbf{0.107 $\pm$ 0.002}
        & \textbf{0.386 $\pm$ 0.006}
        & \textbf{0.054 $\pm$ 0.002}
        & \textbf{0.896 $\pm$ 0.002}
        \\
        & 1,000  
        & \textbf{0.135 $\pm$ 0.007}
        & \textbf{0.440 $\pm$ 0.008}
        & \textbf{0.065 $\pm$ 0.004}
        & \textbf{0.853 $\pm$ 0.005}
        \\
        & 100  
        & \textbf{0.182 $\pm$ 0.008}
        & \textbf{0.594 $\pm$ 0.012}
        & \textbf{0.083 $\pm$ 0.003}
        & \textbf{0.718 $\pm$ 0.010}
        \\
        & 10  
        & \textbf{0.292 $\pm$ 0.031}
        & \textbf{0.814 $\pm$ 0.037}
        & \textbf{0.112 $\pm$ 0.006}
        & \textbf{0.561 $\pm$ 0.021}
        \\
    \bottomrule
    \end{tabular}
	}}\vspace{8pt}
    \caption{\textbf{Quantitative results} on the NYU-Depth-v2 dataset~\cite{silberman2012indoor} in a sparsely-supervised setting. For each row, we trained 5 independent models with randomly selected labels from the entire dataset to calculate the mean and variance.}
    \label{tab:NYU_sparse}
\end{table*}

\vspace{-10 pt}

\section*{Appendix B. Additional Results}
\begin{figure*}[h]
\centering
\renewcommand{\thesubfigure}{}

\subfigure[]
{\includegraphics[width=0.131\textwidth]{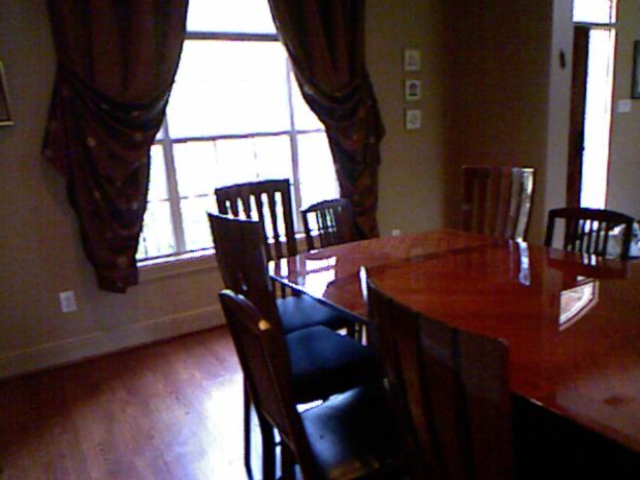}} 
\subfigure[]
{\includegraphics[width=0.131\textwidth]{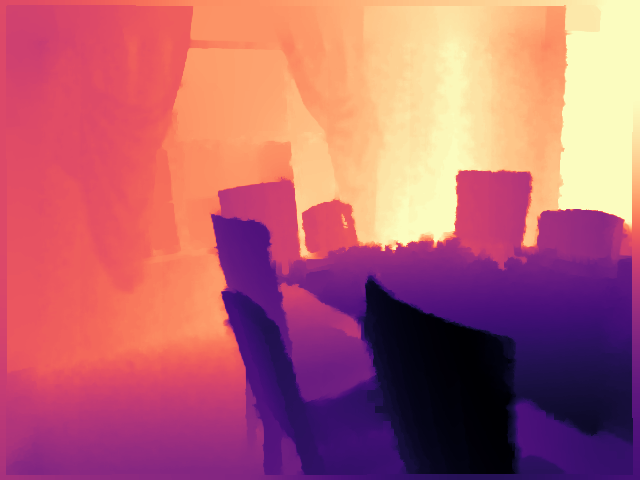}} 
\subfigure[]
{\includegraphics[width=0.131\textwidth]{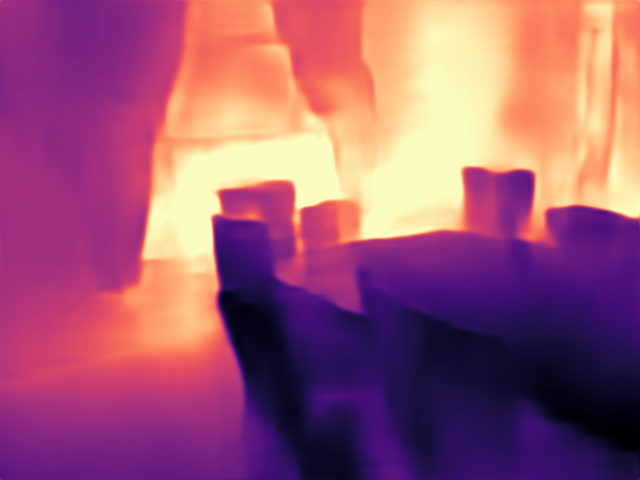}} 
\subfigure[]
{\includegraphics[width=0.131\textwidth]{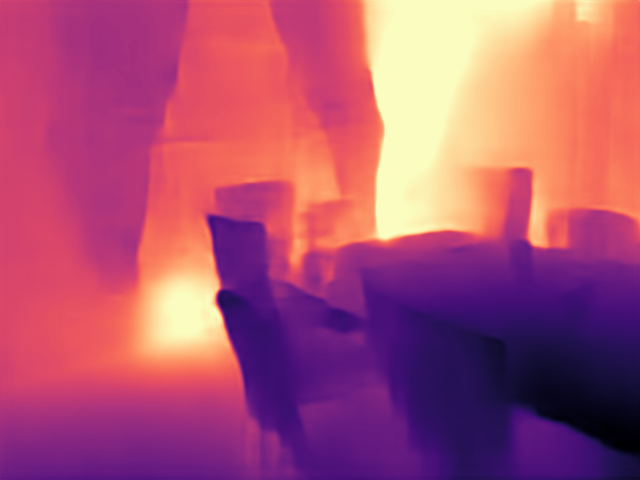}} 
\subfigure[]
{\includegraphics[width=0.131\textwidth]{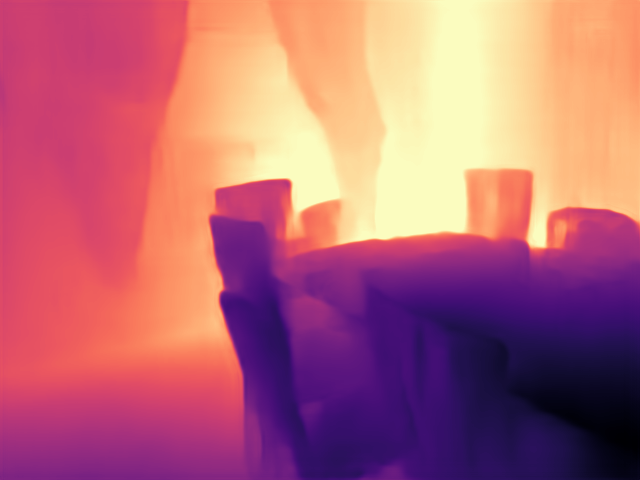}} 
\subfigure[]
{\includegraphics[width=0.131\textwidth]{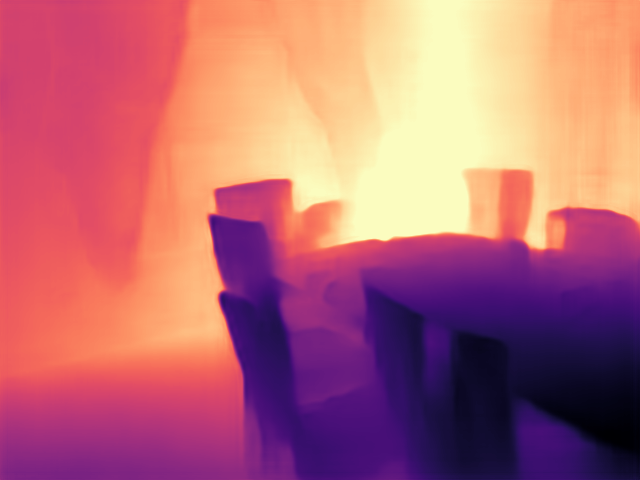}} \\\vspace{-20.5pt}

\subfigure[]
{\includegraphics[width=0.131\textwidth]{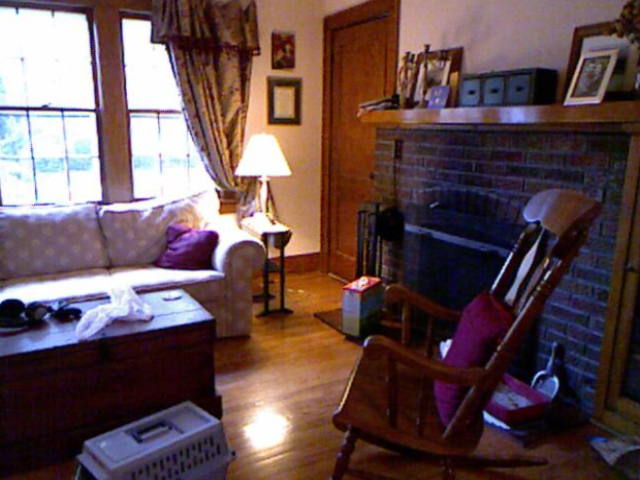}} 
\subfigure[]
{\includegraphics[width=0.131\textwidth]{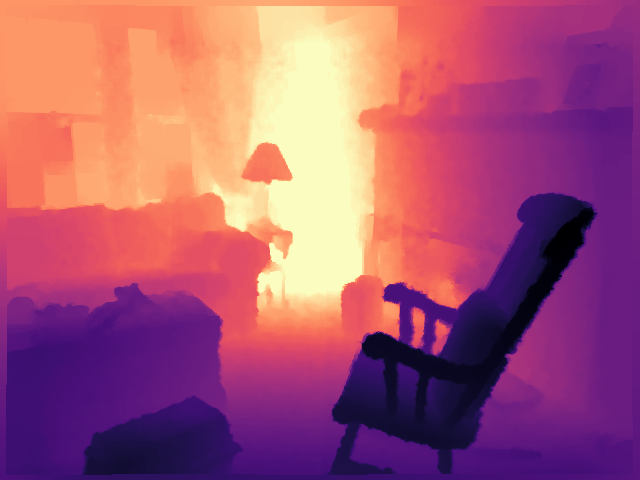}} 
\subfigure[]
{\includegraphics[width=0.131\textwidth]{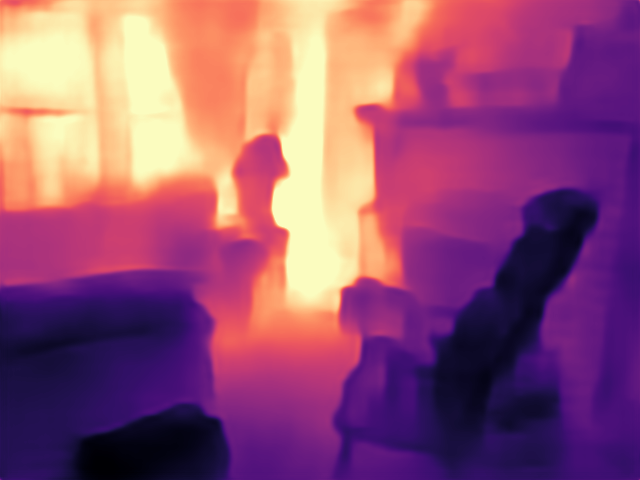}} 
\subfigure[]
{\includegraphics[width=0.131\textwidth]{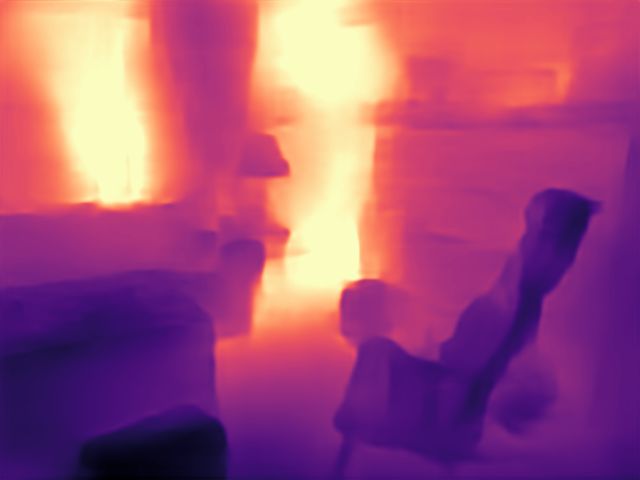}} 
\subfigure[]
{\includegraphics[width=0.131\textwidth]{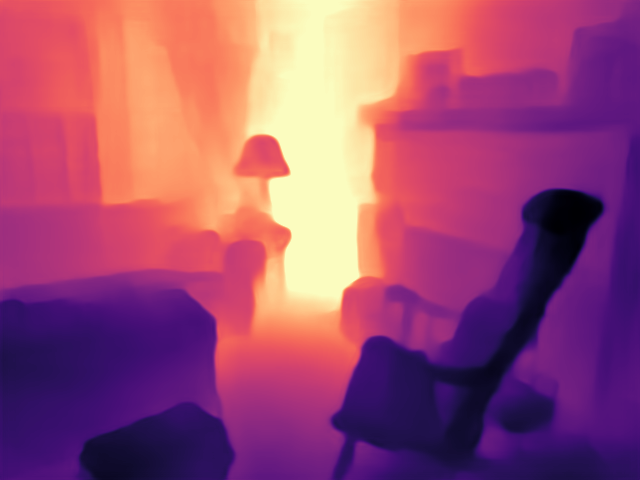}} 
\subfigure[]
{\includegraphics[width=0.131\textwidth]{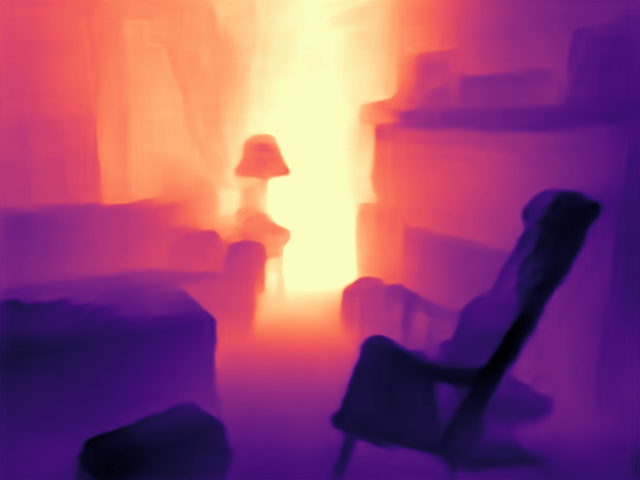}} \\\vspace{-20.5pt}

\subfigure[]
{\includegraphics[width=0.131\textwidth]{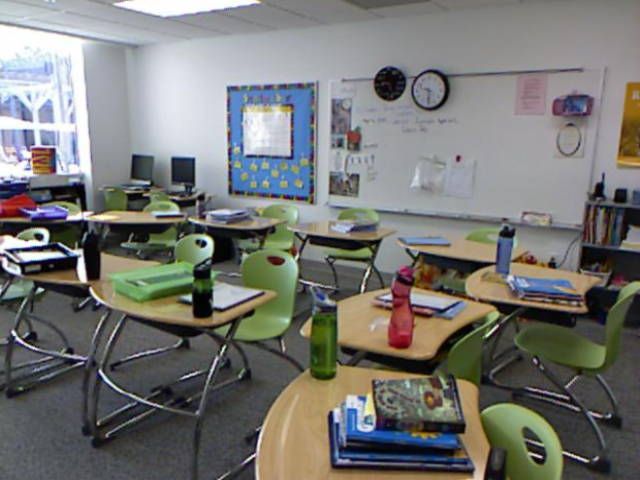}} 
\subfigure[]
{\includegraphics[width=0.131\textwidth]{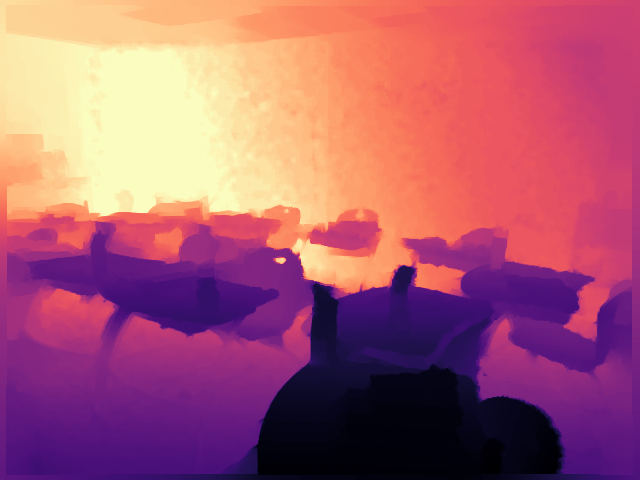}} 
\subfigure[]
{\includegraphics[width=0.131\textwidth]{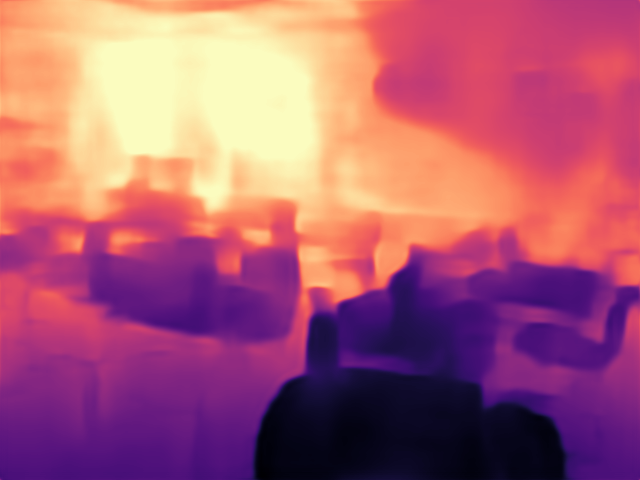}} 
\subfigure[]
{\includegraphics[width=0.131\textwidth]{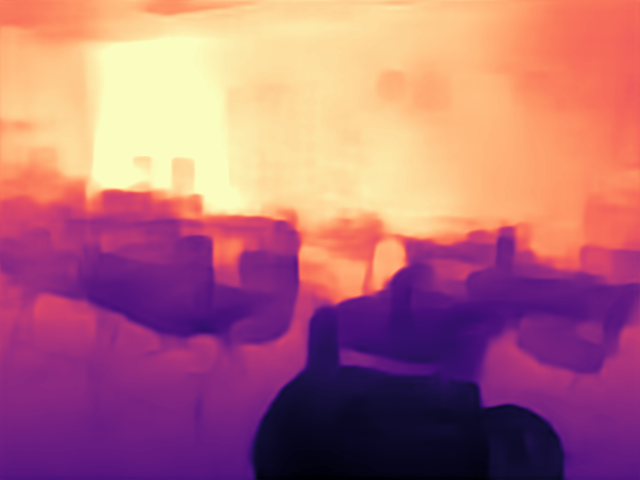}} 
\subfigure[]
{\includegraphics[width=0.131\textwidth]{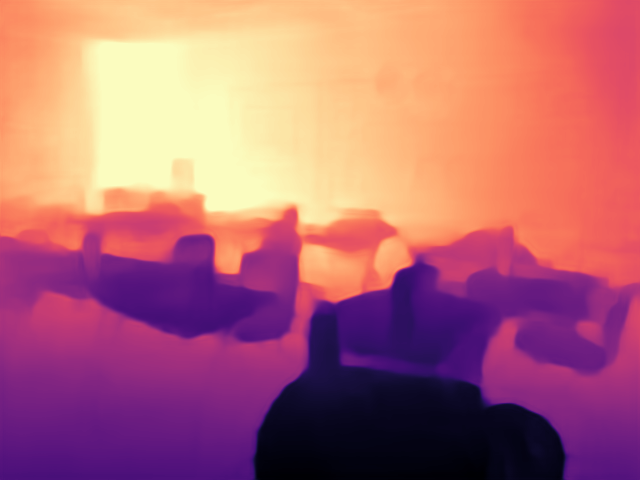}} 
\subfigure[]
{\includegraphics[width=0.131\textwidth]{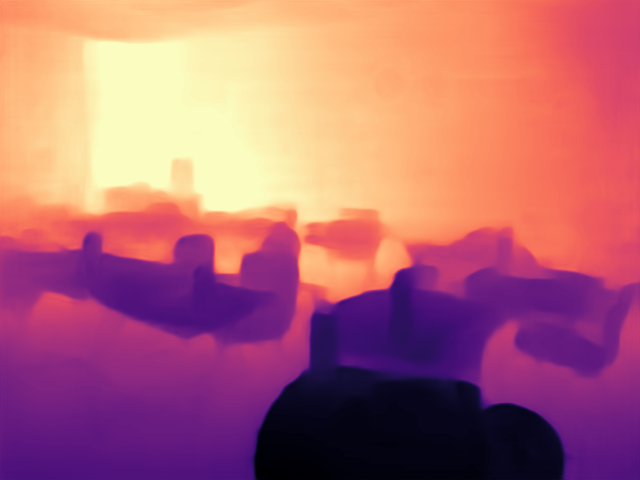}} \\\vspace{-20.5pt}

\subfigure[]
{\includegraphics[width=0.131\textwidth]{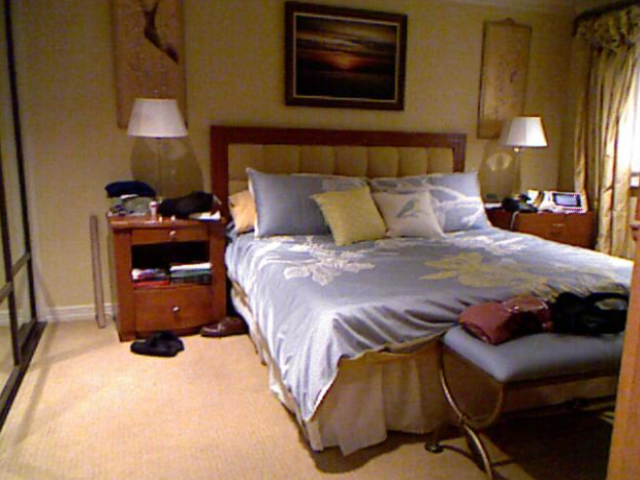}} 
\subfigure[]
{\includegraphics[width=0.131\textwidth]{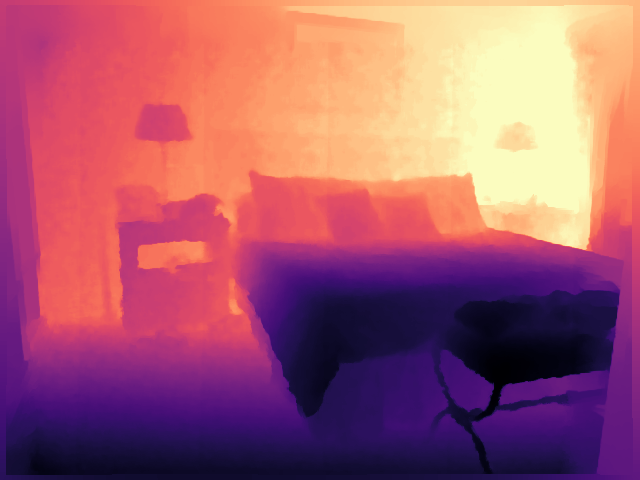}} 
\subfigure[]
{\includegraphics[width=0.131\textwidth]{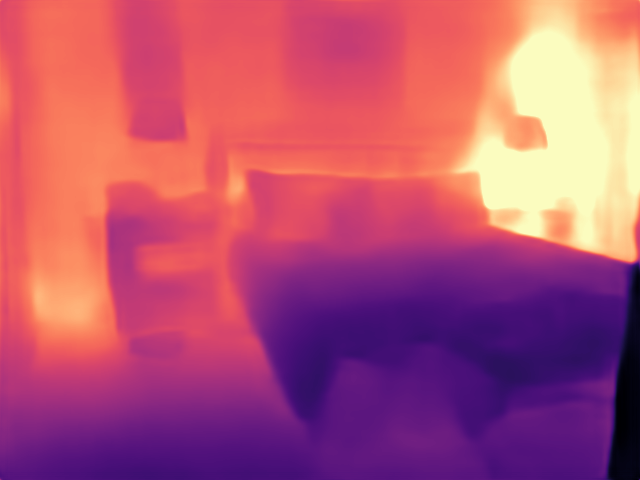}} 
\subfigure[]
{\includegraphics[width=0.131\textwidth]{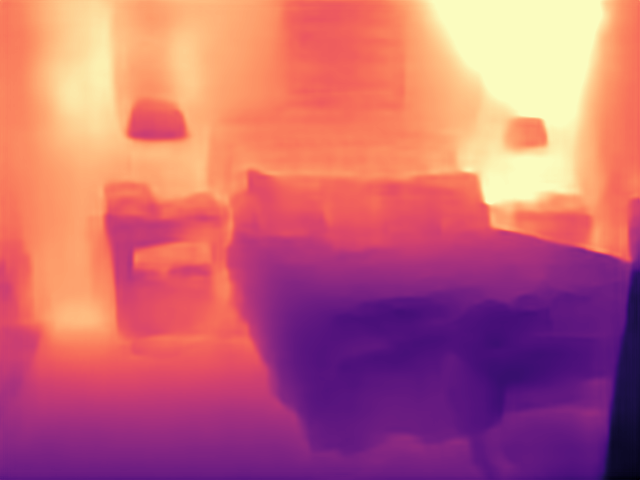}} 
\subfigure[]
{\includegraphics[width=0.131\textwidth]{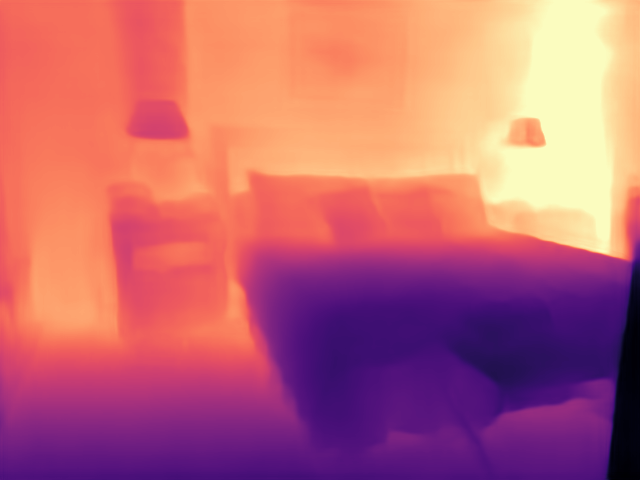}} 
\subfigure[]
{\includegraphics[width=0.131\textwidth]{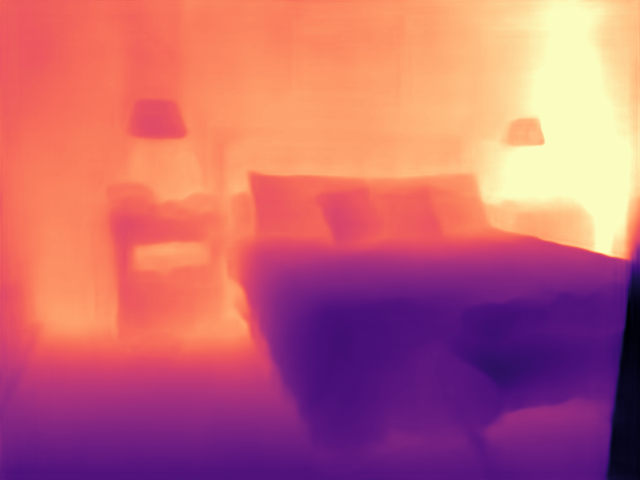}} \\\vspace{-20.5pt}

\centering
\subfigure[(a) RGB]
{\includegraphics[width=0.131\textwidth]{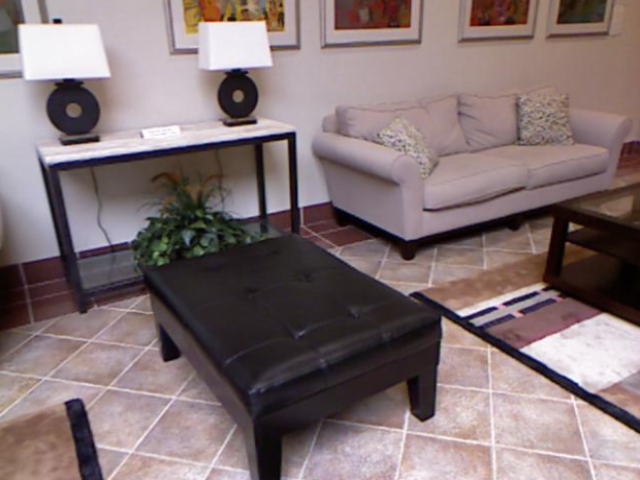}} 
\centering
\subfigure[(b) GT depth]
{\includegraphics[width=0.131\textwidth]{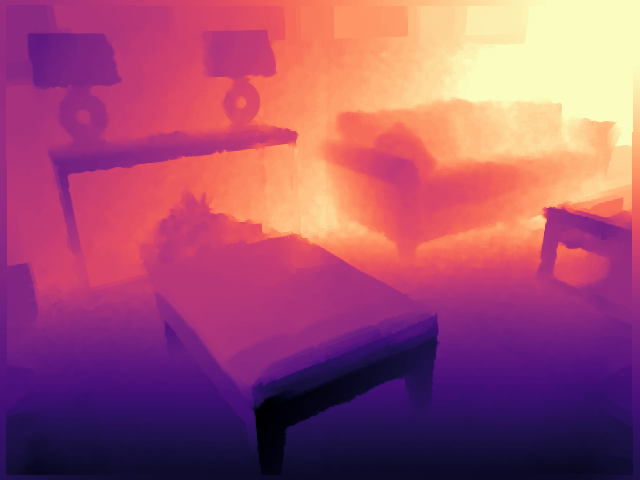}} 
\centering
\subfigure[(c) Baseline]
{\includegraphics[width=0.131\textwidth]{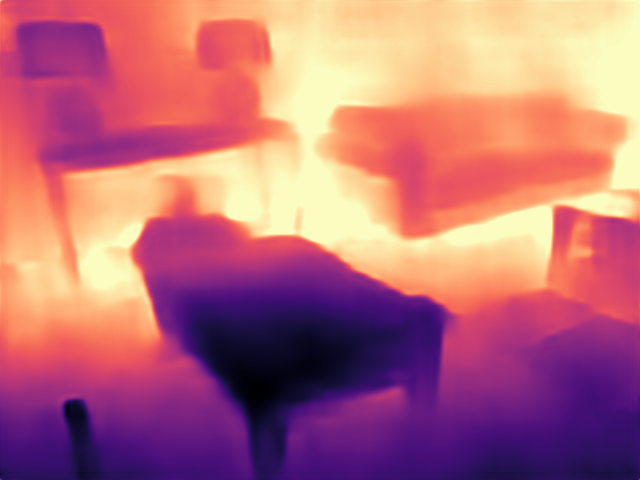}} 
\centering
\subfigure[(d) Ours]
{\includegraphics[width=0.131\textwidth]{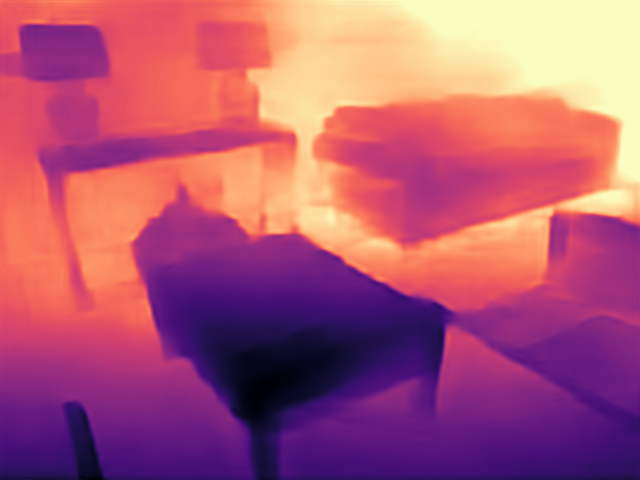}} 
\centering
\subfigure[(e) Baseline]
{\includegraphics[width=0.131\textwidth]{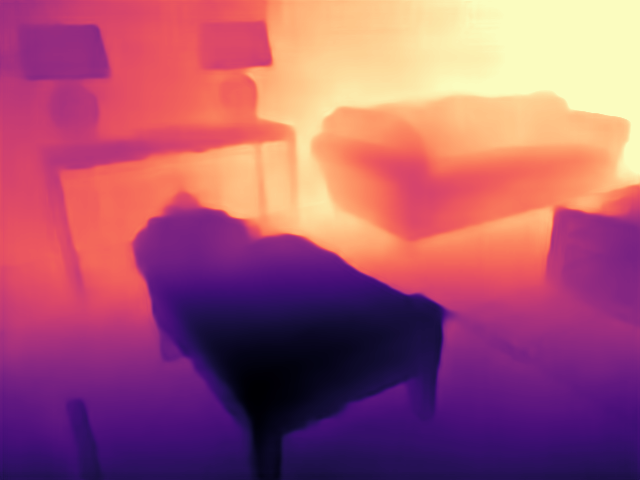}} 
\centering
\subfigure[(f) Ours]
{\includegraphics[width=0.131\textwidth]{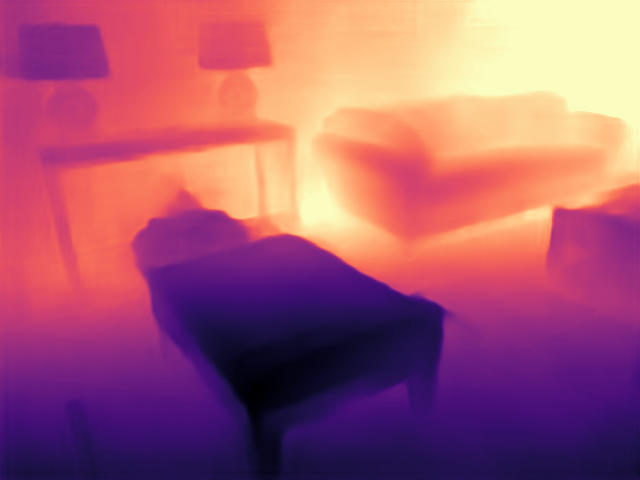}} \\
% \vspace{-5pt}
\vspace{5pt}
\caption{\textbf{Qualitative results on the NYU-Depth-v2 dataset~\cite{silberman2012indoor}:} (a) RGB image, (b) ground-truth depth map, and predicted depth maps by (c), (e) baseline, and (d), (f) ours using 100 and 10,000 labeled frames, respectively.}%
\label{fig_s1}\vspace{-20pt}
\end{figure*}

\begin{figure*}[ht]
\centering
\renewcommand{\thesubfigure}{}
\subfigure[]
{\includegraphics[width=0.19\textwidth]{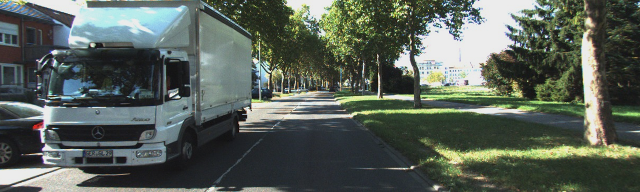}} 
\subfigure[]
{\includegraphics[width=0.19\textwidth]{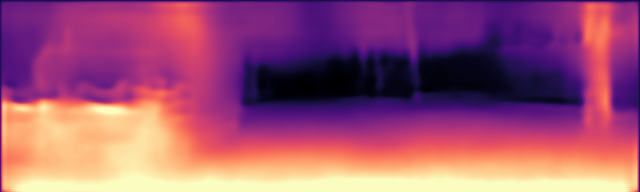}} 
\subfigure[]
{\includegraphics[width=0.19\textwidth]{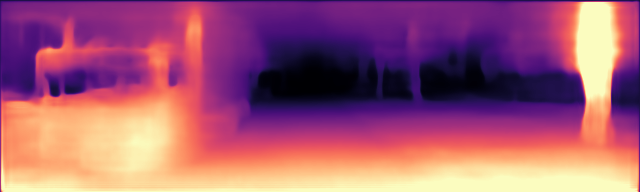}} 
\subfigure[]
{\includegraphics[width=0.19\textwidth]{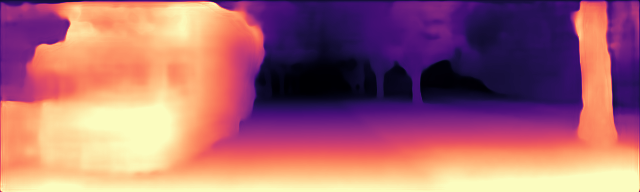}} 
\subfigure[]
{\includegraphics[width=0.19\textwidth]{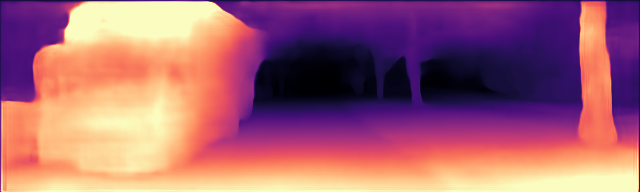}} \\\vspace{-15pt}

\subfigure[]
{\includegraphics[width=0.19\textwidth]{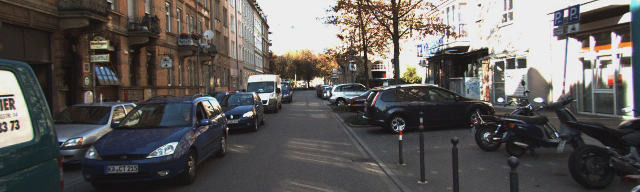}} 
\subfigure[]
{\includegraphics[width=0.19\textwidth]{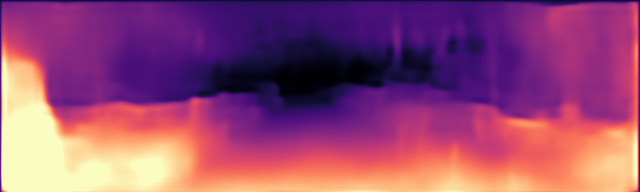}} 
\subfigure[]
{\includegraphics[width=0.19\textwidth]{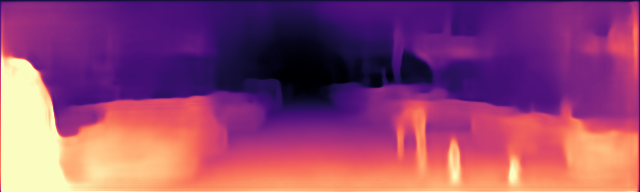}} 
\subfigure[]
{\includegraphics[width=0.19\textwidth]{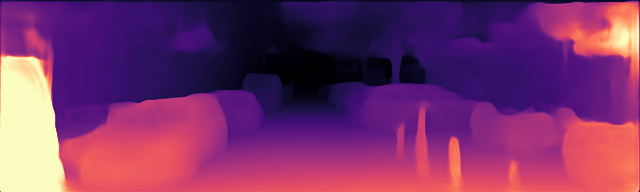}} 
\subfigure[]
{\includegraphics[width=0.19\textwidth]{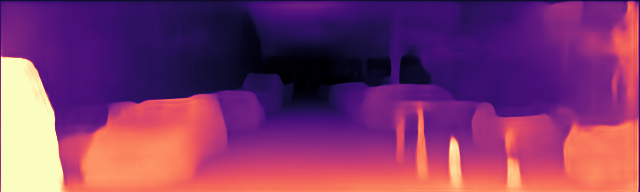}} \\\vspace{-15pt}

\subfigure[]
{\includegraphics[width=0.19\textwidth]{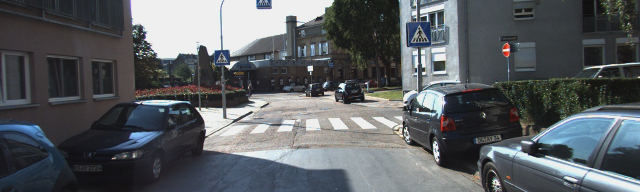}} 
\subfigure[]
{\includegraphics[width=0.19\textwidth]{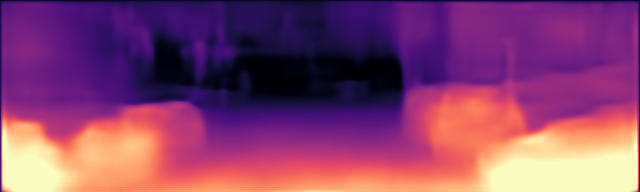}} 
\subfigure[]
{\includegraphics[width=0.19\textwidth]{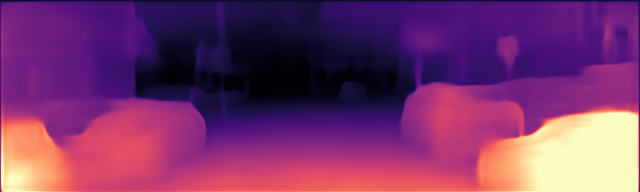}} 
\subfigure[]
{\includegraphics[width=0.19\textwidth]{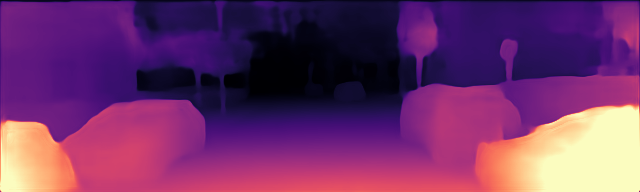}} 
\subfigure[]
{\includegraphics[width=0.19\textwidth]{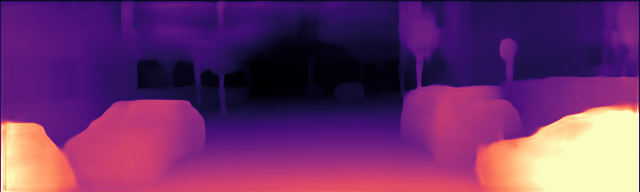}} \\\vspace{-15pt}

\subfigure[]
{\includegraphics[width=0.19\textwidth]{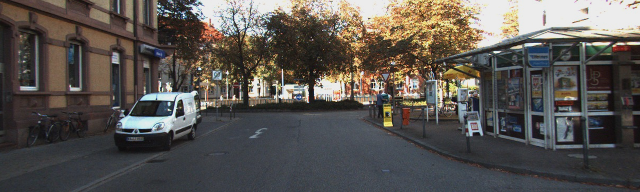}} 
\subfigure[]
{\includegraphics[width=0.19\textwidth]{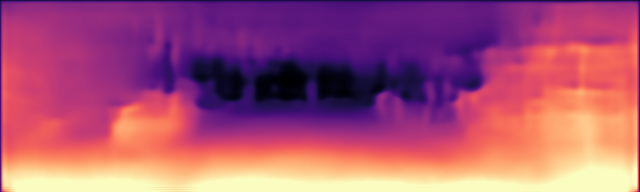}} 
\subfigure[]
{\includegraphics[width=0.19\textwidth]{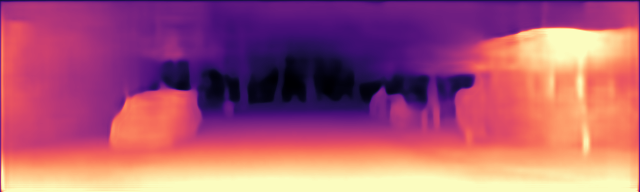}} 
\subfigure[]
{\includegraphics[width=0.19\textwidth]{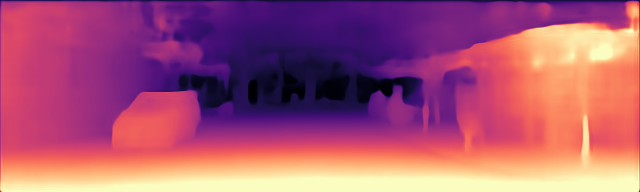}} 
\subfigure[]
{\includegraphics[width=0.19\textwidth]{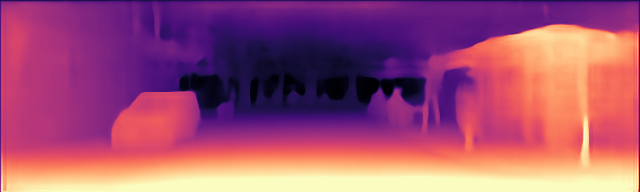}} \\\vspace{-15pt}

\subfigure[]
{\includegraphics[width=0.19\textwidth]{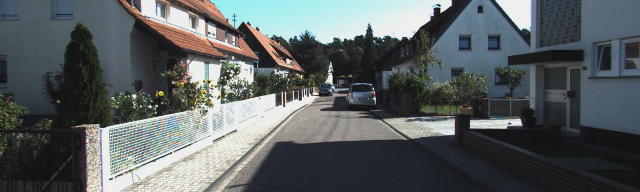}} 
\subfigure[]
{\includegraphics[width=0.19\textwidth]{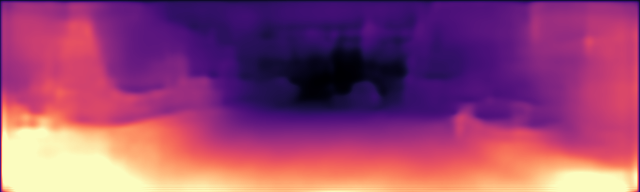}} 
\subfigure[]
{\includegraphics[width=0.19\textwidth]{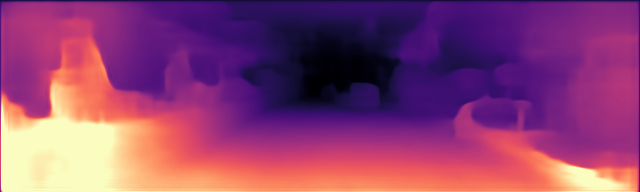}} 
\subfigure[]
{\includegraphics[width=0.19\textwidth]{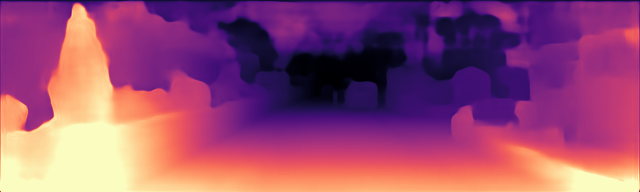}} 
\subfigure[]
{\includegraphics[width=0.19\textwidth]{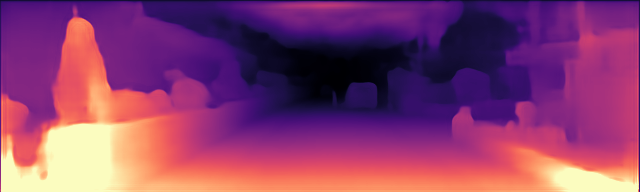}} \\\vspace{-15pt}

\subfigure[]
{\includegraphics[width=0.19\textwidth]{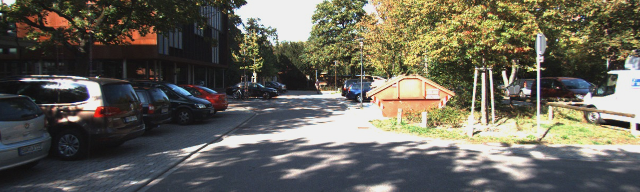}} 
\subfigure[]
{\includegraphics[width=0.19\textwidth]{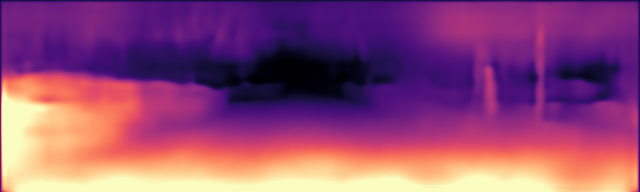}} 
\subfigure[]
{\includegraphics[width=0.19\textwidth]{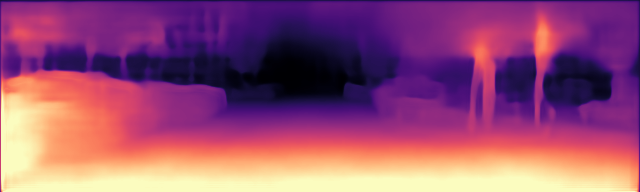}} 
\subfigure[]
{\includegraphics[width=0.19\textwidth]{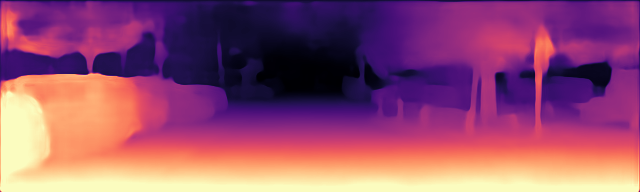}} 
\subfigure[]
{\includegraphics[width=0.19\textwidth]{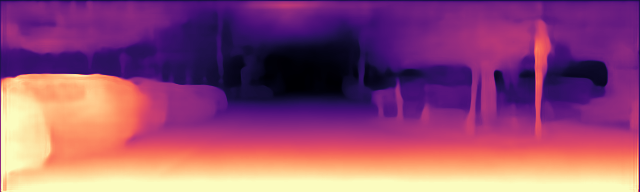}} \\\vspace{-15pt}

\subfigure[]
{\includegraphics[width=0.19\textwidth]{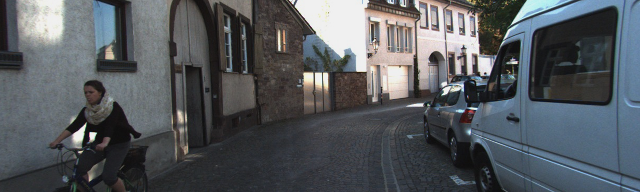}} 
\subfigure[]
{\includegraphics[width=0.19\textwidth]{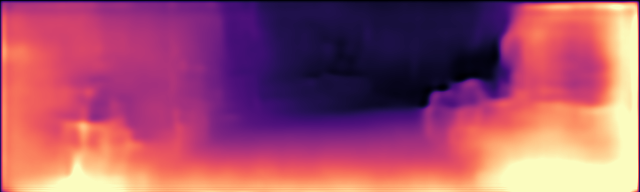}} 
\subfigure[]
{\includegraphics[width=0.19\textwidth]{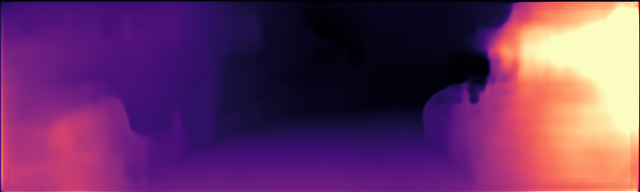}} 
\subfigure[]
{\includegraphics[width=0.19\textwidth]{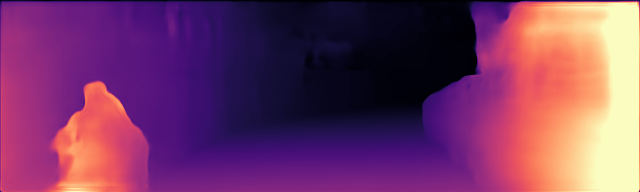}} 
\subfigure[]
{\includegraphics[width=0.19\textwidth]{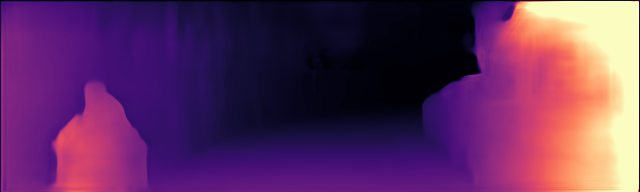}} \\\vspace{-15pt}

\subfigure[]
{\includegraphics[width=0.19\textwidth]{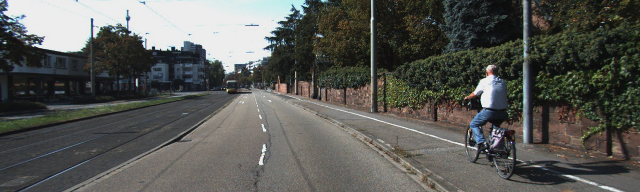}} 
\subfigure[]
{\includegraphics[width=0.19\textwidth]{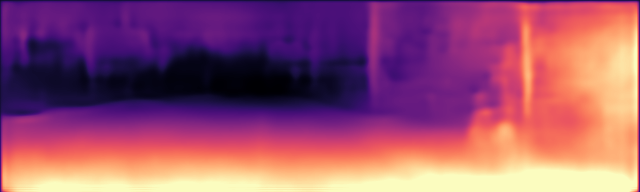}} 
\subfigure[]
{\includegraphics[width=0.19\textwidth]{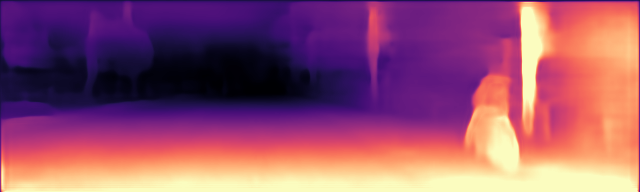}} 
\subfigure[]
{\includegraphics[width=0.19\textwidth]{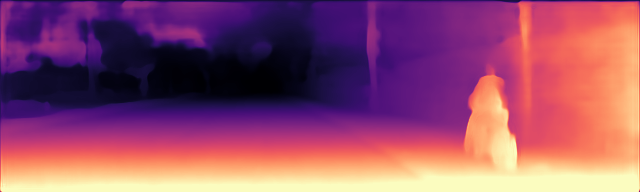}} 
\subfigure[]
{\includegraphics[width=0.19\textwidth]{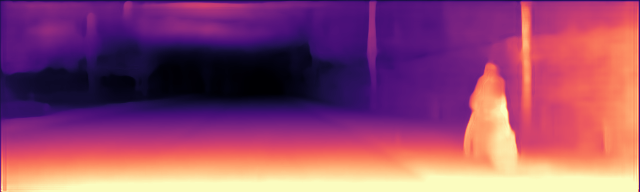}} \\\vspace{-15pt}

\subfigure[]
{\includegraphics[width=0.19\textwidth]{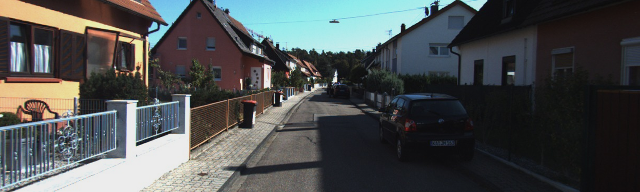}} 
\subfigure[]
{\includegraphics[width=0.19\textwidth]{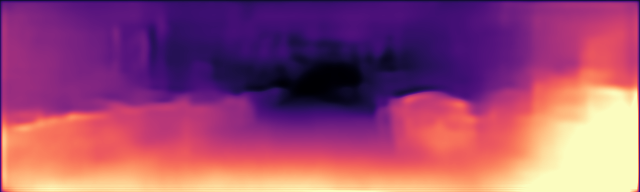}} 
\subfigure[]
{\includegraphics[width=0.19\textwidth]{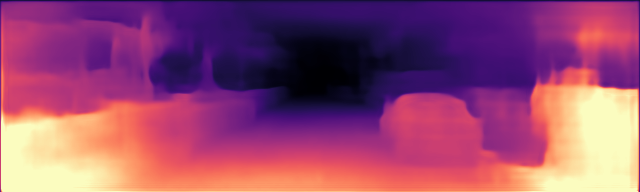}} 
\subfigure[]
{\includegraphics[width=0.19\textwidth]{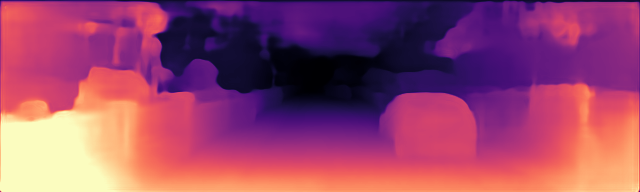}} 
\subfigure[]
{\includegraphics[width=0.19\textwidth]{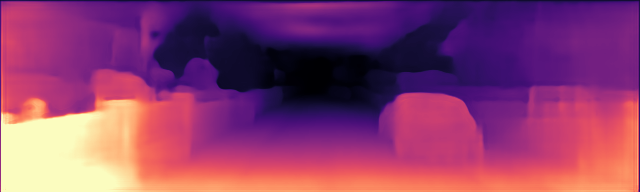}} \\\vspace{-15pt}

\subfigure[]
{\includegraphics[width=0.19\textwidth]{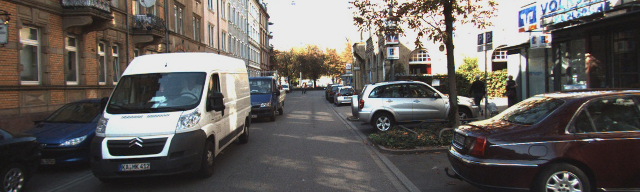}} 
\subfigure[]
{\includegraphics[width=0.19\textwidth]{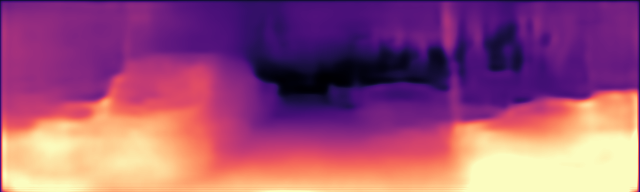}} 
\subfigure[]
{\includegraphics[width=0.19\textwidth]{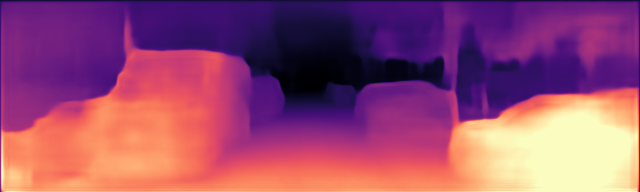}} 
\subfigure[]
{\includegraphics[width=0.19\textwidth]{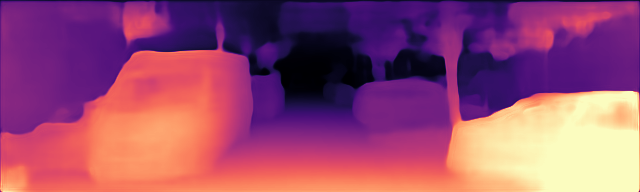}} 
\subfigure[]
{\includegraphics[width=0.19\textwidth]{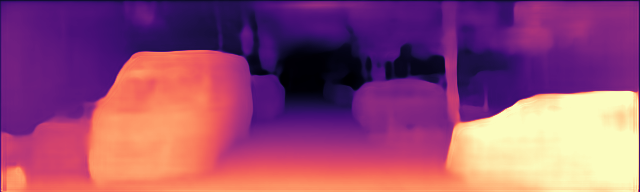}} \\\vspace{-15pt}

\centering
\subfigure[(a) RGB]
{\includegraphics[width=0.19\textwidth]{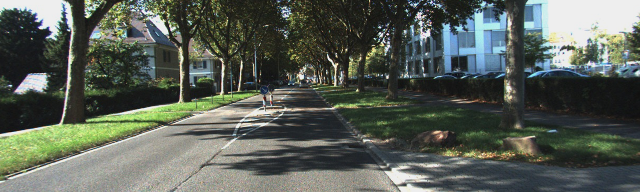}} 
\centering
\subfigure[(b) Baseline]
{\includegraphics[width=0.19\textwidth]{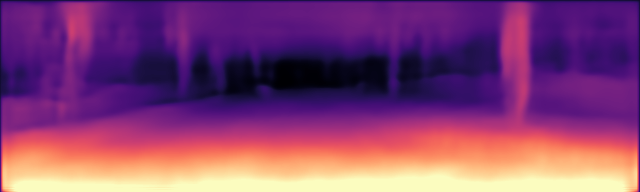}} 
\centering
\subfigure[(c) Ours]
{\includegraphics[width=0.19\textwidth]{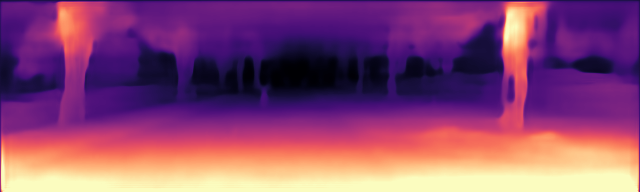}} 
\centering
\subfigure[(d) Baseline]
{\includegraphics[width=0.19\textwidth]{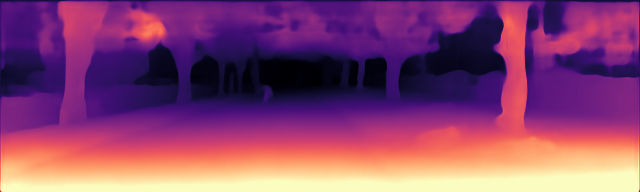}} 
\centering
\subfigure[(e) Ours]
{\includegraphics[width=0.19\textwidth]{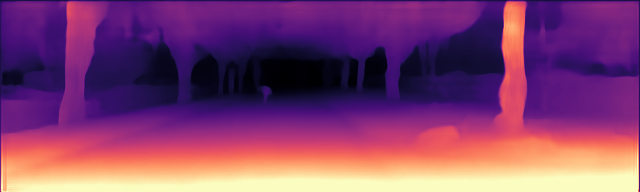}} \\
% \vspace{-5pt}
\vspace{5pt}
\caption{\textbf{Qualitative results on the KITTI dataset~\cite{geiger2012we}:} (a) RGB image, predicted depth maps by (b), (d) baseline, and (c), (e) ours using 100 and 10,000 labeled frames, respectively.}%
\label{fig_s2}\vspace{-10pt}
\end{figure*}

In the main paper, we have visualized comparisons of the baseline and our methods on the NYU-Depth-v2 dataset~\cite{silberman2012indoor} and KITTI dataset~\cite{geiger2012we}.
In this section, we provide additional qualitative results after having trained on 100 and 10,000 frames in Fig.~\ref{fig_s1} and Fig.~\ref{fig_s2}. Our method provides better depth results than the baseline.

\clearpage

\section*{Appendix C. Additional Ablation Study}
In the main paper, we control the intensity of the proposed augmentation strategy to the networks by adjusting $K$. The quantitative evaluation results for this study are presented in Table~\ref{tab:numberK}. 
To further enhance the representation power of the encoder, we incorporate a feature consistency loss between the encoded features of the weak and strong branches. Our experiments reveal that training did not reach convergence when the predictor head was removed, which led to collapsing. Our framework without a predictor head is conceptually analogous to a vanilla siamese network, which cannot prevent collapsing~\cite{zhang2022does}. In this set of experiments, we evaluate the performance of the predictor head used to provide better results with feature consistency loss. Table~\ref{tab:predictor} displays the results, indicating that a simple MLP is preferred for computational efficiency, despite a single Transformer block or an MLP showing comparable performance.

\begin{table*}[h]
\begin{minipage}{.5\linewidth}
    \centering
    \begin{tabular}{l|ccc}
            \toprule
            $K$ & AbsRel $\downarrow$ & RMSE $\downarrow$ &\textbf{$\delta$}$\uparrow$ \\
            \midrule
            $K=4$  &0.131 & 4.456 &0.850\\
            $K=16$ &0.128 & 4.298 &\textbf{0.855}\\
            $K=64$ &\textbf{0.124} & \textbf{4.263} &\textbf{0.855}\\
            $K=128$ &0.132 & 4.382 &0.849\\
            \bottomrule
    \end{tabular}
    \vspace{8pt}
    \caption{\textbf{Influence of the number of $K$.}}
    \label{tab:numberK}
\end{minipage}%
\begin{minipage}{.5\linewidth}
    \centering
    \begin{tabular}{c|c|c|c}
            \toprule
            Method & Blocks &  AbsRel $\downarrow$ & \textbf{$\delta$}$\uparrow$ \\
            \midrule
            w/o head       &- &0.317 &0.423\\
            Transformer    &1 &0.125 &\textbf{0.856}\\
            Transformer    &2 &0.129 &0.848\\
            MLP            &2 &\textbf{0.124} &0.855\\
            \bottomrule
    \end{tabular}
    \vspace{8pt}
    \caption{\textbf{Comparison of using different predictor head.}}
    \label{tab:predictor}
\end{minipage}
\end{table*}

\section*{Appendix D. Domain Adaptation} 
\begin{figure*}[ht]
\centering
\renewcommand{\thesubfigure}{}

\subfigure[]
{\includegraphics[width=0.22\textwidth]{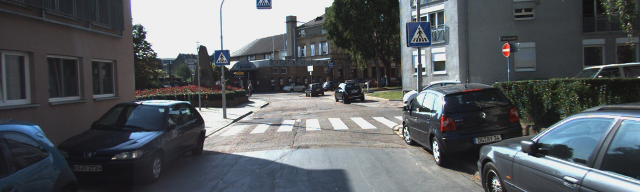}} 
\subfigure[]
{\includegraphics[width=0.22\textwidth]{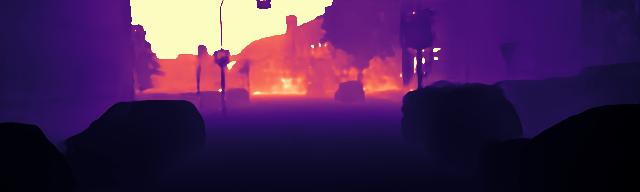}} 
\subfigure[]
{\includegraphics[width=0.22\textwidth]{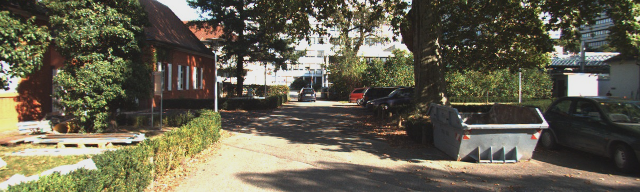}} 
\subfigure[]
{\includegraphics[width=0.22\textwidth]{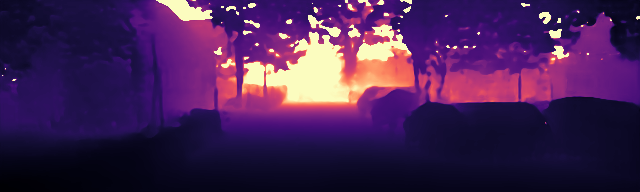}} \\\vspace{-20.5pt}

\subfigure[]
{\includegraphics[width=0.22\textwidth]{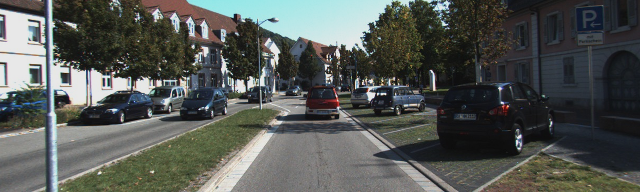}} 
\subfigure[]
{\includegraphics[width=0.22\textwidth]{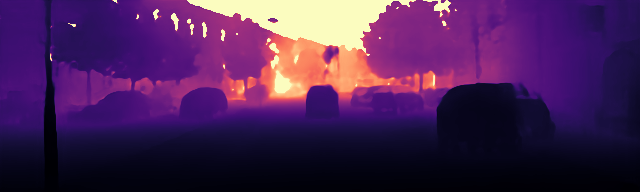}} 
\subfigure[]
{\includegraphics[width=0.22\textwidth]{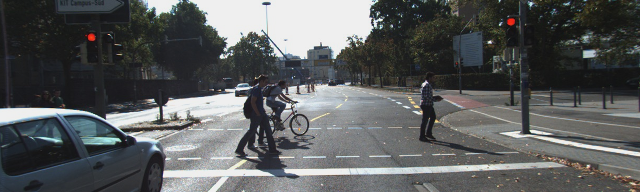}} 
\subfigure[]
{\includegraphics[width=0.22\textwidth]{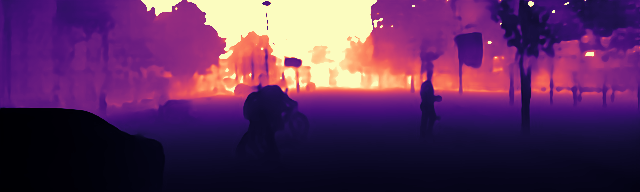}} \\\vspace{-20.5pt}

\subfigure[]
{\includegraphics[width=0.22\textwidth]{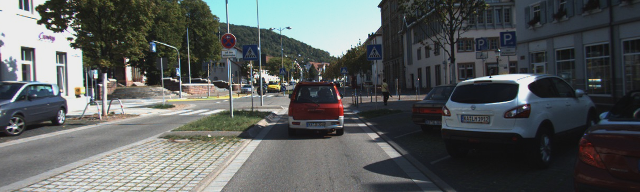}} 
\subfigure[]
{\includegraphics[width=0.22\textwidth]{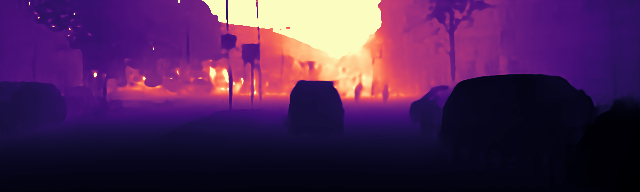}} 
\subfigure[]
{\includegraphics[width=0.22\textwidth]{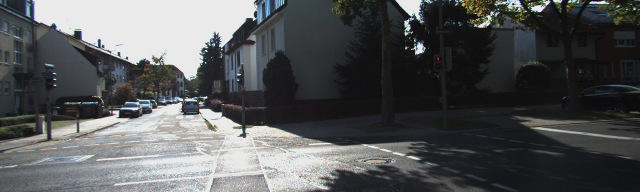}} 
\subfigure[]
{\includegraphics[width=0.22\textwidth]{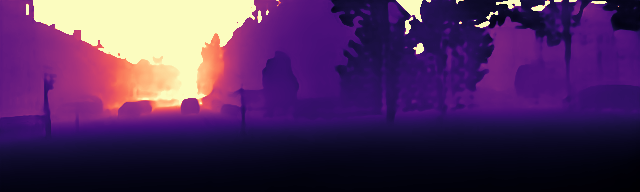}} \\\vspace{-20.5pt}

\subfigure[]
{\includegraphics[width=0.22\textwidth]{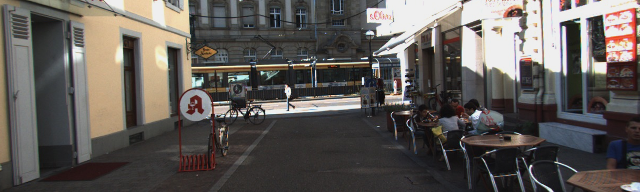}} 
\subfigure[]
{\includegraphics[width=0.22\textwidth]{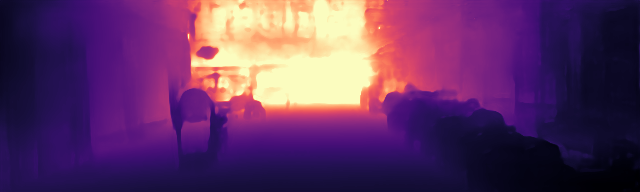}} 
\subfigure[]
{\includegraphics[width=0.22\textwidth]{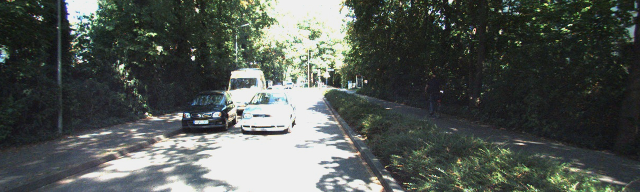}} 
\subfigure[]
{\includegraphics[width=0.22\textwidth]{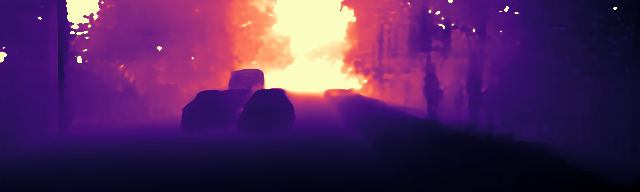}} \\\vspace{-20.5pt}

\centering
\subfigure[(a) RGB]
{\includegraphics[width=0.22\textwidth]{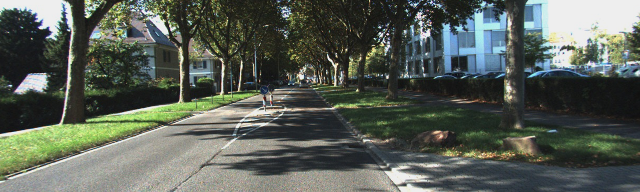}} 
\centering
\subfigure[(b) Depth]
{\includegraphics[width=0.22\textwidth]{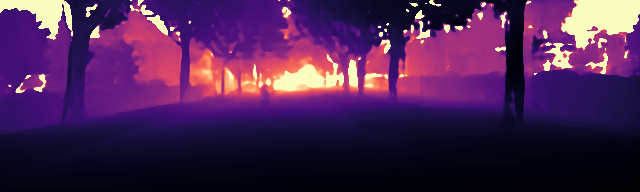}} 
\centering
\subfigure[(c) RGB]
{\includegraphics[width=0.22\textwidth]{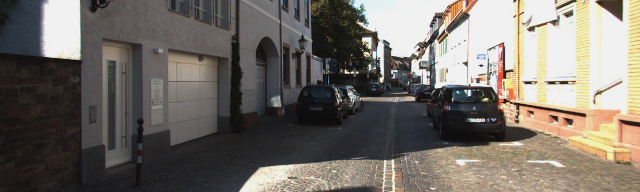}} 
\centering
\subfigure[(d) Depth]
{\includegraphics[width=0.22\textwidth]{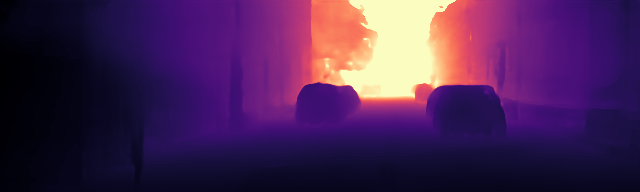}} \\
% \vspace{-5pt}
\vspace{5pt}
\caption{\textbf{Qualitative results on the KITTI dataset~\cite{geiger2012we}:} (a), (c) RGB image, and (b), (d) depth map. Our framework proves to work well in domain adaptation task on real-world images.
}%
\label{figdom}\vspace{-10pt}
\end{figure*}

In the main paper, we simply extend MaskingDepth to domain adaptation by following \cite{melas2021pixmatch} and provide the additional qualitative results in this section. As demonstrated in Fig.~\ref{figdom}, unlike generative models, MaskingDepth is simply adaptable and lightweight while achieving reasonable performance in the domain adaptation from virtual KITTI (synthetic)~\cite{gaidon2016virtual} to KITTI (real)~\cite{geiger2012we}.

\clearpage

\section*{Appendix E. Limitations of Na\"ive Masking}
\begin{figure*}[ht]
\centering
\renewcommand{\thesubfigure}{}

\subfigure[]
{\includegraphics[width=0.19\textwidth]{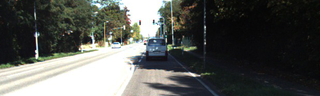}} 
\subfigure[]
{\includegraphics[width=0.19\textwidth]{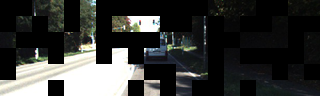}} 
\subfigure[]
{\includegraphics[width=0.19\textwidth]{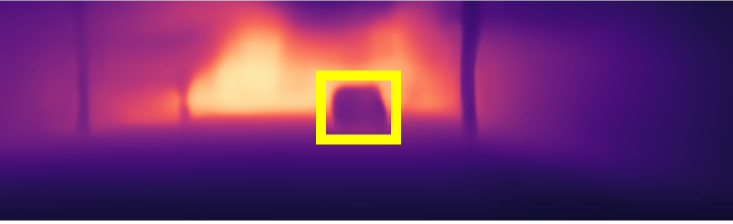}} 
\subfigure[]
{\includegraphics[width=0.19\textwidth]{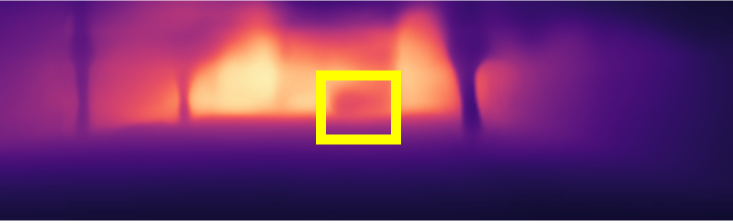}} 
\subfigure[]
{\includegraphics[width=0.19\textwidth]{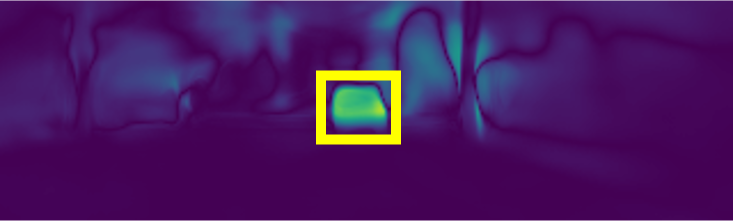}} \\\vspace{-15pt}

\subfigure[]
{\includegraphics[width=0.19\textwidth]{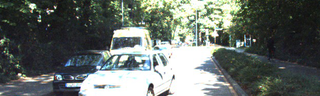}} 
\subfigure[]
{\includegraphics[width=0.19\textwidth]{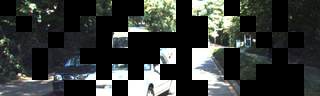}} 
\subfigure[]
{\includegraphics[width=0.19\textwidth]{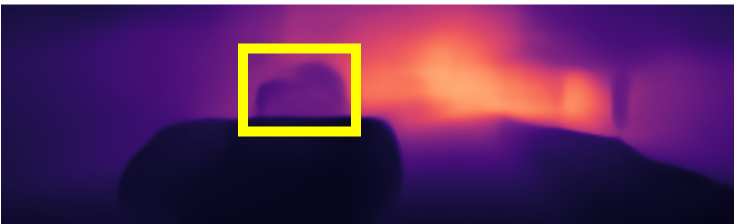}} 
\subfigure[]
{\includegraphics[width=0.19\textwidth]{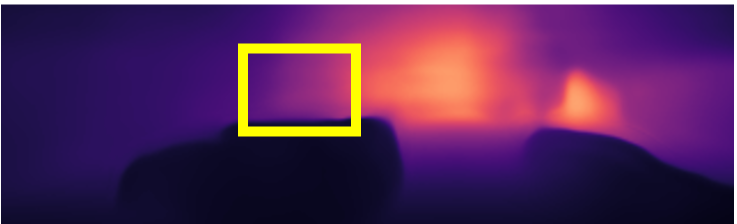}} 
\subfigure[]
{\includegraphics[width=0.19\textwidth]{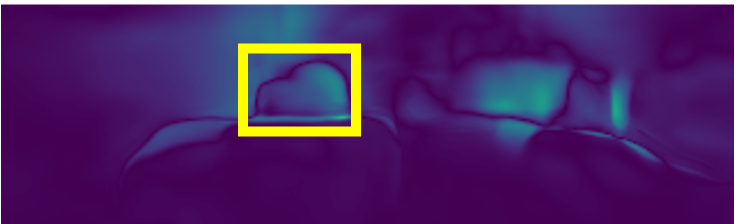}} \\\vspace{-15pt}

\subfigure[]
{\includegraphics[width=0.19\textwidth]{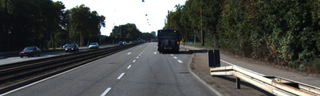}} 
\subfigure[]
{\includegraphics[width=0.19\textwidth]{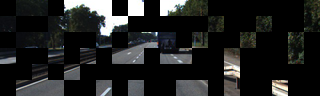}} 
\subfigure[]
{\includegraphics[width=0.19\textwidth]{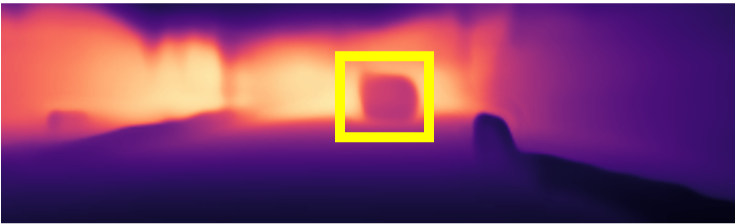}} 
\subfigure[]
{\includegraphics[width=0.19\textwidth]{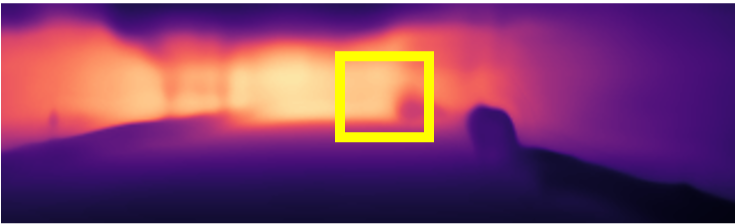}} 
\subfigure[]
{\includegraphics[width=0.19\textwidth]{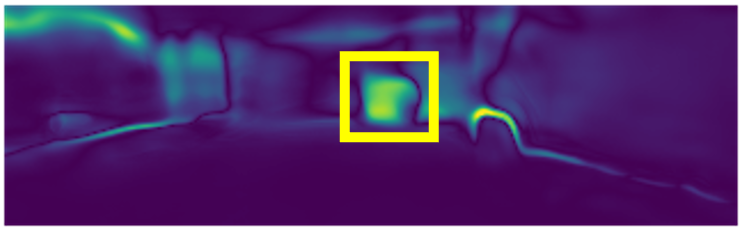}} \\\vspace{-15pt}

\centering
\subfigure[(a) RGB]
{\includegraphics[width=0.19\textwidth]{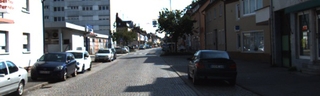}} 
\centering
\subfigure[(b) Masked RGB]
{\includegraphics[width=0.19\textwidth]{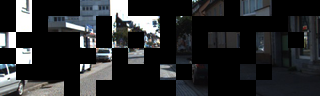}} 
\centering
\subfigure[(c) Depths (weak)]
{\includegraphics[width=0.19\textwidth]{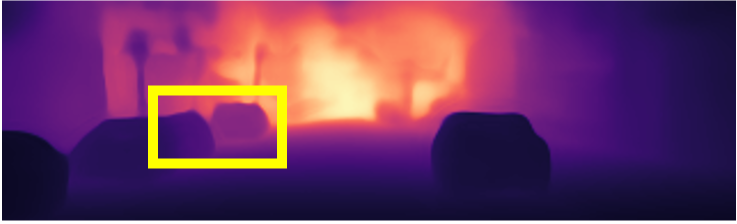}} 
\centering
\subfigure[(d) Depths (strong)]
{\includegraphics[width=0.19\textwidth]{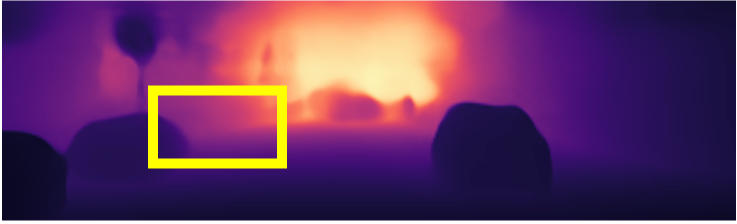}} 
\centering
\subfigure[(e) Difference maps]
{\includegraphics[width=0.19\textwidth]{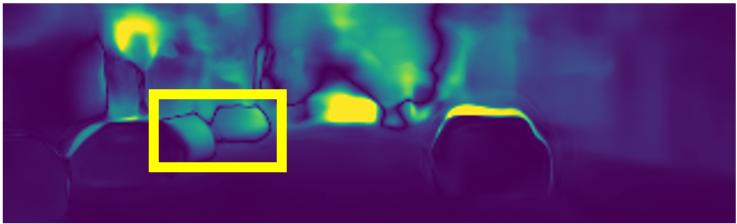}} \\
% \vspace{-5pt}
\vspace{5pt}
\caption{\textbf{Missing object cases of na\"ive masking on the KITTI dataset~\cite{geiger2012we}:} (a) RGB images, (b) masked RGB images, (c) depth maps predicted from (a), (d) depth maps predicted from (b), and (e) difference maps between (c) and (d).}%
\label{fig_s4}\vspace{-10pt}
\end{figure*}

\begin{figure*}[ht]
\centering
\renewcommand{\thesubfigure}{}
\subfigure[]
{\includegraphics[width=0.19\textwidth]{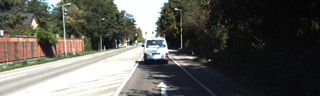}} 
\subfigure[]
{\includegraphics[width=0.19\textwidth]{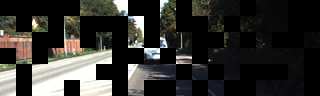}} 
\subfigure[]
{\includegraphics[width=0.19\textwidth]{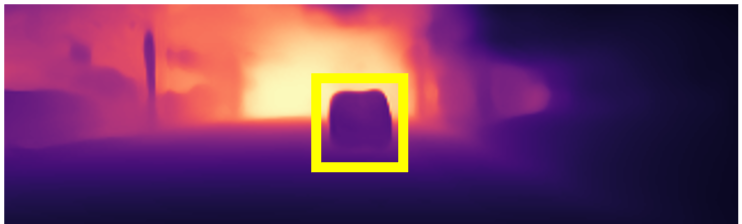}} 
\subfigure[]
{\includegraphics[width=0.19\textwidth]{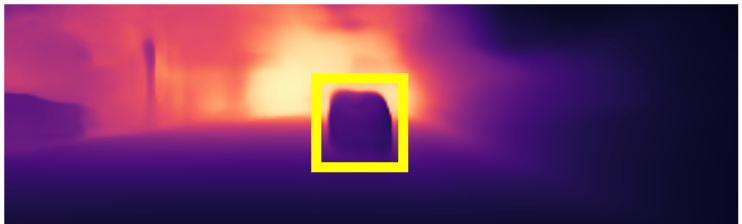}} 
\subfigure[]
{\includegraphics[width=0.19\textwidth]{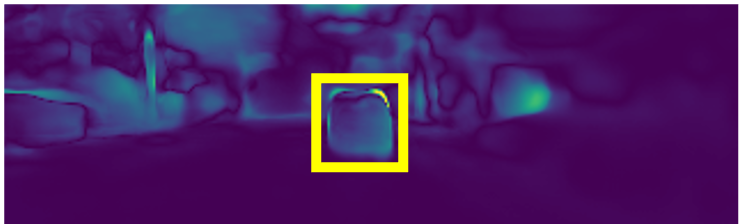}} \\\vspace{-15pt}

\subfigure[]
{\includegraphics[width=0.19\textwidth]{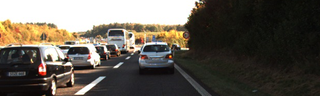}} 
\subfigure[]
{\includegraphics[width=0.19\textwidth]{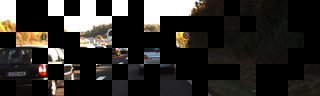}} 
\subfigure[]
{\includegraphics[width=0.19\textwidth]{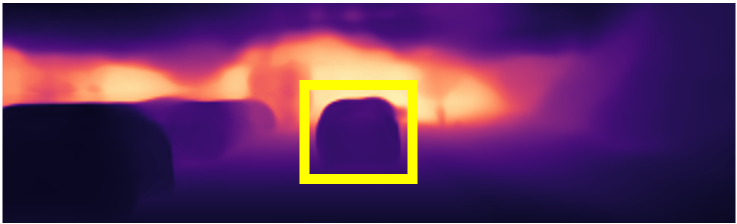}} 
\subfigure[]
{\includegraphics[width=0.19\textwidth]{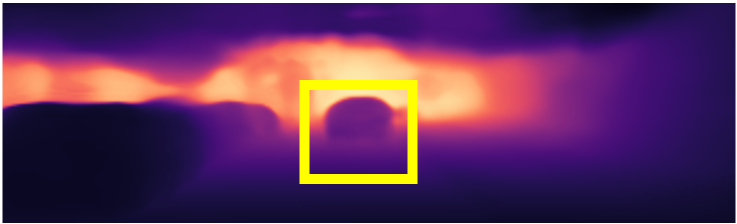}} 
\subfigure[]
{\includegraphics[width=0.19\textwidth]{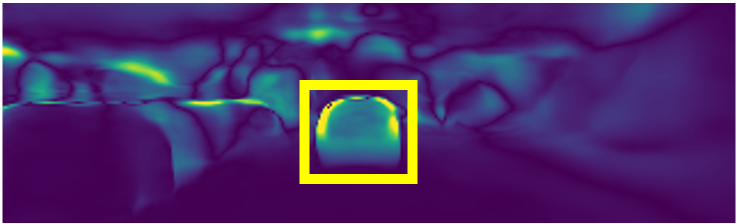}} \\\vspace{-15pt}

\subfigure[]
{\includegraphics[width=0.19\textwidth]{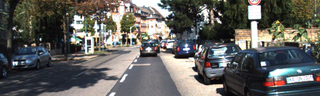}} 
\subfigure[]
{\includegraphics[width=0.19\textwidth]{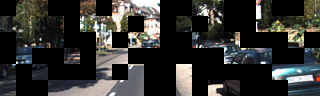}} 
\subfigure[]
{\includegraphics[width=0.19\textwidth]{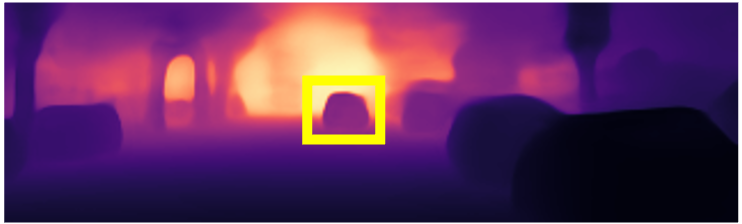}} 
\subfigure[]
{\includegraphics[width=0.19\textwidth]{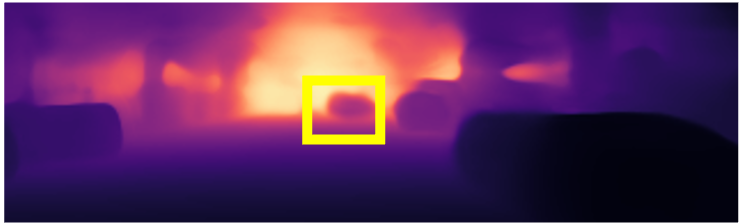}} 
\subfigure[]
{\includegraphics[width=0.19\textwidth]{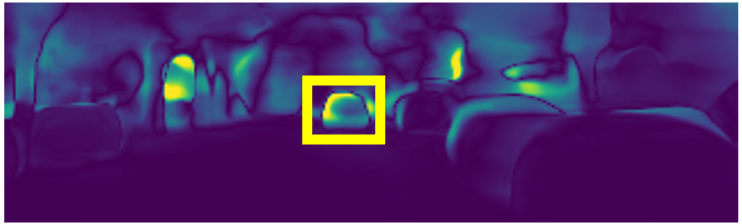}} \\\vspace{-15pt}

\centering
\subfigure[(a) RGB]
{\includegraphics[width=0.19\textwidth]{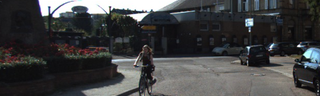}} 
\centering
\subfigure[(b) Masked RGB]
{\includegraphics[width=0.19\textwidth]{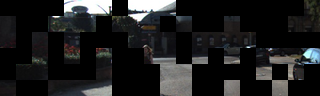}} 
\centering
\subfigure[(c) Depth map]
{\includegraphics[width=0.19\textwidth]{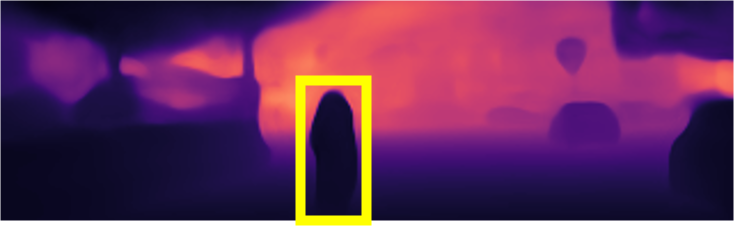}} 
\centering
\subfigure[(d) Masked depth map]
{\includegraphics[width=0.19\textwidth]{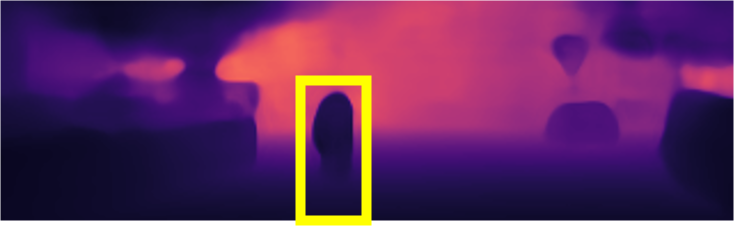}} 
\centering
\subfigure[(e) Difference maps]
{\includegraphics[width=0.19\textwidth]{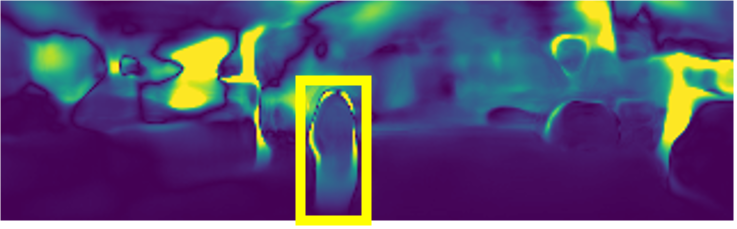}} \\
% \vspace{-5pt}
\vspace{5pt}
\caption{\textbf{Scale ambiguity cases of na\"ive masking on the KITTI dataset~\cite{geiger2012we}:} (a) RGB images, (b) masked RGB images, (c) depth maps predicted from (a), (d) depth maps predicted from (b), and (e) difference maps between (c) and (d).}%
\label{fig_s5}\vspace{-10pt}
\end{figure*}

In this section, we visualize failure cases of na\"ive masking strategy, specifically ``missing object" and ``scale ambiguity" cases. As shown in Fig.~\ref{fig_s4} we found that small objects often disappear even very small parts of them are masked out. In addition, locally inconsistent depth maps are shown in Fig.~\ref{fig_s5}. These cases can cause the performance drop in consistency regularization framework.

\clearpage
\section*{Appendix F. More Implementation Details}
\paragraph{PyTorch-like Pseudo-code for $K$-way disjoint masking.} We provide pseudo-code for $K$-way disjoint masking, in order to show how our novel technique can be implemented independently. This enables the effect of data augmentation while preserving the geometry.

\begin{figure}[ht]
\begin{algorithm}[H]
\small
\caption{\small Pseudo Code of $K$-way disjoint masking}
\label{alg:guide}
\definecolor{codeblue}{rgb}{0.25,0.5,0.5}
\definecolor{codegreen}{rgb}{0,0.6,0}
\definecolor{codekw}{rgb}{0.85, 0.18, 0.50}
\lstset{
  backgroundcolor=\color{white},
  basicstyle=\linespread{1.0}\fontsize{9pt}{9pt}\ttfamily\selectfont,
  columns=fullflexible,
  breaklines=false,
  captionpos=b,
  commentstyle=\fontsize{9pt}{9pt}\color{codegreen},
  keywordstyle=\fontsize{9pt}{9pt}\color{codekw},
  escapechar={|}, 
  xleftmargin=0.0in,
  xrightmargin=0.0in
}
\begin{lstlisting}[language=python]
    # Input: Transformer are basic transformer encoder blocks
    # Input: X (B x N x P) is patch embeddings
    # where N is number of patchs and P is embedding dimension
    # Input: K is number of masking subsets. 
    # Output: Z (B x N x P) is K-way disjointly encoded features
    def K_way_masked_transformer(transformer, X, K):
        B, N, _, _ = X.shape
        batch_range = torch.arange(B)[:, None]
        rand_indices = torch.rand(B, N).argsort(dim = -1)
        X = X[batch_range,rand_indices]
        # patch wise random shuffling on X
        # X is no longer spatial order after shuffling
    
        v = sorted([random.randint(1,N-1) for i in range(int(K-1))] + [0, N])
        # sampling subset split point randomly
        mask_v = torch.zeros(len(v[:-1]), N)
        for i in range(len(v[:-1])):
            mask_v[i, v[i]:v[i+1]] = 1.0
    
        attention_mask = mask_v.transpose(0,1) @ mask_v
        # mask has shape of (B x N x N)
        # attention matrix masking from different subset
        # by applying this attention mechanism proceeds semi-globally
    
        Z_ = transformer(x, attention_mask)
        # Z (B x N x P)) is K-way masked token embeddings
        reform_indices = torch.argsort(rand_indices, dim=1)
    
        return Z_[batch_range, reform_indices]
        # reassemble features to spatial order and return
    
    # In the transformer, attention method
    # Input: Q, K, V (B x N x P) is Query, Key, Value vectors
    # Input: mask (B x N x N) attention mask
    # Output: V' (B x N x P) mask attentioned Value vectors
    def attention(Q, K, V, mask):
        dots = torch.matmul(Q, K.transpose(-1, -2)) * sqrt(Q.shape[-1])
        ## general attention matrix that has shape of (B x N x N)
        if not(mask == None):
            dots[:,:, mask==0.0] = -10.0
        ## replace value of masked region with negatives befor softmax
        ## so it's attention distribution value goes nearly zero after softmax 
        attn = Softmax(dots, dim=-1)    
        V = torch.matmul(attn, V)
        return V
\end{lstlisting}
\end{algorithm}
% \end{minipage}
\end{figure}

\clearpage
\paragraph{Network Architecture Details.}
We summarize the detailed network architecture of our model in Table~\ref{tab:s_network}. Our encoder follows ViT~\cite{dosovitskiy2020image} without class token. We use 12 transformer blocks(ViT-Base) and extract features 2, 5, 8, and 11 for skip connection input ($0<=\mathtt{N}<=11$). The Decoder structure follows DPT~\cite{ranftl2021vision}, and the Decoder part in the Table~\ref{tab:s_network} indicates 4 skip connections. However, there are no readout operations because no class token is used. 

\section*{Appendix G. Limitations and Broader Impact}
\paragraph{Limitations.} In this paper, we introduce a novel consistency regularization framework for semi-supervised monocular depth estimation, called MaskingDepth. However, our approach learns unlabeled data by following the guidance of a small number of labeled data, so it may fail to predict depth in sparse annotations such as sky regions. Another limitation is that our method performs well on the KITTI dataset~\cite{geiger2012we}, while the performance improves relatively less on images with a higher variety of content, e.g., as in the NYU-Depth-v2 dataset~\cite{silberman2012indoor}.

\paragraph{Broader Impact.} In the future, we aim to apply our method in various data domains, and develop the performance of our framework by leveraging the newly studied data augmentation techniques. 
Our work is an essential study in order to be completely free from the dependence of the data, and provides the possibility to replace the existing self-supervised monocular depth estimation methods which require specific dataset, such as stereo pairs and video sequences. 

\begin{table}[h] 
    \centering
    \scalebox{0.85}{
    \begin{tabular}{>{\centering}m{0.30\linewidth}
    >{\centering}m{0.30\linewidth} 
    >{\centering}m{0.30\linewidth}}
    &\textbf{PatchEmbedding}& \tabularnewline
    Layer & Parameters $(\mathtt{in},\mathtt{out},\mathtt{k},\mathtt{s},\mathtt{p})$ & Output shape $(C \times H \times W)$ \tabularnewline
    \hlinewd{1.6pt}
    Conv & $(3,768,16,16,0)$ & $(768,12,40)$ \tabularnewline
    Rerange & $ - $ & $(HW=480,C=768)$ \tabularnewline
    \hlinewd{1.6pt}

    &&\tabularnewline 
    \multicolumn{3}{c}{\textbf{Transformer Encoder $\mathtt{N}$ blocks}}\tabularnewline
    Layer & Parameters $(\mathtt{in},\mathtt{out})$ & Output shape $(C)$ \tabularnewline
    \hlinewd{1.6pt}
    % Encoder
    LayerNorm & - & $(768)$ \tabularnewline
    Linear-N-1 & $(768,2304)$ & $(2304)$ \tabularnewline
    Attention & - & $(768)$ \tabularnewline
    Linear-N-2 & $(768,768)$ & $(768)$ \tabularnewline
    LayerNorm & - & $(768)$ \tabularnewline
    Linear-N-3 & $(768,3072)$ & $(3072)$ \tabularnewline
    GELU & - & $(3072)$ \tabularnewline
    Linear-N-4 & $(3072,768)$ & $(768)$ \tabularnewline 
    \hlinewd{1.6pt}

    &&\tabularnewline 
    &\textbf{Decoder}& \tabularnewline 
    % Transformer Aggregator
    Layer & Parameters $(\mathtt{in},\mathtt{out},\mathtt{k},\mathtt{s},\mathtt{p})$ & Output shape $(C \times H \times W)$ \tabularnewline 
    \hlinewd{1.6pt}
    %\multicolumn{3}{c}{Postprocessing feature[2,5,8,11]} \tabularnewline \hline
    Conv-1(Linear-2-4) & (768,96,1,1,0) & (96,12,40) \tabularnewline
    ConvTranspose-1 & (96,96,4,4,0) & (96,48,160) \tabularnewline
    Conv-2 & (96,256,3,1,1) & (256,48,160) \tabularnewline \hline
    Conv-3(Linear-5-4) & (768,192,1,1,0) & (192,12,40) \tabularnewline
    ConvTranspose-2 & (192,192,2,2,0) & (192,24,80) \tabularnewline
    Conv-4 & (192,256,3,1,1) & (256,24,80) \tabularnewline \hline
    Conv-5(Linear-8-4) & (768,384,1,1,0) & (384,12,40) \tabularnewline
    Conv-6 & (384,256,3,1,1) & (256,12,40) \tabularnewline \hline
    Conv-7(Linear-11-4) & (768,768,1,1,0) & (768,12,40) \tabularnewline
    Conv-8 & (768,768,3,2,1) & (768,6,20) \tabularnewline
    Conv-9 & (768,256,3,1,1) & (256,6,20) \tabularnewline 
    \hlinewd{1.6pt}

    &&\tabularnewline 
    &\textbf{Fusion(x)} & \tabularnewline
    Layer & Parameters $(\mathtt{in},\mathtt{out},\mathtt{k},\mathtt{s},\mathtt{p})$ & Output shape $(C \times H \times W)$ \tabularnewline
    \hlinewd{1.6pt}
    ReLU & - & (256,h,w) \tabularnewline
    Conv-10 & (256,256,3,1,1) & (256,h,w) \tabularnewline
    RELU & - & (256,h,w) \tabularnewline
    Conv-11 & (256,256,3,1,1) &(256,h,w) \tabularnewline
    RELU & - & (256,h,w) \tabularnewline
    Fusion-1(Conv-9) & - & (256,6,20) \tabularnewline
    UpSample & - & (256,12,40) \tabularnewline 
    Conv-12 & (256.256,1,1,1) & (256,12,40) \tabularnewline \hline
    Fusion-2(Conv-6) & - & (256,12,40) \tabularnewline
    add-2(Fusion-2,Conv-12)& - & (256,12,40) \tabularnewline
    Fusion-4(add-2) & - & (256,12,40) \tabularnewline
    UpSample & - & (256,24,80) \tabularnewline 
    Conv-13 & (256.256,1,1,1) & (256,24,80) \tabularnewline \hline
    Fusion-5(Conv-4) & - & (256,24,80)\tabularnewline
    add-3(Fusion-5,Conv-13) & - & (256,24,80) \tabularnewline
    Fusion-6(add-3) & - & (256,24,80) \tabularnewline
    UpSample & - & (256,48,160) \tabularnewline 
    Conv-14 & (256.256,1,1,1) & (256,48,160) \tabularnewline \hline
    Fusion-7(Conv-2) & - & (256,48,160) \tabularnewline
    add-4(Fusion-7,Conv-14) & - & (256,48,160) \tabularnewline
    Fusion(add-4) & - & (256,48,160) \tabularnewline
    UpSample & - & (256,96,320) \tabularnewline 
    Conv-15 & (256.256,1,1,1) & (256,96,320) \tabularnewline\hline
    Conv-16 & (256,128,3,1,1) & (128,96,320) \tabularnewline
    UpSample & - & (256,192,640) \tabularnewline
    Conv-17 & (128,32,3,1,1) & (32,192,640) \tabularnewline
    ReLU & - & (32,192,640) \tabularnewline
    Conv-18 & (32,1,1,1,0) & (1,192,640) \tabularnewline
    sigmoid & - & (1,192,640) \tabularnewline 
    \hlinewd{1.6pt}
    \end{tabular}
    }\vspace{8pt}
\caption{\textbf{Network architecture of our model.}}
\label{tab:s_network}
\end{table}

\twocolumn

\clearpage

% % \clearpage

% % % \bibliographystyle{plain}
% % % \bibliography{aaai23}

% % %-------------------------------------------------------------------------
% % {\small
% % \bibliographystyle{ieee_fullname}
% % \bibliography{egbib}
% % }

% % \end{document}
%-------------------------------------------------------------------------

{\small
\bibliographystyle{ieee_fullname}
\bibliography{egbib}
}

\newpage
\clearpage

\end{document}